\documentclass[manuscript,screen]{acmart}
\usepackage{amsmath}
\usepackage{amsthm}
\usepackage{nicefrac}
\usepackage{algorithm}
\usepackage{tcolorbox}
\usepackage{apxproof}
\usepackage{xcolor}
\usepackage{lineno}
\usepackage{listings}

\definecolor{plotgreen}{HTML}{008400}
\definecolor{plotred}{HTML}{E80000}
\definecolor{grey}{HTML}{404060}
\definecolor{maroon}{HTML}{900010}

\theoremstyle{definition}

\usepackage{algorithm}
\usepackage[noend]{algpseudocode}

\usepackage{listings}
\usepackage{color} 

\definecolor{codegreen}{rgb}{0,0.6,0}
\definecolor{codegray}{rgb}{0.5,0.5,0.5}
\definecolor{codepurple}{rgb}{0.58,0,0.82}
\definecolor{backcolour}{rgb}{0.95,0.95,0.92}

\lstdefinestyle{mystyle}{
    backgroundcolor=\color{backcolour},   
    commentstyle=\color{codegreen},
    keywordstyle=\color{magenta},
    numberstyle=\tiny\color{codegray},
    stringstyle=\color{codepurple},
    basicstyle=\ttfamily\footnotesize,
    breakatwhitespace=false,         
    breaklines=true,                 
    captionpos=b,                    
    keepspaces=true,                 
    numbers=left,                    
    numbersep=5pt,                  
    showspaces=false,                
    showstringspaces=false,
    showtabs=false,                  
    tabsize=2
}

\AtBeginDocument{%
  }

\setcopyright{acmlicensed}
\copyrightyear{2018}
\acmYear{2018}
\acmDOI{XXXXXXX.XXXXXXX}

\acmConference[Conference acronym 'XX]{Make sure to enter the correct
  conference title from your rights confirmation emai}{June 03--05,
  2018}{Woodstock, NY}
\acmISBN{978-1-4503-XXXX-X/18/06}

\begin{document}

\title{EVOTER: Evolution of Transparent Explainable Rule sets}

\author{Hormoz Shahrzad}
\email{hshahrzad@cs.utexas.edu}
\orcid{0000-0002-5983-4531}
\affiliation{%
  \institution{University of Texas at Austin}
  \city{Austin}
  \state{Texas}
  \country{USA}
}
\affiliation{%
  \institution{Cognizant AI Labs}
  \city{San Francisco}
  \state{California}
  \country{USA}
}

\author{Babak Hodjat}
\email{babak@cognizant.com}
\orcid{0000-0002-4547-4731}
\affiliation{%
  \institution{Cognizant AI Labs}
  \city{San Francisco}
  \state{California}
  \country{USA}}

\author{Risto Miikkulainen}
\email{risto@cs.utexas.edu}
\orcid{0000-0002-0062-0037}
\affiliation{%
  \institution{University of Texas at Austin}
  \city{Austin}
  \state{Texas}
  \country{USA}
}
\affiliation{%
  \institution{Cognizant AI Labs}
  \city{San Francisco}
  \state{California}
  \country{USA}
}

\renewcommand{\shortauthors}{Shahrzad, Hodjat, and Miikkulainen}


\begin{abstract}
Most AI systems are black boxes generating reasonable outputs for given inputs. Some domains, however, have explainability and trustworthiness requirements that cannot be directly met by these approaches. Various methods have therefore been developed to interpret black-box models after training. This paper advocates an alternative approach where the models are transparent and explainable to begin with. This approach, EVOTER, evolves rule sets based on extended propositional logic expressions. The approach is evaluated in several prediction/classification and prescription/policy search domains with and without a surrogate.  It is shown to discover meaningful rule sets that perform similarly to black-box models. The rules can provide insight into the domain and make hidden biases explicit. It may also be possible to edit the rules directly to remove biases and add constraints. EVOTER thus forms a promising foundation for building trustworthy AI systems for real-world applications in the future.
\end{abstract}


%


\maketitle

\section{Introduction}
\label{sc:introduction}
Most of today's popular and powerful Artificial intelligence and machine learning approaches rely on inherently black-box methods like neural networks, deep learning, and random forests. Therefore, they are difficult to explain, and their functionality in critical/sensitive use cases cannot be thoroughly understood. It is also difficult to modify them to remove biases or add known constraints. In some domains, such lack of transparency can lead to costly failures (e.g., in self-driving cars), and in others, mandatory regulations (e.g. insurance) or ethical concerns (e.g. gender or racial biases) cannot be met \cite{Tjoa2021ASO}.

Methods have been developed to try to explain the behavior of black-box models (e.g., deep learning networks) by interrogating them after training. In this manner, it is possible to e.g.\ identify the areas of the input to which the model is paying attention when making a prediction. However, post-training interrogation-based explanations are not always accurate or complete, and they can be open to interpretation. 

Interpretability and explainability, therefore, are different concepts \cite{Explainable-AI,lage2019evaluation}. Interpretability can be defined as "the ability to explain or to present in understandable terms to a human" \cite{Doshi_2017}. Another popular definition is "the degree to which a human can understand the cause of a decision" \cite{miller2019}. These definitions point to a behavior-based interpretation of systems where the actual cause of the behavior is not transparent. In contrast, the internal processes of explainable models are transparent and can be understood in mechanical terms. Thus, explainability means interpretability with transparency. As a result, explainable systems can be audited for compliance, which is useful for various stakeholders, from data scientists and business owners to risk analysts and regulators \cite{explainable-ML, xai-survey-2018}.

This paper introduces such an approach, the \textbf{EVO}lution of \textbf{T}ransparent \textbf{E}xplainable \textbf{R}ule sets, or EVOTER. These rule sets are transparently explainable and lead to domain insights that would be difficult to achieve through other machine-learning methods.  The paper begins with a review of earlier methods that paved the way for the EVOTER approach, emphasizing key distinctions between them. Next, the EVOTER rule representation approach, applicable to both prediction/classification and prescription/policy search tasks, is presented. Experiments with EVOTER in a range of real-world domains then demonstrate that the approach is general and effective:
\begin{enumerate}
\item EVOTER is as accurate and less complex, and thus more explainable, than other transparent machine learning methods such as naive Bayes, decision trees, and random forests (shown in UCI iris, breast cancer, and human activity recognition tasks).

\item Rule-set evolution results in effective classification (shown in the iris, breast cancer, and static human activity recognition tasks) and effective static prescription (shown in the heart-failure prevention and diabetes treatment tasks).

\item Rule-set evolution results in effective time-series prediction (shown in the dynamic human activity recognition and blood-pressure classification tasks) and effective policy search (shown in the flappy bird and cart-pole control task).

\item It results in more interpretable explanations than other transparent methods such as naive Bayes, decision trees, and random forests (in cart-pole control)

\item Similar results are obtained both through direct evolution in the task itself and through evolution with a surrogate model of the task (in cart-pole control).

\item EVOTER leads to useful insights, including possibly identifying biases in the results (in diabetes treatment).

\item Explainability does not incur a major cost in accuracy. The performance of evolved rule sets is similar to that of neural networks trained with the same data, both when the task has a single objective (in heart-failure prevention) and when it has two objectives (in diabetes treatment).

\item The evolved rule sets are compressible, and only a small number of rules apply to each case, making the approach focused and economical (in heart-failure prevention).
\end{enumerate}
The paper thus demonstrates that EVOTER is an effective and versatile approach to forming transparent and explainable models in machine-learning tasks.

\section{Background}
\label{sc:background}

Explainability refers to the ease with which a human can understand the mechanisms behind a model's behavior \cite{Schwalbe2021ACT}. Explainability is difficult to capture quantitatively because it would require quantifying psychological processes, and there are outside factors that matter: Models are more interpretable if they align well with domain knowledge and domain-specific expectations, and if their behavior can be visualized e.g.\ through plotting decision surfaces or clustering the types of behaviors. In terms of building explainable AI models, there are two main directions: (1) use a black-box modeling method and make it more transparent; and (2) use a transparent method and make it more interpretable. These approaches are reviewed in the subsections below.

\subsection{Making Black-Box Methods More Transparent}

Black-box learning methods such as deep learning have been shown powerful in many domains. However, they are inherently difficult to explain because behavior is defined by nonlinear interactions of a very large number of parameters (i.e.\ connection weights). Recently, techniques have been developed to make their behaviors transparent \cite{darpa_xai_2019,xai-survey-2018,shap,LIME,MUSE,explain-deepNN}. For instance, SHAP (SHapley Additive exPlanations; \cite{shap}) explains the output of any model as a sum of the effects of its features, offering local interpretability based on Shapley values from cooperative game theory. It can allocate credit consistently and fairly among features and is model-agnostic. However, it is also computationally intensive and the information it provides is often overwhelming. Similarly, LIME (Local Interpretable Model-agnostic Explanations; \cite{LIME}) provides explanations for individual predictions by approximating the model locally with a transparent model. It can be applied to any model type and it is good at explaining specific predictions, but it may not explain global behavior well. 

Such explainability techniques can be evaluated along several dimensions \cite{Bodria2021BenchmarkingAS,Nauta2022FromAE}, i.e.\ whether the explanations they provide have high \textit{fidelity}, which measures how accurately the explanations reflect the behavior of the original model; \textit{stability}, evaluating the consistency of explanations under slight variations in input; \textit{comprehensiveness}, ensuring that explanations capture all relevant aspects influencing the model's decisions; \textit{sufficiency}, assessing whether the provided explanations are adequate for understanding the model's decisions; and \textit{transparency}, quantifying how transparent the model is in terms of its decision-making process. These metrics facilitate a quantitative assessment of how interpretable and understandable a model is as a whole, enabling stakeholders to gauge the trustworthiness and reliability of the model's decisions.

\subsection{Making Transparent Models More Interpretable}

In the second direction, the substrate of the model is replaced with an inherently explainable structure, such as a probability set, decision tree, random forest, or rule set. Instead of e.g.\ running gradient descent in a deep neural network, a transparent representation of knowledge is constructed.  Thus, the resulting model consists of a human-readable set of rules or equations. By reviewing this set, it is possible to understand what input features are used and how they are brought together in rules to make a decision. It is thus possible to explain the behavior of the model comprehensively in an exact human-readable form.

Classic such approaches include naive Bayes for learning class probabilities, decision trees for learning a hierarchy of tests on selected features to arrive at a classification, and random forests as an ensemble of such trees. They each have their own learning algorithms, or they can be evolved, especially to optimize their structure and hyperparameters automatically as well \cite{miikkulainen:emlchapter23}. The most general such approach, however, is to learn sets of rules. Because rule sets are highly structured, evolution is well-suited for learning them. Many early such systems were based on a fixed- or variable-length-chromosome genetic algorithm (GA) or a tree- or list-based genetic programming (GP) \cite{Srinivasan2011,Pearroya2004PittsburghGM,MOCA,RIPPER}. In these methods, the rules were propositional logic expressions where each term is built from a feature compared against a constant, for example, IF (Quality = Medium OR High) AND (Advertisement = Yes OR Telemarketing = Yes) AND (Gifts = Yes) AND (Sales Profile = Good OR Medium) THEN (Profit = Good) ELSE (Profit = Low)\cite{Srinivasan2011}. These techniques were applied to several data sets in the UCI machine learning repository \cite{Dua:2019}.

EVOTER, a type of Grammatical Evolution (GE), advances rule-evolution systems by using a list-based grammatical approach to evolve interpretable rule sets. However, unlike traditional GE methods, which often focus on evolving programs or symbolic expressions for tasks like symbolic regression \cite{oneill1998grammatical, fonseca2023comparing, fernandes2023hotgp}, EVOTER focuses on producing interpretable rule sets for classification and decision-making tasks. Traditional GE approaches often result in complex models, which can be challenging to interpret, especially in domains like healthcare and finance that require high transparency.

In contrast, EVOTER emphasizes simplicity by employing a list-based grammar to evolve rules that are easy to understand, while maintaining fair accuracy. This focus on interpretability ensures that EVOTER's models can be easily understood, verified, and adjusted by domain experts, balancing accuracy and transparency in practical applications. In EVOTER, rules are extended to a more powerful syntax that brings about three key advantages:
\begin{enumerate}
    \item Time lags can be evolved directly for features that form a time series;
    \item Features can be compared with each other; and
    \item Coefficients and exponents can be evolved to implement linear and nonlinear operations on features.
\end{enumerate}
This approach makes more complex applications possible, as will be demonstrated systematically in this paper. However, they also make the rule-set solutions more interpretable. Each of these extensions allows for more compact coding of behavior. Therefore, the resulting rule sets are simpler, with fewer features, clauses, and rules, and therefore easier for humans to understand.

Whereas human interpretability in general is difficult to quantify, simplicity (or complexity) of rule sets and other transparent machine-learning approaches can be readily measured, simply by counting the free parameters in them, or the number of decision points in them. Moreover, metrics have been developed for model selection that combine simplicity with performance, thus identifying models that represent good tradeoffs between quality and explainability. For instance, AIC (Akaike Information Criterion; \cite{akaike1974}) estimates the relative amount of information lost by a given model: the less information a model loses, the higher the quality of that model:
\begin{equation}
    \text{AIC} = 2k - 2\ln(L)
\end{equation}
where $k$ is the number of parameters in the model and $L$ is the maximum value of the likelihood function for the model. AIC balances the model accuracy and complexity, even when models are non-nested or have different numbers of parameters, and does not require prior knowledge of the true model. However, it may favor more complex models, potentially leading to overfitting, and does not incorporate sample size, which can be a drawback in large datasets. In contrast, BIC (Bayesian Information Criterion; \cite{schwarz1978}) adjusts the AIC by incorporating a penalty for the number of parameters proportional to the log of the sample size:
\begin{equation}
    \text{BIC} = \ln(n)k - 2\ln(L)
\end{equation}
where $n$ is the sample size. BIC results in more consistent model selection, favoring the true model as sample size increases. On the other hand, it can be overly punitive towards models with more parameters in small samples, and it may not perform well if the true model is not among the considered  candidates. Therefore, this paper includes both the AIC and BIC metrics in evaluating the useful complexity of the models. Given that the models are already transparent, these metrics serve as proxies for their interpretability, and thus explainability.

\subsection{Deploying Explainable Models}

How does one determine which explainability technique to apply to a given ML model? Post-facto explainability methods are useful for pre-existing deep learning-based models that are hard to replace, given the cost of training. In many cases, deep learning is the preferred model type, especially in very large data problems such as unstructured text, sound, images, and video. In contrast, in problems where feature engineering has produced a structured data set with familiar input and output features (e.g., tabular data), rule-set evolution can be competitive and, in some cases, even more accurate, making it a preferable choice.

However, rule-set evolution can also be applied to unstructured data sets with continuous input or output features, and deep learning methods can sometimes do well in structured and tabular data problems. So the second, equally important consideration is whether interpretability is sufficient, or whether transparency is required. If explainability is, for example, mandated by regulations (as it is e.g.\ in healthcare, insurance, and finance), then it is important to consider transparently explainable modeling techniques such as EVOTER. For that reason, many of the examples in this paper are drawn from the healthcare domain.

\section{Method}
\label{sc:method}

In this section, the design of the rule-set representations is first reviewed, followed by the methods for evolving them effectively.

\subsection{Representing Behavior as Rule Sets}

\begin{figure}[!ht]
\centering
\begin{minipage}{\textwidth}
\centering
{\small
\setlength{\lineskip}{-1pt}
\begin{align*}
\textcolor{blue}{<\textit{rules}>} &::=~\hspace*{0.5ex} \textcolor{orange}{<\textit{rule}>} \,|\, \textcolor{orange}{<\textit{rule}>} \textbf{OR} \textcolor{blue}{<\textit{rules}>}\\
\textcolor{orange}{<\textit{rule}>} &::=~\hspace*{0.5ex} \textcolor{purple}{<\textit{conditions}>}~ \longrightarrow [\textcolor{green}{<\textit{certainty}>}]\textcolor{black}{<\textit{action}>}\\
\textcolor{purple}{<\textit{conditions}>} &::=~\hspace*{0.5ex} \textcolor{teal}{<\textit{condition}>} \,|\, \textcolor{teal}{<\textit{condition}>} \textbf{AND} \textcolor{purple}{<\textit{conditions}>}\\
\textcolor{teal}{<\textit{condition}>} &::=~\hspace*{0.5ex} \textcolor{grey}{<\textit{leading}>}\textcolor{violet}{<\textit{operator}>}\textcolor{maroon}{<\textit{trailing}>}\\
\textcolor{grey}{<\textit{leading}>} &::=~\hspace*{0.5ex} \textcolor{red}{<\textit{coefficient}>}*\textcolor{brown}{<\textit{feature}>}^{[\textcolor{olive}{<\textit{power}>}]}{[\textcolor{olive}{(\textit{lag})}]}\\
\textcolor{red}{<\textit{coefficient}>} &::=\hspace*{0.5ex} 0.[0-9]+\\
\textcolor{brown}{<\textit{feature}>} &::=\hspace*{0.5ex} \textcolor{black}{\text{(an arbitrary input feature)}}\\
\textcolor{olive}{<\textit{power}>} &::=\hspace*{0.5ex} [1-3] \textcolor{black}{\text{(keep it to more meaningful values.)}}\\
\textcolor{olive}{<\textit{lag}>} &::=\hspace*{0.5ex} [0-9] \textcolor{black}{\text{(or the desired lag range to be explored.)}}\\
\textcolor{violet}{<\textit{operator}>} &::=\hspace*{0.5ex} "\textcolor{black}{=}" \hspace*{2ex}|\hspace*{2ex} "\textcolor{black}{\neq}" \hspace*{2ex}|\hspace*{2ex} "\textcolor{black}{<}" \hspace*{2ex}|\hspace*{2ex} "\textcolor{black}{\le}" \hspace*{2ex}|\hspace*{2ex} "\textcolor{black}{>}" \hspace*{2ex}|\hspace*{2ex} "\textcolor{black}{\ge}"\\
\textcolor{maroon}{<\textit{trailing}>} &::=~\hspace*{0.5ex} \textcolor{grey}{<\textit{leading}>} \,|\, \textcolor{purple}{<\textit{value}>}\\
\textcolor{purple}{<\textit{value}>} &::=\hspace*{0.5ex} \textcolor{black}{\text{(a number between min and max feature value)}}\\
\textcolor{green}{<\textit{certainty}>} &::=~\hspace*{0.5ex} \textcolor{red}{<\textit{coefficient}>} \textcolor{black}{\text{(confidence of a prediction or action)}}\\
\textcolor{black}{<\textit{action}>} &::=\hspace*{0.5ex} \textcolor{black}{\text{(a predicted class or prescribed action)}}
\end{align*}
{\footnotesize ($a$) BNF representation of EVOTER rules}
}
\end{minipage}\\[2ex]
\begin{minipage}{\textwidth}
\centering
\includegraphics[width=0.8\linewidth]{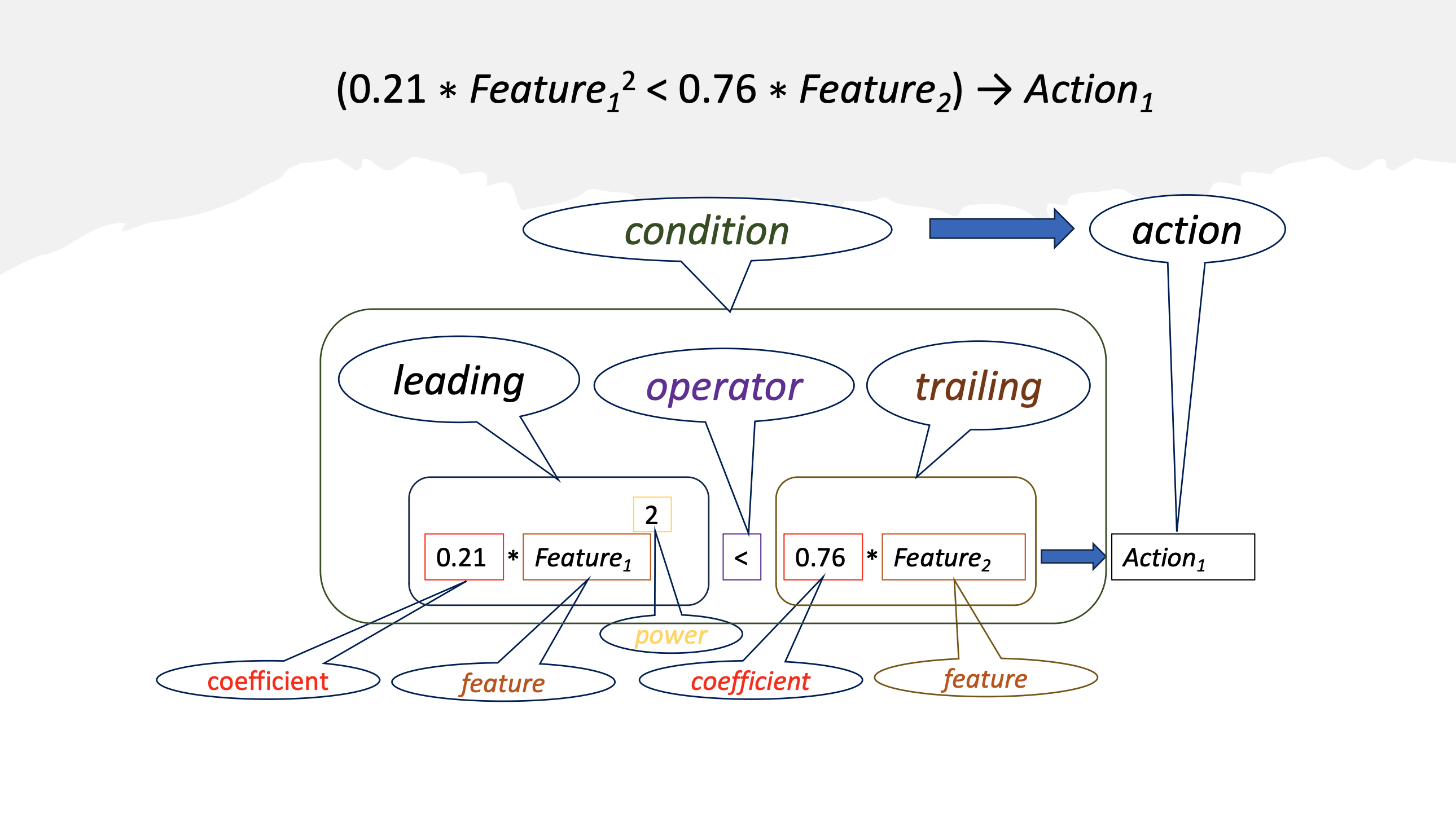}\\[-6ex]
{\footnotesize ($b$) An example EVOTER rule}
\end{minipage}
\vspace*{-1ex}
\caption{\emph{Rule-set Representation in EVOTER} ($a$) The BNF grammar for the rules, including time lags, feature comparisons, and feature exponentiation. ($b$) An example rule illustrating these concepts, with elements color-coded as in the grammar. Rules can have multiple conditions and a rule set consists of multiple such rules (not shown). In addition, EVOTER keeps track of how many times the antecedent is satisfied, and the rule set includes a default rule that is applied if none of the antecedents apply in the current situation.}
\vspace*{-2ex}
\label{fig:rules}
\end{figure}

Rule sets in EVOTER are based on propositional logic expressions. They are collections of statements of the form “IF antecedent A is met THEN consequent B occurs”; the antecedents are conjunctions of conditions, and logical OR applies between the rules (Figure~\ref{fig:rules}).  Conditions compare single features with constant values or compare features with each other; coefficients are used to scale the features linearly, power expressions to scale them nonlinearly, and time lags to access feature values at different points in time.

There are several reasons why such rules are a good representation. First, they closely resemble verbal reasoning we use in our daily lives. Thus, they are interpretable and transparent, which is vital in applications where predictions need to be auditable so that experts can understand how and why a forecast or a recommendation was made. Second, such rules have a linear list structure, which helps avoid the usual problems of tree evolution (e.g., bloat and complex mutation and crossover operators) while remaining logically complete. Third, they can uncover nonlinear relationships and interactions among the domain features, as well as their temporal dynamics. Fourth, the conditions can be easily augmented to represent probabilities, as well as a variety of further mathematical operations and functions \cite{Hodjat2018}.

When parsing an individual's rule set, conditions are evaluated in order, and all actions for conditions that are met are returned. In some domains (e.g.\ those in Sections~\ref{sc:prediction}-~\ref{sc:esp}), only the first action is executed; in others (e.g.\ Sections~\ref{sc:nn}--\ref{sc:insight}), a hard-max filter is used to select one of the actions based on the action coefficients; in yet others, all actions may be executed in parallel or in sequence. There are two additional extensions at this level: (1) Each rule has a $\mathit{times\_applied}$ field, which is incremented each time its antecedent evaluates to true when applied to a data point. This bookkeeping helps simplify the rule set and to reduce bloat, as described below. (2) A \emph{default} rule is added at the end of each rule set, ensuring that some action is always produced even if no specific rules are triggered. This mechanism also makes evolution more efficient: Often the most common action is discovered as the default, and other rules then fine-tuned to address exceptions.

\begin{figure}[!t]
\begin{minipage}{0.6\textwidth}
\centering
\includegraphics[width=\linewidth]{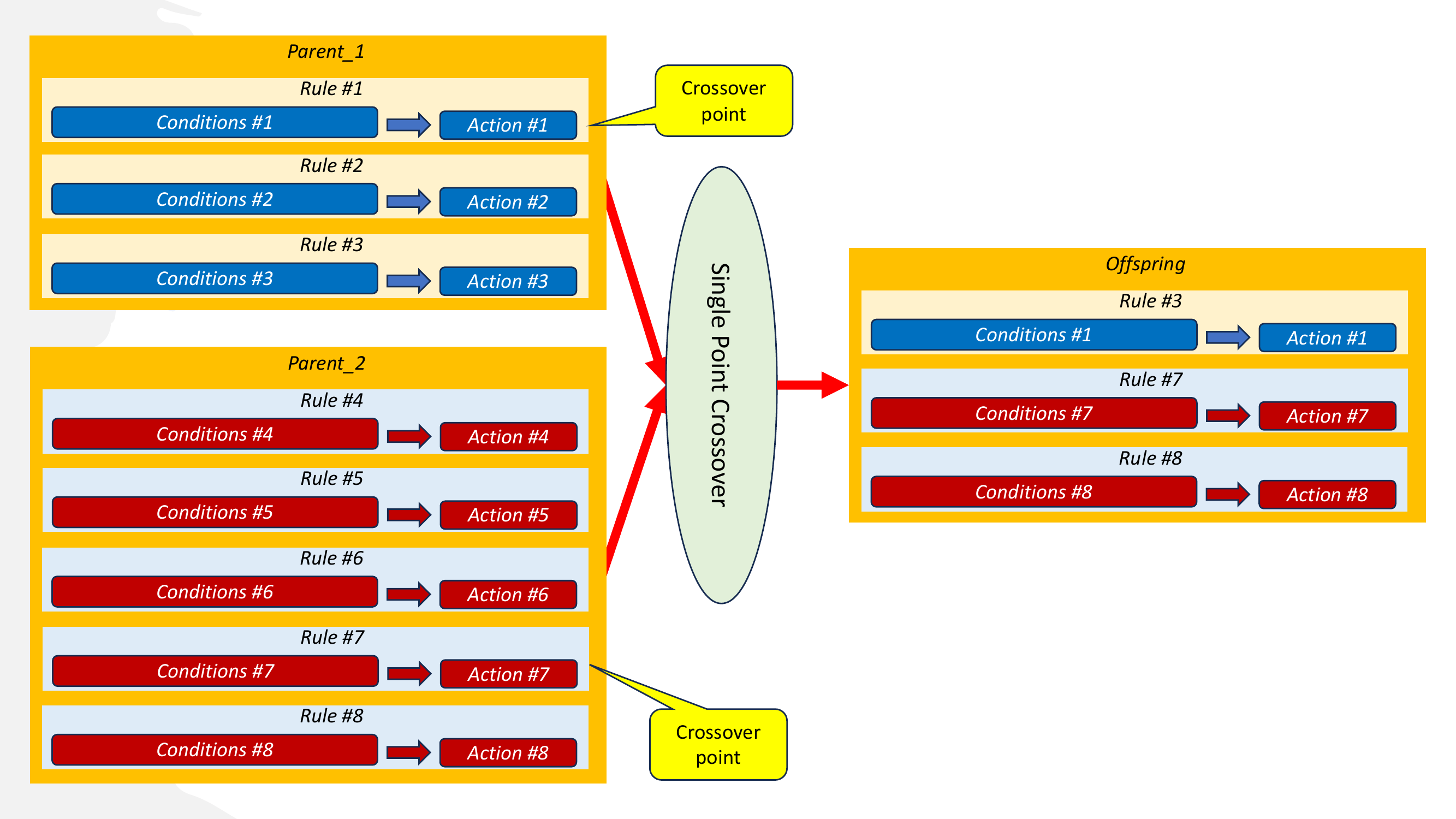}\\[-1ex]
{\footnotesize ($a$) Single-point crossover\rule{18ex}{0ex}}
\end{minipage}
\hfill\mbox{}\\[20ex]
\begin{minipage}{0.6\textwidth}
\centering
\includegraphics[width=\linewidth]{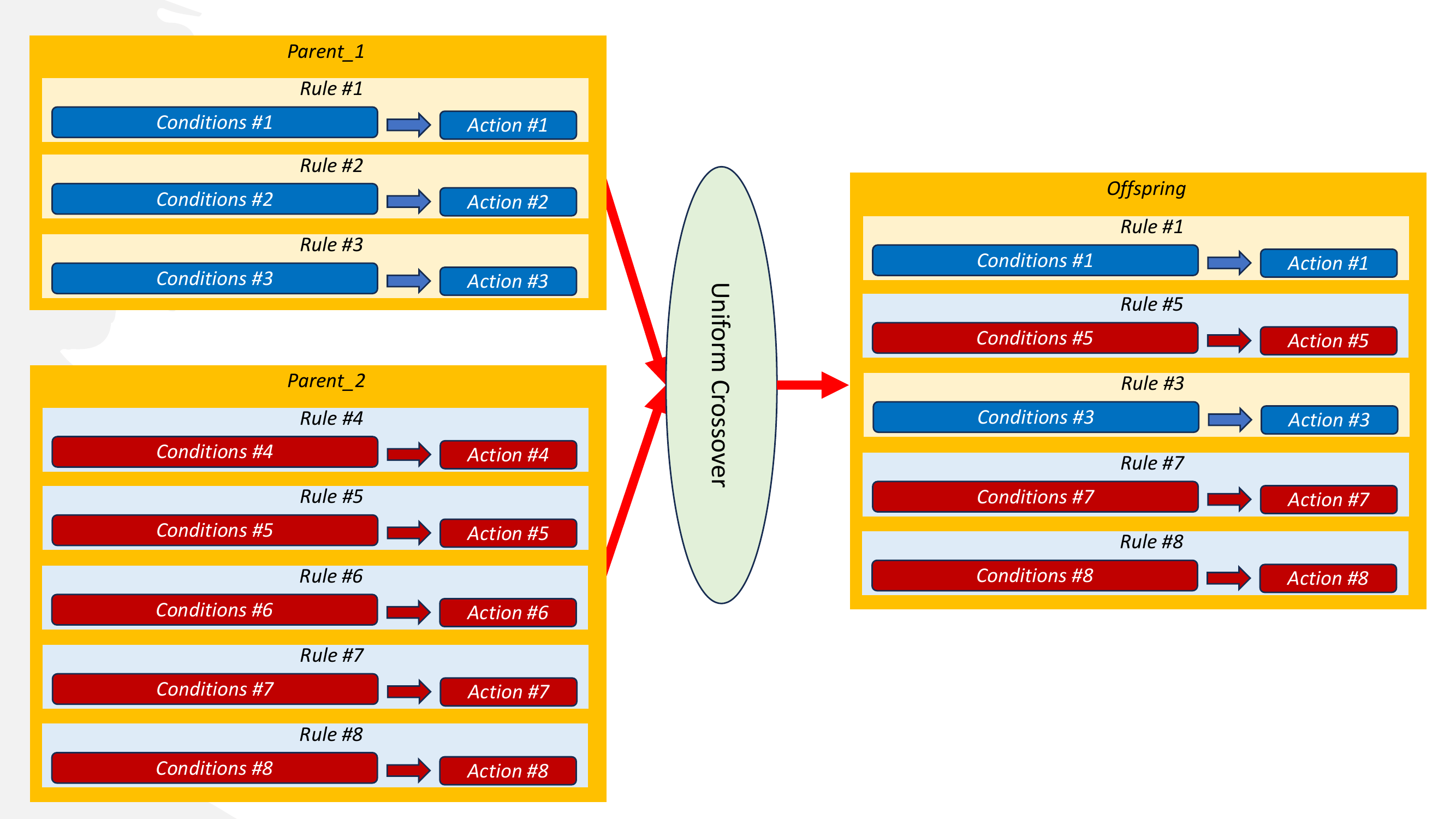}\\[-1ex]
{\footnotesize ($b$) Uniform crossover\rule{18ex}{0ex}}
\end{minipage}
\hfill\mbox{}\\[-63ex]
\hfill\mbox{}
\begin{minipage}{0.6\textwidth}
\centering
\includegraphics[width=\linewidth]{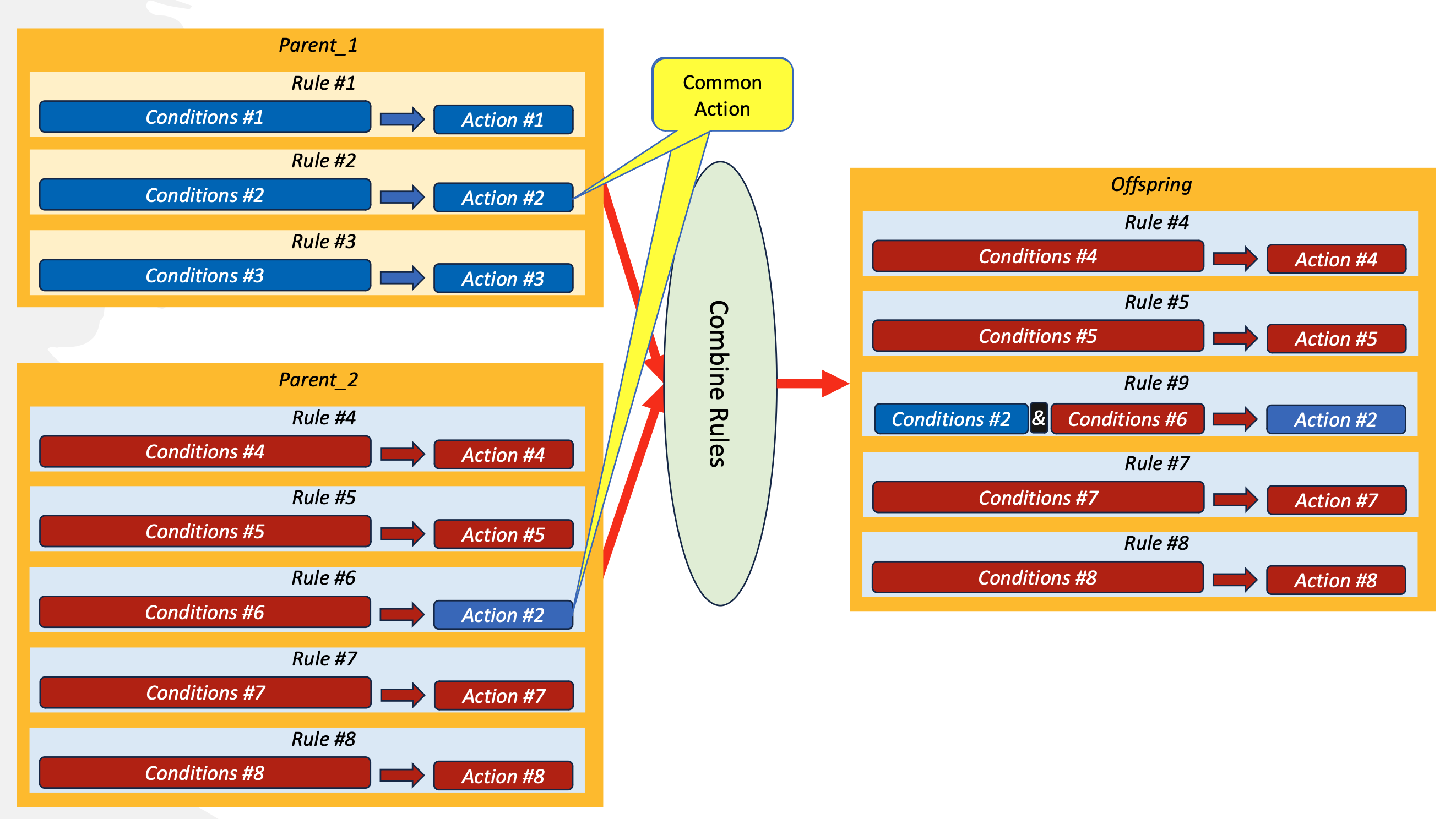}\\[-1ex]
{\footnotesize\rule{20ex}{0ex} ($c$) Combine-rules crossover}
\end{minipage}\\[28ex]
\caption{\emph{The three crossover strategies in EVOTER.} ($a$) In single-point crossover, the offspring inherits rules from $parent_1$ up to its crossover point, followed by rules from $parent_2$ after its crossover point. ($b$) In uniform crossover, each rule is selected uniformly randomly from either parent at each point in the rule set; in addition, all remaining rules are included from the larger set. ($c$) In combine-rules crossover, the offspring is initially cloned from $parent_2$; a rule with a matching action in $parent_1$ and $parent_2$ is then selected, and its conditions are augmented with conditions from $parent_1$. Each of these crossover strategies serves a different purpose, and they are selected randomly during each crossover.}
\label{fig:rule_crossover}
\end{figure}

\subsection{Evolving Rule Sets}

Rule sets are evolved through genetic algorithms, adapted to take advantage of their structure. Two parents are first chosen through tournament selection, and offspring are generated through various crossover and mutation methods. These methods are specified in pseudocode in Appendix~\ref{ap:algos}.

To implement crossover, a high-level rule-set crossover method (Algorithm~\ref{alg:crossover}) calls one of three different strategies.  First, in \emph{single-point crossover} (Algorithm~\ref{alg:single_point_crossover}; Figure~\ref{fig:rule_crossover}$a$), a crossover point is first chosen separately in each parent's rule set. Offspring is then generated by concatenating the rules above the crossover point of the first parent with the rules below the crossover point of the second parent. In this manner, the number of rules in the offspring can grow or shrink compared to the parents. Second, in \emph{uniform crossover} (Algorithm~\ref{alg:uniform_crossover}; Figure~\ref{fig:rule_crossover}$b$), rules are randomly selected from either parent at each point in the rule set, thus forming a uniform mixture of their abilities. Third, in \textit{combine-rules crossover} (Algorithm~\ref{alg:combine_rules}; Figure~\ref{fig:rule_crossover}$c$), conditions for rules that have the same action are combined, thus potentially forming a more precise rule that replaces the old rule in one parent. The three strategies thus each serve a different purpose. They are selected randomly at each crossover.

\begin{figure}[!t]
\centering
\footnotesize
\setlength{\lineskip}{-1pt}
\includegraphics[width=\linewidth]{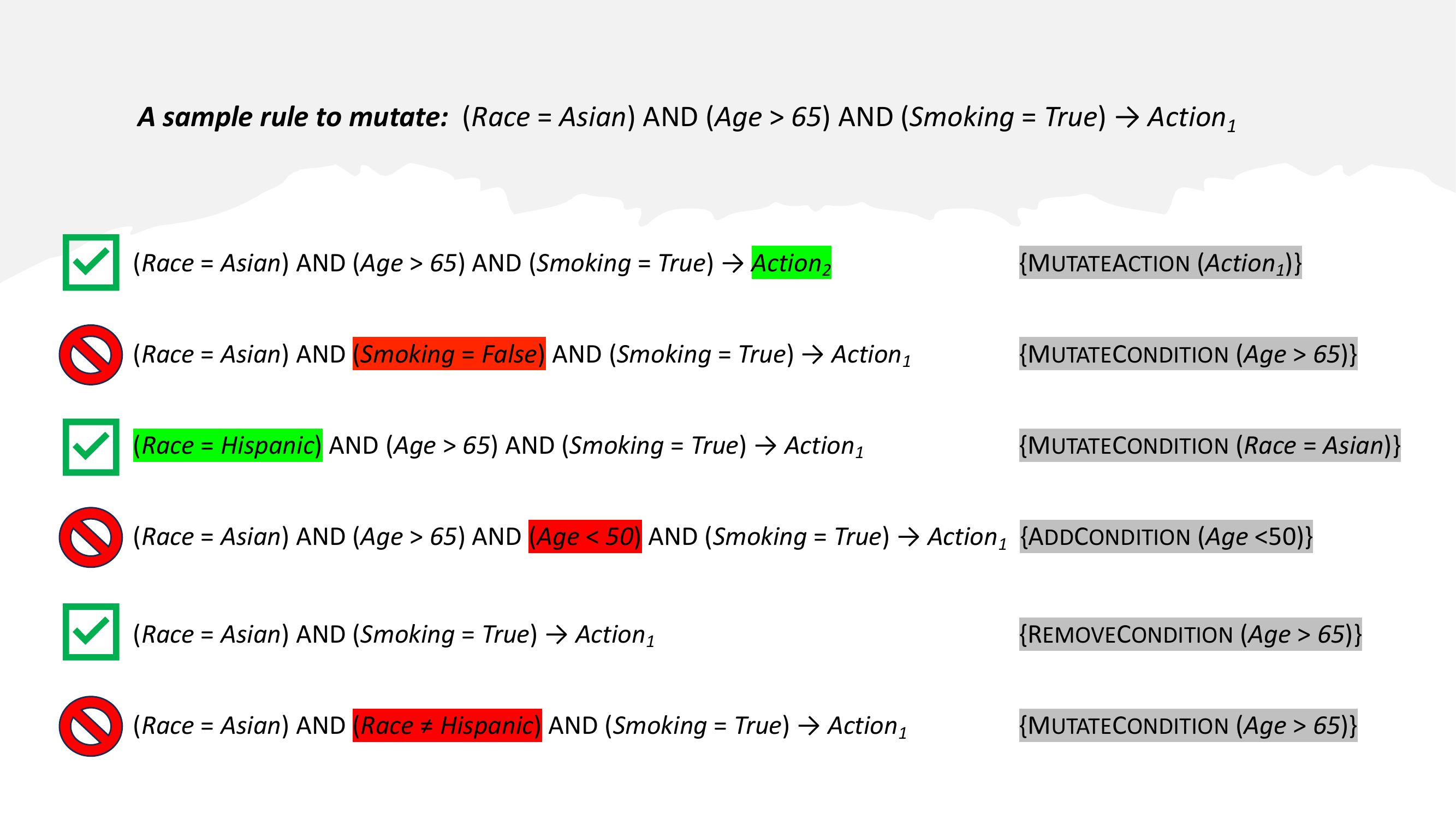}\\
\vspace*{-6ex}
\caption{\emph{Examples of permitted and rejected rule mutations in EVOTER}. Gray highlighting on the right specifies the proposed mutation, green highlighting on the left identifies the results of permitted mutations, and red highlighting identifies the rejected mutation. In this manner, mutations that lead to contradictions and redundancies are automatically ruled out, making evolution more efficient.}
\label{fig:rule_mutation}
\end{figure}

Mutations are implemented as single random changes to elements in the rule set. \emph{MutateConditions} (Algorithm~\ref{alg:mutate_condition}) operates at the condition level, altering elements like the condition's value, operator, or time lags. \emph{MutateRule} (Algorithm~\ref{alg:mutate_rule}) operates at the rule level: It can modify a rule's action, replace, add, or remove conditions, or change the rule's certainty. If any such mutation would result in a contradiction or redundancy among the conditions, it is immediately rejected, thus making evolution more efficient (Figure~\ref{fig:rule_mutation}). \emph{MutateRuleSet} (Algorithm~\ref{alg:mutate_ruleset}) applies at the rule-set level, adding a random rule to the individual, removing an entire rule from the individual, changing the default rule, or changing the rule order. Note that these mutations can make the offspring smaller or larger than the original individual.

The $\mathit{times\_applied}$ counter is used to filter out inactive individual rules from participating in offspring generation (Algorithm ~\ref{alg:crossover}). To reduce bloat, all conditions recognized as tautologies or falsehoods are removed from offspring individuals (Algorithm~\ref{alg:prune}). While this process works well, several extensions can be implemented to make it more efficient in large-scale experiments, as outlined in Section~\ref{sc:discussion}. The experiments that follow focus primarily on quantitative comparisons to other transparent machine learning methods, and on illustrating the capabilities of the approach in various challenging real-world domains.

\section{Quantitative Comparison of Transparent Methods}
\label{sc:quantitative}
\label{subsec:comparison_class_models}

The  experiments in this section show quantitatively that EVOTER is equally accurate than other transparent machine learning methods, while at the same time significantly less complex and more explainable in terms of metrics such as AIC and BIC. Three standard methods were included: naive Bayes, decision trees, and random forests. The comparison was run on six UCI classification datasets, each with unique characteristics and challenges:
\begin{enumerate}
    \item \textbf{Iris} tests the method's ability to classify flower species based on morphological measurements. It comprises three classes (versicolor, virginica, setosa) across 150 data points with four features.
    \item \textbf{Breast Cancer} challenges methods to distinguish between benign and malignant breast cancer tumors. It contains two classes (malignant and benign) with 30 features across 561 data points.
    \item \textbf{Human Activity Recognition (Static Features)} tests the method's capacity to recognize human activities from featurized sensor data. Six activity classes (WALKING, WALKING UPSTAIRS, WALKING DOWNSTAIRS, SITTING, STANDING, and LAYING) are described by 561 static features derived from nine different gyro signals sampled at 50 Hz from smartphones. The dataset has 10,299 data points.
    \item \textbf{Human Activity Recognition (Time-series Features)} evaluates how well the methods handle time-series data. The same six activity classes are represented by 1152 features, i.e.\ 128 samples of the nine gyro measurements over time. This variant also contains 10,299 data points.
    \item \textbf{Credit Approval} focuses on predicting whether a credit card application should be approved based on various personal and financial attributes. It has two classes (approved or denied) with 690 instances and 15 attributes.
    \item \textbf{Contraceptive Method Choice} involves classifying the choice of contraceptive methods (no use, long-term, or short-term methods) based on demographic and socio-economic attributes. The dataset consists of 1,473 instances with 10 attributes.

\end{enumerate}

As usual with UCI datasets, accuracy was measured through the proportion of correct classifications. For the Naive Bayes experiments, the \texttt{GaussianNB()} implementation from the \texttt{scikit-learn} library was used; it is specifically designed to handle continuous data by modeling each feature as a Gaussian (normal) distribution. To measure model complexity, the number of parameters was calculated. For Naive Bayes models, they include the mean and standard deviation for each feature within each class; thus, the total number of parameters equals two times the number of classes times the number of features.
For decision trees, the parameters comprise the key decision-making points in the model's structure, i.e.\ the number of decision nodes (excluding leaf nodes.  For random forests, they consist of all such decision nodes across all trees. For EVOTER, analogously to decision nodes, the total number of conditions across all rules in the rule set was counted. As proxies for explainability, the AIC and BIC metrics were calculated for each model. The structure of the methods and the datasets made the AIC and BIC calculations efficient, as detailed in Appendix~\ref{ap:aicbic}.

All models were trained on a randomly chosen 70\% partition of the data, with the remaining 30\% reserved as an unseen test set to obtain the accuracy measurement. This procedure was repeated ten times for each experiment using bootstrapping, a statistical method where the training and test sets are re-sampled with replacement for each run. Bootstrapping captures the variability in the data and provides a more accurate assessment of the model’s generalization ability than e.g.\ k-fold cross-validation for small datasets and for datasets where maintaining the exact class distribution is critical~\cite{efron1992bootstrap, fushiki2011bootstrap, hastie2009elements}. The average accuracy, complexity, AIC, and BIC were calculated across the ten runs and statistical significance of differences between models estimated.

The results are summarized in Tables~\ref{tab:model_comparison_iris}--\ref{tab:model_comparison_contraceptive_choice}. In terms of accuracy, EVOTER performs as well as the best model in each case (the 0.01 difference from RF is not statistically significant; $p> 0.064$ in a paired $t$-test). However, in terms of the number of parameters and AIC and BIC, EVOTER is in a category of its own, reducing these metrics by at least an order of magnitude, and sometimes up to seven orders of magnitude. Thus, EVOTER is significantly more explainable than the other methods. This result is further elucidated in the accuracy vs.\ complexity trade-offs shown in Appendix~\ref{ap:tradeoffs}.

The domain of human activity recognition using time-series features (as shown in Table~\ref{tab:model_comparison_activity_raw}) highlights the first major extension of EVOTER over other rule-set systems. Out of 1,154 total time-series features, EVOTER utilized an average of only 142. The set of features was remarkably consistent across the ten runs: On average, 88\% of these features were the same, whereas picking 142 features randomly would result in only 6\% overlap. Thus, EVOTER consistently identified the points in the time series where each feature was most useful, and included only those in the rule sets. This ability can be coded more compactly in EVOTER by allowing it to evolve a time lag as a parameter of the feature directly. This extension will be evaluated in the experiment in Section~\ref{sc:prediction}.

\begin{table}[H]
\centering
{\footnotesize
\begin{tabular}{lrrrr}
\hline
Model         & Accuracy & Number of Parameters & \rule{30pt}{0pt}AIC    & \rule{30pt}{0pt}BIC      \\ \hline
Naive-Bayes   & 0.95     & 24                & 532 & 604   \\
Decision Tree & \textbf{1.00}     & 50                & 100 & 252   \\
Random Forest & \textbf{1.00}     & 525               & 1,085 & 2,666 \\
EVOTER        & 0.99     & \textbf{5}         & \textbf{83}  & \textbf{98}    \\ \hline
              &          &                      &        &         \\
\end{tabular}}
\vspace*{-2ex}
\caption{\emph{Model Comparison for Iris.}}
\label{tab:model_comparison_iris}

\centering
{\footnotesize
\begin{tabular}{lrrrr}
\hline
Model         & Accuracy & Number of Parameters & \rule{30pt}{0pt}AIC    & \rule{30pt}{0pt}BIC      \\ \hline
Naive-Bayes   & 0.94     & 120               & 2,589 & 3,110  \\
Decision Tree & 0.98     & 127               & 1,111 & 1,662  \\
Random Forest & \textbf{0.99}     & 2014     & 4,574 & 13,322 \\
EVOTER        & 0.98     & \textbf{5}         & \textbf{795}  & \textbf{816}   \\ \hline
              &          &                      &         &          \\
\end{tabular}}
\vspace*{-2ex}
\caption{\emph{Model Comparison for Breast Cancer.}}
\label{tab:model_comparison_breast}

\centering
{\footnotesize
\begin{tabular}{lrrrr}
\hline
Model         & Accuracy & Number of Parameters & \rule{30pt}{0pt}AIC    & \rule{30pt}{0pt}BIC      \\ \hline
Naive-Bayes   & 0.75     & 6,732              & 189,059   & 237,797    \\
Decision Tree & 0.96     & 262,143            & 552,635   & 2,450,499   \\
Random Forest & \textbf{0.97}     & 41,184,451          & 82,388,229 & 380,555,502 \\
EVOTER        & 0.96     & \textbf{35}        & \textbf{28,580}    & \textbf{29,025}     \\ \hline
              &          &                      &             &              \\
\end{tabular}}
\vspace*{-2ex}
\caption{\emph{Model Comparison for Human Activity Recognition (Static Features).}}
\label{tab:model_comparison_activity}

\centering
{\footnotesize
\begin{tabular}{lrrrr}
\hline
Model         & Accuracy & Number of Parameters & AIC           & BIC           \\ \hline
Naive-Bayes   & 0.76     & 13,824             & 196,957     & 297,040     \\
Decision Tree & 0.92     & 2,097,151           & 4,250,289    & 19,433,248   \\
Random Forest & \textbf{0.94}     & 587,779,267         & 1,175,600,063 & 5,431,005,623 \\
EVOTER        & 0.93     & \textbf{61}        & \textbf{49,871}      & \textbf{50,126}      \\ \hline
              &          &                      &               &               \\
\end{tabular}}
\vspace*{-2ex}
\caption{\emph{Model Comparison for Human Activity Recognition (Time-series Features).}}
\label{tab:model_comparison_activity_raw}
\vspace*{-4ex}
\end{table}

\begin{table}[H]
\centering
{\footnotesize
\begin{tabular}{lrrrr}
\hline
Model         & Accuracy & Number of Parameters & \rule{30pt}{0pt}AIC    & \rule{30pt}{0pt}BIC      \\ \hline
Naive-Bayes   & 0.85     & 987                & 2,457 & 2,589   \\
Decision Tree & 0.90     & 15,432             & 3,290 & 5,012   \\
Random Forest & \textbf{0.93}     & 56,743             & 12,543 & 23,891 \\
EVOTER        & 0.91     & \textbf{150}        & \textbf{1,125}  & \textbf{1,342}    \\ \hline
              &          &                      &        &         \\
\end{tabular}}
\vspace*{-2ex}
\caption{\emph{Model Comparison for Credit Approval.}}
\label{tab:model_comparison_credit_approval}
\vspace*{-4ex}
\end{table}

\begin{table}[H]
\centering
{\footnotesize
\begin{tabular}{lrrrr}
\hline
Model         & Accuracy & Number of Parameters & \rule{30pt}{0pt}AIC    & \rule{30pt}{0pt}BIC      \\ \hline
Naive-Bayes   & 0.73     & 232                & 3,142 & 3,356   \\
Decision Tree & 0.78     & 8,932              & 4,512 & 8,567   \\
Random Forest & \textbf{0.80}     & 45,341             & 9,231 & 18,123 \\
EVOTER        & 0.79     & \textbf{97}         & \textbf{1,578}  & \textbf{1,892}    \\ \hline
              &          &                      &        &         \\
\end{tabular}}
\vspace*{-2ex}
\caption{\emph{Model Comparison for Contraceptive Method Choice.}}
\label{tab:model_comparison_contraceptive_choice}
\vspace*{-4ex}
\end{table}

\section{Illustrating EVOTER's Performance}
\label{sc:experiments}

The experiments in this section each illustrate a different aspect of EVOTER's performance.  In Section~\ref{sc:prediction}, EVOTER's ability to predict/classify time series is illustrated again, but now through its extension to evolving time-lags for features.  This ability is then extended to prescriptions/policy search in Section~\ref{sc:prescription}, where the value of its two other extensions, i.e.\ comparing features with each other and evolving exponents are demonstrated. In Section~\ref{sc:esp}, similar prescriptions are shown to emerge when evaluation is done directly and through a surrogate model. Section~\ref{sc:insight} demonstrates that EVOTER results in explainability that is understandable to humans: The evolved rules are meaningful to domain experts and, in some cases, provide useful insight to them. Finally, Section~\ref{sc:nn} shows that the prescription performance of evolved rule sets is similar to the performance of evolved neural networks, and that the sets can be compressed and executed efficiently.

\subsection{Rule Sets for Prediction/Classification}
\label{sc:prediction}

\begin{figure}[!t]
\centering
\footnotesize
\setlength{\lineskip}{-1pt}
\begin{align*}
1. &(\textit{Mean}[4] < 72.75\text{mmHg}) \And \\
&(\textit{Kurtosis}[3] < 4.09) &&\longrightarrow \textit{Low}\\
2. &(\textit{Skew}[10] > 2.01) \And \\
&(\textit{Mean}[8] < 88.92\text{mmHg}) \And \\
&(\textit{Skew}[4] < 0.15) &&\longrightarrow Normal\\
3. &(\textit{Mean}[0] < 72.75\text{mmHg}) &&\longrightarrow \textit{Low}\\
4. &(\textit{Mean}[10] < 73.10\text{mmHg}) &&\longrightarrow \textit{Low}\\
5. &(\textit{Mean}[1] < 121.96\text{mmHg}) \And \\
&(\textit{Mean}[4] > 88.92\text{mmHg}) \And \\
&(\textit{Mean}[1] > 73.10\text{mmHg}) &&\longrightarrow \textit{High}\\
6. &(\textit{Mean}[0] < 97.53\text{mmHg}) &&\longrightarrow \textit{Normal}\\
7. &(\textit{Mean}[0] < 97.53\text{mmHg}) \And \\
&(\textit{Kurtosis}[0] > 12.71) &&\longrightarrow \textit{Normal}\\
8. &(\textit{Mean}[4] < 72.75\text{mmHg}) \And \\
&(\textit{Kurtosis}[7] > 4.03) &&\longrightarrow \textit{Low}\\
9. &(\textit{Mean}[4] > 121.96\text{mmHg}) \And \\
&(\textit{Kurtosis}[5] > 12.71) \And \\
&(\textit{Kurtosis}[3] > 1.00) &&\longrightarrow \textit{Normal}\\
10. &(\textit{Std}[0] < 10.76) &&\longrightarrow \textit{High}\\
11. &(\textit{Kurtosis}[0] > 1.00) &&\longrightarrow \textit{High}\\
12. &(\textit{Mean}[0] < 72.75\text{mmHg}) \And \\
&(\textit{Std}[4] > 0.01) &&\longrightarrow \textit{Low}\\
13. &(\textit{Kurtosis}[0] < 4.09) \And \\
&(\textit{Skew}[3] > 2.01) &&\longrightarrow \textit{Normal}\\
14. &(\textit{Skew}[9] > 0.06) &&\longrightarrow \textit{High}\\
15. &(\textit{Skew}[0] < 1.95) &&\longrightarrow \textit{High}\\
16. &(\textit{Mean}[0] < 72.75\text{mmHg}) \And \\
&(\textit{Mean}[5] < 52.12\text{mmHg}) &&\longrightarrow \textit{Low}\\
17. &~\text{Default} &&\longrightarrow \textit{Normal}\\
\end{align*}
\vspace*{-8ex}
\caption{\emph{A sample rule set evolved to predict blood pressure.} 
\label{fig:bp_ref}
All the features are extracted from aggregations of MAP \cite{bp-sepsis}. EVOTER discovered sets of features at specific time points to provide a useful signal for prediction. For instance, Std[4] specifies the standard deviation of the aggregated mean arterial pressure (MAP) over four minutes earlier.  The evolved rules predict sepsis accurately and are interpretable and meaningful to experts.}
\end{figure}

In this experiment, EVOTER was again applied to a time series prediction/classification task: Based on a series of blood pressure measurements over time, predict whether a patient in the ICU will go into septic shock in the next 30 minutes. However, instead of choosing among fixed features at different time points (i.e.\ features $x_{t_1}, x_{t_2}..x_{t_n}$), EVOTER evolved time lags for the features directly (i.e.\ lags $t$ of $x[t]$).  The experiment thus demonstrated that EVOTER's time-lag evolution extension is effective in handling temporal dependencies, and showed how its explainability can provide valuable insight for decision-making in critical care.

\paragraph{Experiment}
Consider the mean arterial pressure (MAP) signal at time $t$ as $x(t) \in \mathbb{R}$. The goal is to predict if the value of the statistic $\Bar{m} = \mathrm{avg}([x(t + \alpha), ... , x(t+\alpha +\beta)])$ falls into one of the three intervals:
\begin{itemize}
\item If $\Bar{m} \le 55\text{mmHg} \longrightarrow $ Low;
\item If $\text{55mmHg} < \Bar{m} \le 85\text{mmHg} \longrightarrow $ Normal;
\item If $\Bar{m} > 85\text{mmHg} \longrightarrow $ High.
\end{itemize}
The first of these intervals indicates acute hypotension indicative of septic shock. In this sense, the task is a prediction/classification task.

\paragraph{Results} EVOTER was trained with approximately 4000 patients' arterial blood pressure waveforms from the MIMIC II v3 data set \cite{bp-sepsis}. Rules were evolved to observe various features of these waveforms at various times and predict whether acute hypotension was likely to develop within the prediction window of 30 minutes (Figure~\ref{fig:bp_ref}). In a massively parallel implementation on 1000 clients running for several days, EVOTER achieved an accuracy of 0.895 risk-weighted error on the withheld set, with a true positive rate of 0.96 and a false positive rate of 0.394.

Note that the approach is general and does not need a transformation of the time-series data into large temporal feature sets. Even though many machine learning methods could be used for this prediction task, EVOTER provides a solution that is transparent and interpretable. Indeed, the rules were evaluated by emergency-room physicians who found them meaningful \cite{bp-sepsis}. Without such an understanding and verification, it would be difficult to trust the system enough to deploy it. Thus, working in tandem with human experts, EVOTER can be useful in interpreting and understanding the progression of the patient's health and sensitivity and serve as an early indicator of problems.

\subsection{Rule Sets for Prescription/Policy Search}
\label{sc:prescription}

The second set of experiments focuses on problems where the goal is not to predict a set of known labels but instead to prescribe, i.e.\ generate actions for an agent in a reinforcement learning environment. Results from two standard such domains are included: flappy bird and cart-pole. In both domains, evaluation is done by observing the effects of the actions directly in a simulation of the domain.

\subsubsection{Flappy Bird}
\label{sc:flappybird}
\paragraph{Experiment}
Flappy Bird is a side-scroller game where the player attempts to fly an agent between columns of pipes without hitting them by performing flapping actions at carefully chosen times. The experiment was based on a PyGame~\cite{tasfi2016PLE} implementation running at a speed of 30 frames per second. The goal of the game is to finish ten episodes of two minutes, or 3,600 frames each, through random courses of pipes. A reward is given for each frame where the bird does not collide with the boundaries or the pipes; otherwise, the episode ends. The score of each candidate is the average reward over the ten episodes \cite{esp-rl}.

\begin{figure}[!t]
\centering
\footnotesize
\setlength{\lineskip}{-1pt}
\begin{align*}
1.&(0.99*\textit{next.pipe.dist.to.player} < 0.93*\textit{next.next.pipe.bottom.y}) \And\\
&(0.99*\textit{next.pipe.dist.to.player} < 0.83*\textit{next.next.pipe.bottom.y}) \And\\
&(0.98*\textit{player.y} \le 0.78*\textit{next.pipe.bottom.y}) \And\\
&(0.95*\textit{player.y} \le 0.65*\textit{next.pipe.bottom.y}) \And\\
&(0.76*\textit{player.vel} > -0.98 [-8.0..10.0]) \And\\
&(0.47*\textit{next.next.pipe.bottom.y} > 0.82*\textit{player.vel}) \And\\
&(0.41*\textit{player.y} \le 0.78*\textit{next.pipe.bottom.y}) \And\\
&(0.26*\textit{next.pipe.top.y} < 0.76*\textit{player.y}) \And\\
&(0.17*\textit{next.pipe.top.y} \le 84.48 [0..192.0]) &&\hspace*{-3ex}\longrightarrow \textit{Flap}\\
2.&(0.95*\textit{player.y} \le 0.65*\textit{next.pipe.bottom.y}) \And\\
&(0.76*\textit{player.vel} > -0.98 [-8.0..10.0]) \And\\
&(0.47*\textit{next.next.pipe.bottom.y} > 0.82*\textit{player.vel}) \And\\
&(0.41*\textit{player.y} \le 0.78*\textit{next.pipe.bottom.y}) \And\\
&(0.19*\textit{next.pipe.dist.to.player} < 0.64*\textit{next.pipe.bottom.y}) \And\\
&(0.17*\textit{next.pipe.top.y} \le 84.48 [0..192.0]) &&\hspace*{-3ex}\longrightarrow \textit{Flap}\\
3. &(0.92*\textit{next.pipe.dist.to.player} < 0.95*\textit{next.pipe.top.y}) \And\\
&(0.78*\textit{next.pipe.bottom.y} \ge 175.2 [0..292.0]) \And\\
&(0.71*\textit{next.next.pipe.bottom.y} > 0.71*\textit{next.pipe.dist.to.player}) \And\\
&(0.49*\textit{next.next.pipe.top.y} \ge 0.12*\textit{next.pipe.dist.to.player}) \And\\
&(0.53*\textit{next.pipe.top.y} < 0.63*\textit{next.pipe.dist.to.player}) &&\hspace*{-3ex}\longrightarrow \textit{No flap}\\
4. &~\text{Default Action} &&\hspace*{-3ex}\longrightarrow \textit{No flap}\\
\end{align*}
\vspace*{-8ex}
\caption{\emph{A sample solution rule set discovered for the Flappy Bird domain by direct evolution.} The rules primarily identify situations where a flap is warranted. Rule 3 is redundant, and the conditions in all rules have redundancy, which is common in evolved solutions, making the search robust.}\label{fig:flappy-de}
\end{figure}

\paragraph{Results} EVOTER finds perfect solutions to this problem typically within 400 generations. Figure~\ref{fig:flappy-de} shows a sample solution rule set. As is typical in this domain, the rules identify cases where the agent should flap. The conditions appear complex, however, several of the clauses are redundant and can be removed to form a final solution (as will be demonstrated in Section~\ref{sc:hf}). Such redundancy is common in evolved solutions, likely because it makes the solutions robust to mutations and crossover. Innovation is unlikely to break them completely, and they can be refined to be useful. While this principle applies to many evolutionary settings (e.g.\ neuroevolution \cite{gomez}), rule sets make it explicit.

\subsubsection{Cart-Pole}
\label{sc:cartpole}
\paragraph{Experiment}
Cart-pole is one of the standard reinforcement learning benchmarks. In the popular CartPole-v0 implementation in the OpenAI Gym platform \cite{brockman2016gym}, there is a single pole on a cart that moves left and right depending on the force applied to it.  The controller inputs are the position of the cart, the velocity of the cart, the angle of the pole, and the rotation rate of the pole. A reward is given for each time step that the pole stays near vertical, and the cart stays near the center of the track; otherwise, the episode ends \cite{esp-rl}.

\paragraph{Results} EVOTER finds solutions reliably within five generations. These rule sets will keep the pole from falling indefinitely when started with situations in the standard validation set. Figure~\ref{fig:cartpole-de} shows a sample of such a rule set. In contrast to the usual rule-set solutions that identify different situations and develop rules for each of them, this set is remarkably simple. It is based on a relationship between pole angle and cart velocity that applies across situations. In this manner, EVOTER can discover and take advantage of physical relationships, which is, in general, a powerful ability of evolutionary optimization \cite{lipson}.

\begin{figure}[!t]
\centering
\footnotesize
\setlength{\lineskip}{-1pt}
\begin{alignat*}{2}
1. &(0.11*{\textit{velocity.of.cart}}^3 < 0.87*\textit{angle.of.pole}) &&\longrightarrow {\textcolor{teal}{\textbf{\textit{Left}}}}\\
2. &~\text{Default Action} &&\longrightarrow {\textcolor{red}{\textbf{\textit{Right}}}}\\
\end{alignat*}
\vspace*{-8ex}
\caption{\emph{A sample solution rule set for the Cart-Pole problem discovered by direct evolution.} This solution is remarkably simple: The first rule expresses a mathematical relationship between two input features, making the usual case-by-case rule approach unnecessary. Such discoveries can be useful in uncovering insights into the nature of the physical system.\label{fig:cartpole-de}}
\end{figure}

\begin{figure*}[!ht]
  \centering
  \begin{minipage}[t]{0.22\linewidth}
  \centering
  \includegraphics[width=\linewidth]{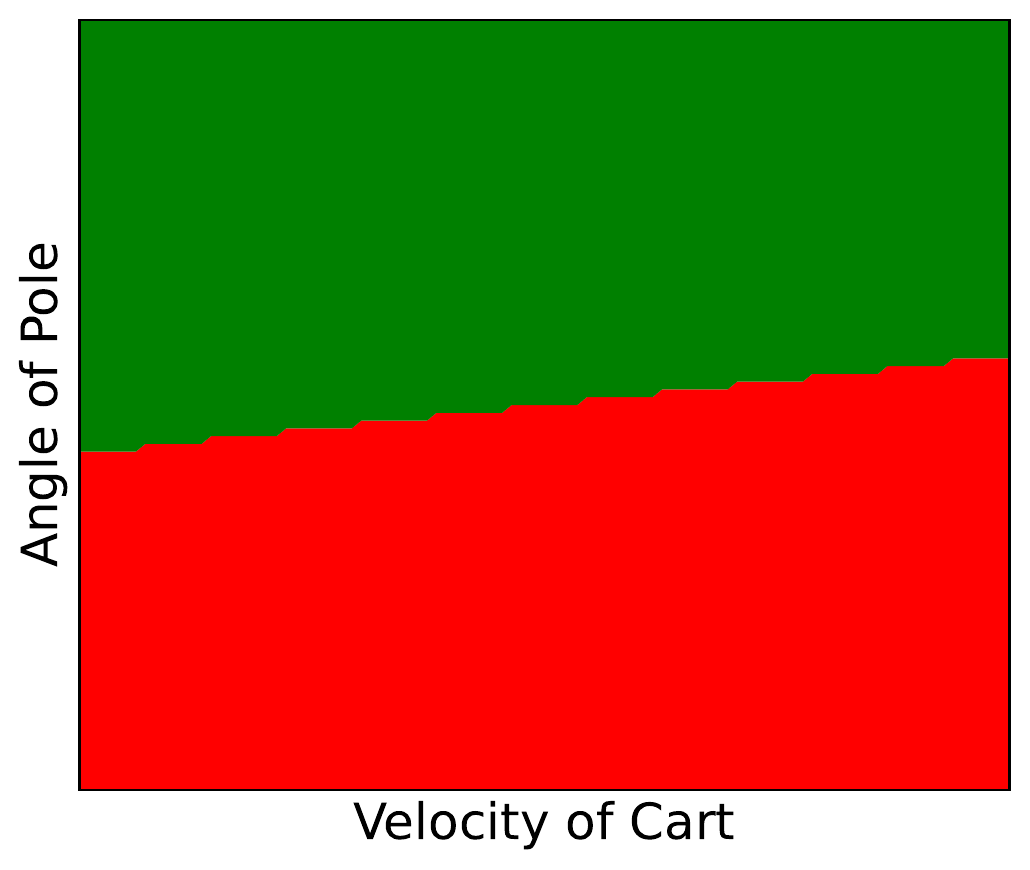}\\[-1ex]
  {\footnotesize ($a$) EVOTER}
  \end{minipage}
  \hfill
  \begin{minipage}[t]{0.22\linewidth}
  \centering
  \includegraphics[width=\linewidth]{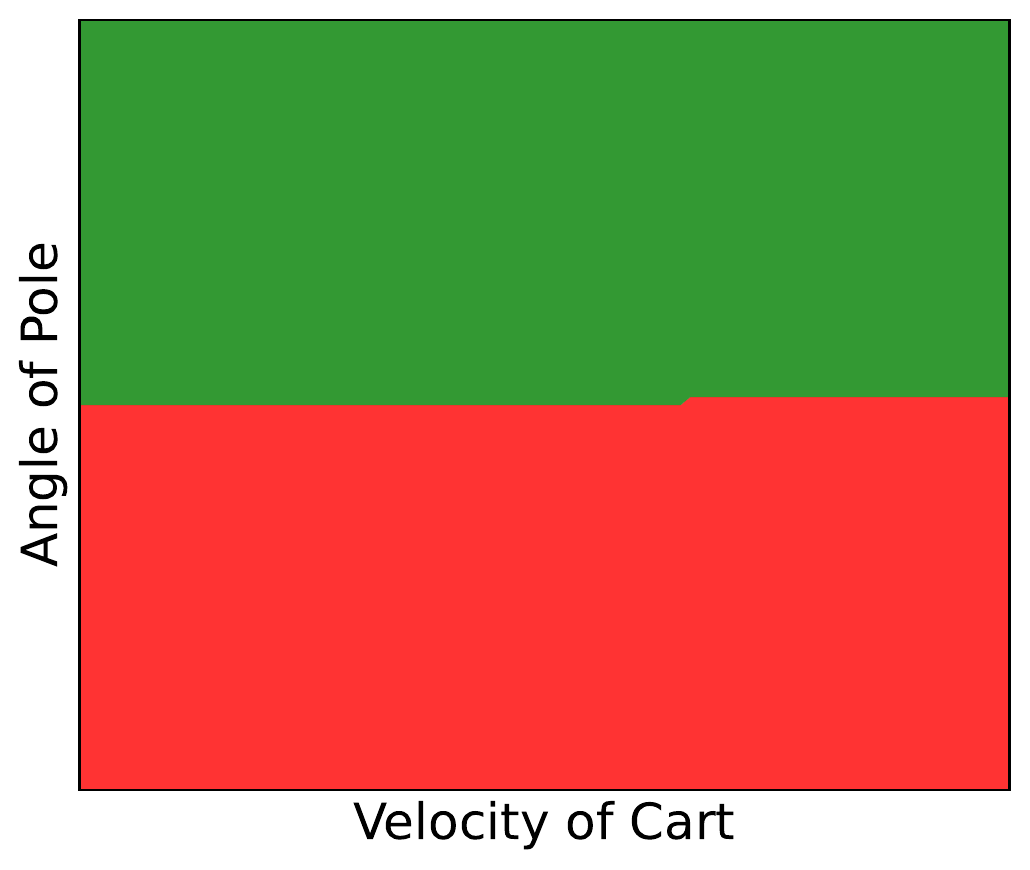}\\[-1ex]
  {\footnotesize ($b$) Naive Bayes}
  \end{minipage}
  \hfill
  \begin{minipage}[t]{0.22\linewidth}
  \centering
  \includegraphics[width=\linewidth]{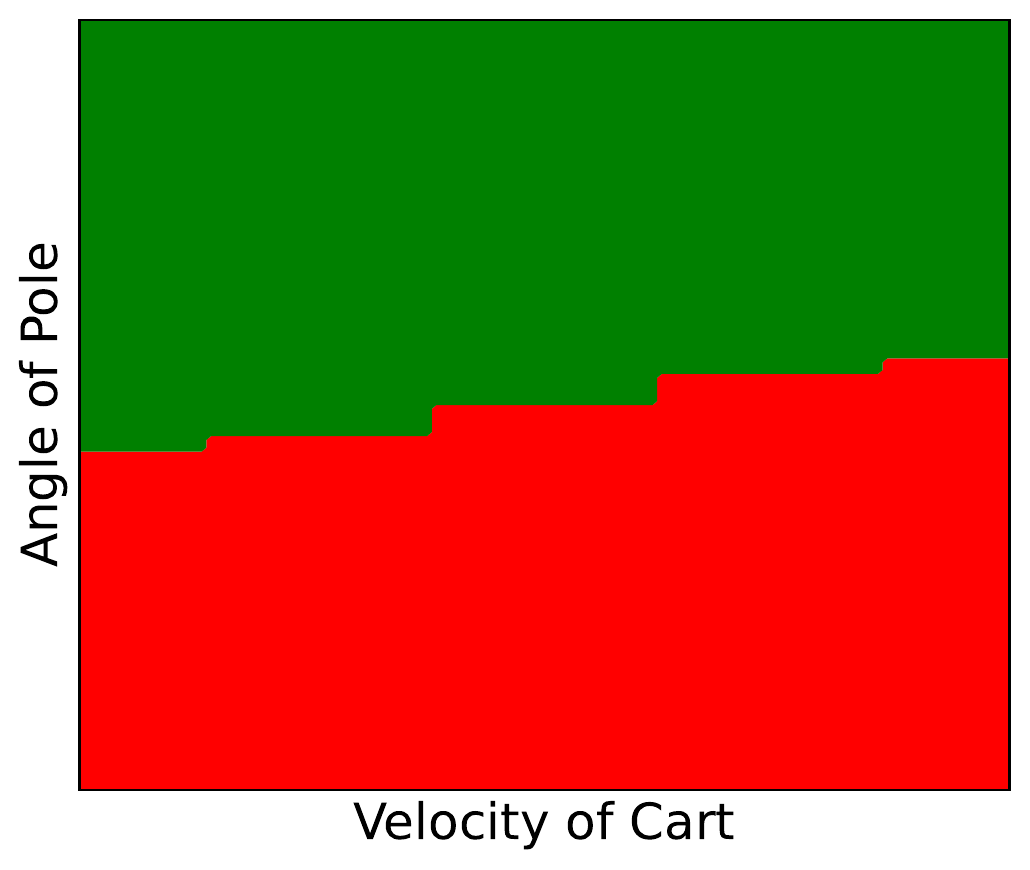}\\[-1ex]
  {\footnotesize ($c$) Decision tree}
  \end{minipage}
  \hfill
  \begin{minipage}[t]{0.22\linewidth}
  \centering
  \includegraphics[width=\linewidth]{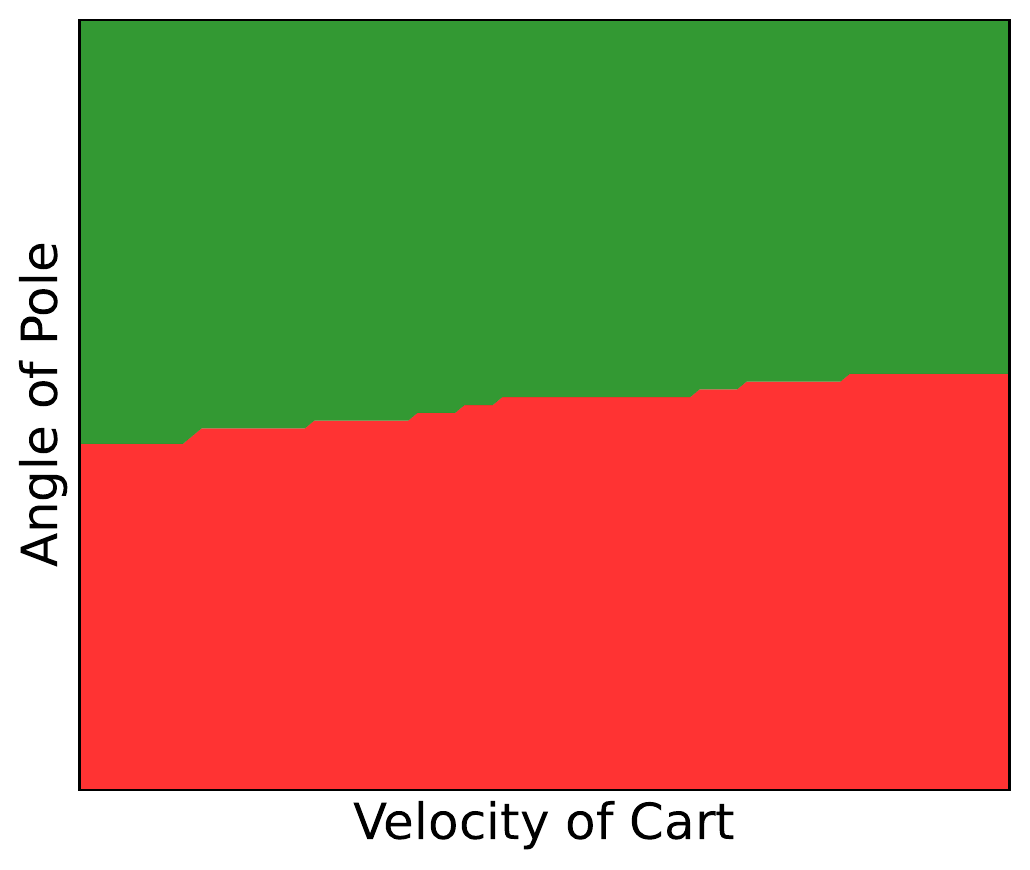}\\[-1ex]
  {\footnotesize ($d$) Random Forest}
  \end{minipage}
  \begin{minipage}[t]{0.062\linewidth}
  \centering
  \includegraphics[width=\linewidth]{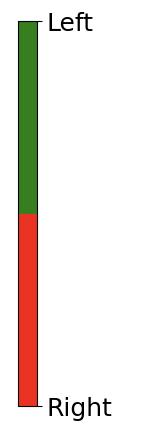}
  \end{minipage}\\
  \includegraphics[width=0.8\linewidth, height=125pt]{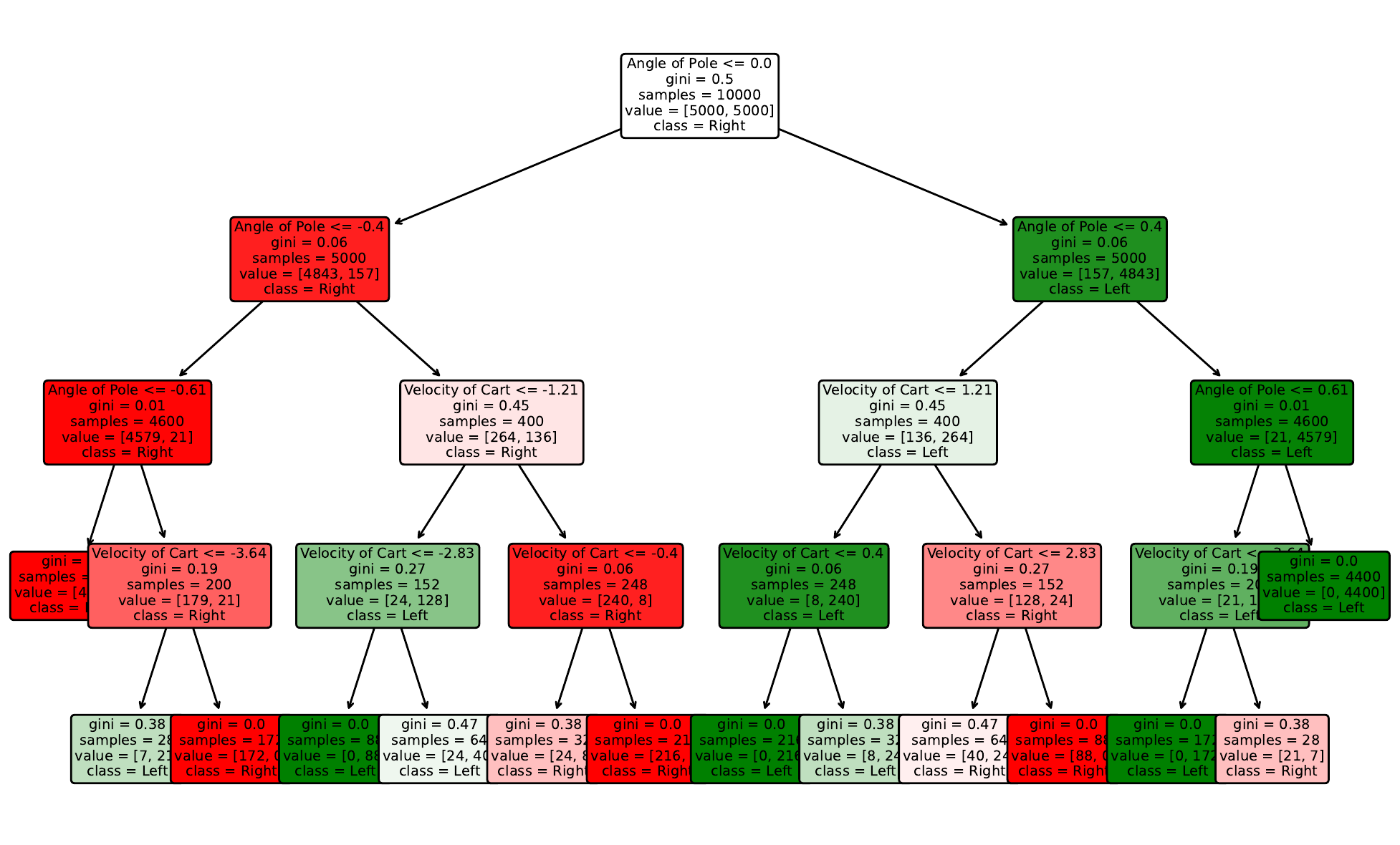}\\
  \vspace*{-2ex}
  {\footnotesize ($e$) The four-level decision tree approximating the linear rule as shown in ($c$)}
\vspace*{-1ex}
\caption{\emph{EVOTER versus other transparent models representing the linear rule} "IF $(0.11 * \textit{velocity.of.cart} < 0.87 * \textit{angle.of.pole})$ THEN \textit{Left}; ELSE \textit{Right}." ($a$) The decision boundary of this rule in EVOTER is a linear line. ($b$) The decision boundary of the naive-Bayes model largely misses the linear relationship and instead keys on the pole angle only. ($c$) The decision boundary of the decision tree with four levels struggles to approximate the relationship, even though it is quite complex, as shown in $e$ (with colors representing increased certainty from red to green). It would need many more levels to be as accurate as the simple rule in EVOTER. ($d$) The decision boundary of the random-forest model with ten estimators is better, but this model also has many more parameters. Thus, the other models are less accurate or more complex (i.e., have more parameters) than the explanation discovered by EVOTER through its feature comparison extension.}
\label{fig:feature}
\end{figure*}

The cart-pole domain can be used to illustrate the value of the second and third extensions of EVOTER, i.e.\ the ability to compare features and to exponentiate them. First, consider the discovered rule shown in Figure~\ref{fig:cartpole-de} but without the exponent, thus reducing it to a straightforward comparison between the two features. This rule establishes a linear decision boundary, as shown in Figure~\ref{fig:feature}$a$. Naive Bayes largely misses it, making decisions based on pole angle only (Figure~\ref{fig:feature}$b$). The decision tree captures this relationship, but struggles to represent it accurately (Figure~\ref{fig:feature}$c$, even with the relatively complex model consisting of four levels (shown in Figure~\ref{fig:feature}$e$). In effect, it can only do it by adding many decision nodes, like the random forest does  Figure~\ref{fig:feature}$e$). This illustration emphasizes an important point: Performance of these other transparent models comes with a significant cost in complexity, making them less explainable. Section~\ref{sc:quantitative} demonstrated this conclusion quantitatively; the visualization of the decision tree in Figure~\ref{fig:feature}$e$ makes it concrete: it is much more complex and difficult to understand that the simple linear rule it approximates, i.e.\ "IF $(0.11 * \textit{velocity.of.cart} < 0.87 * \textit{angle.of.pole})$ THEN \textit{Left}; ELSE \textit{Right}."

\begin{figure*}[!ht]
  \centering
  \begin{minipage}[t]{0.22\linewidth}
  \centering
  \includegraphics[width=\linewidth]{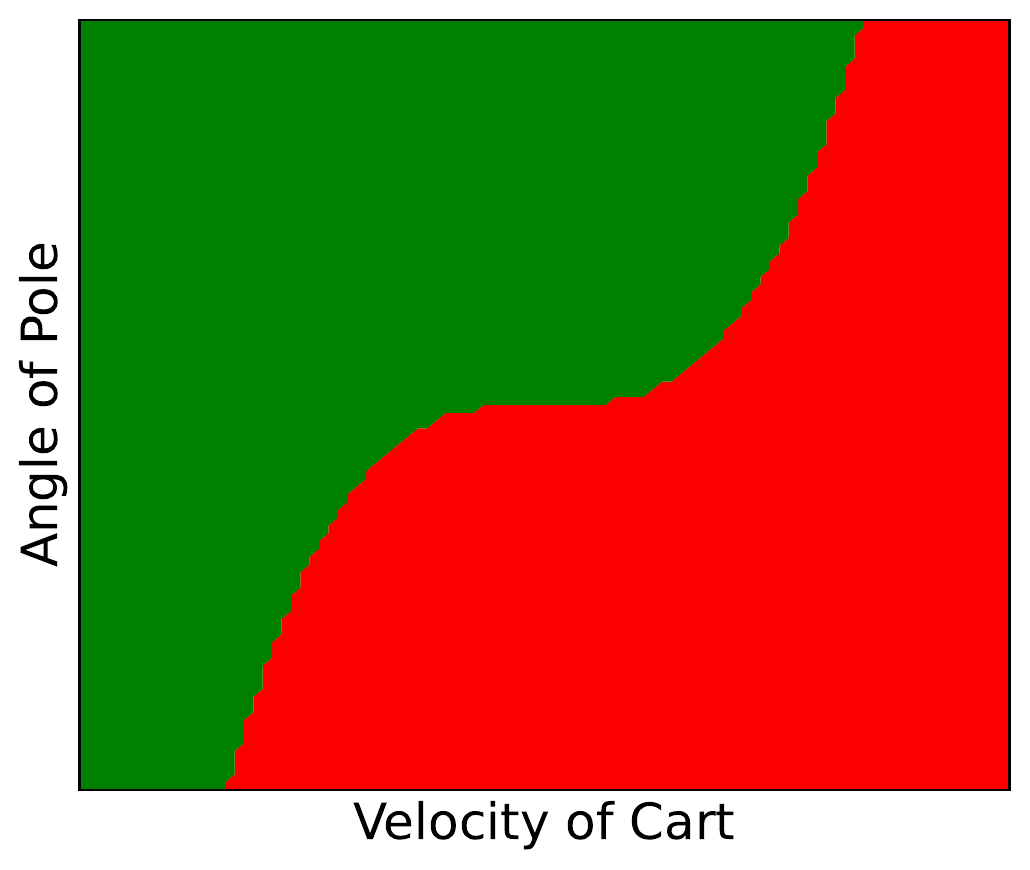}\\[-1ex]
  {\footnotesize ($a$) EVOTER}
  \end{minipage}
  \hfill
  \begin{minipage}[t]{0.22\linewidth}
  \centering
  \includegraphics[width=\linewidth]{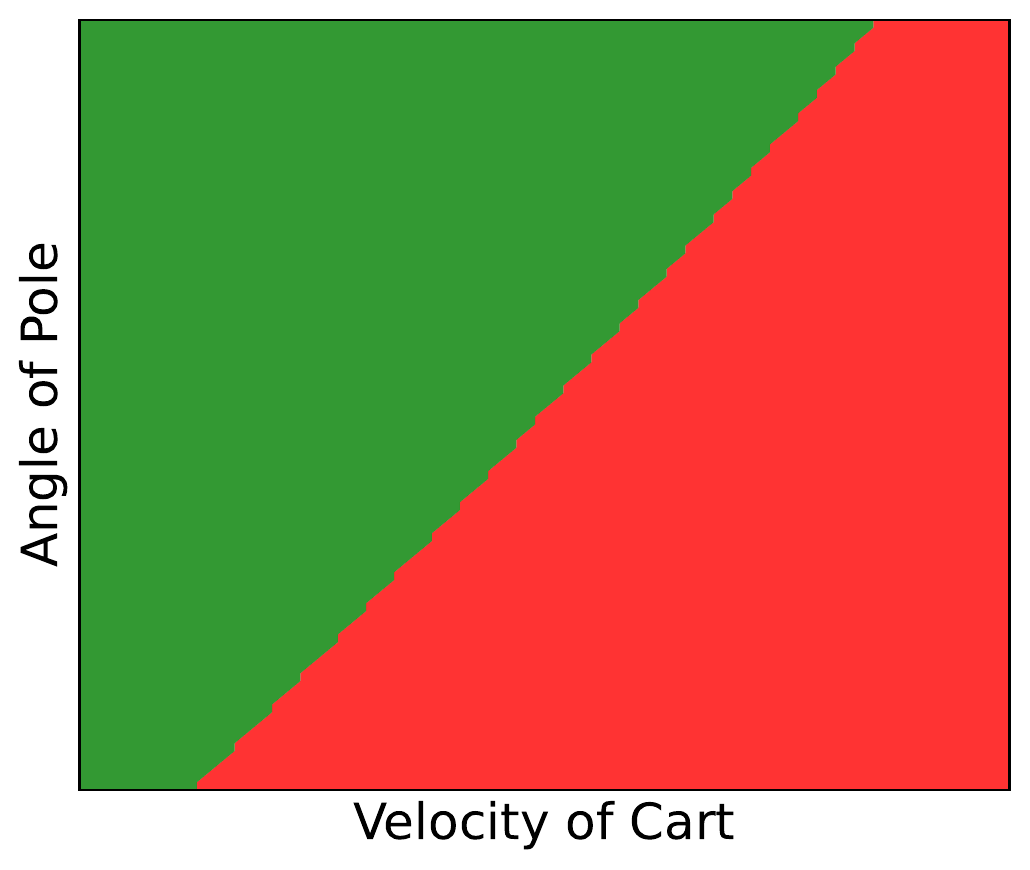}\\[-1ex]
  {\footnotesize ($b$) Naive Bayes}
  \end{minipage}
  \hfill
  \begin{minipage}[t]{0.22\linewidth}
  \centering
  \includegraphics[width=\linewidth]{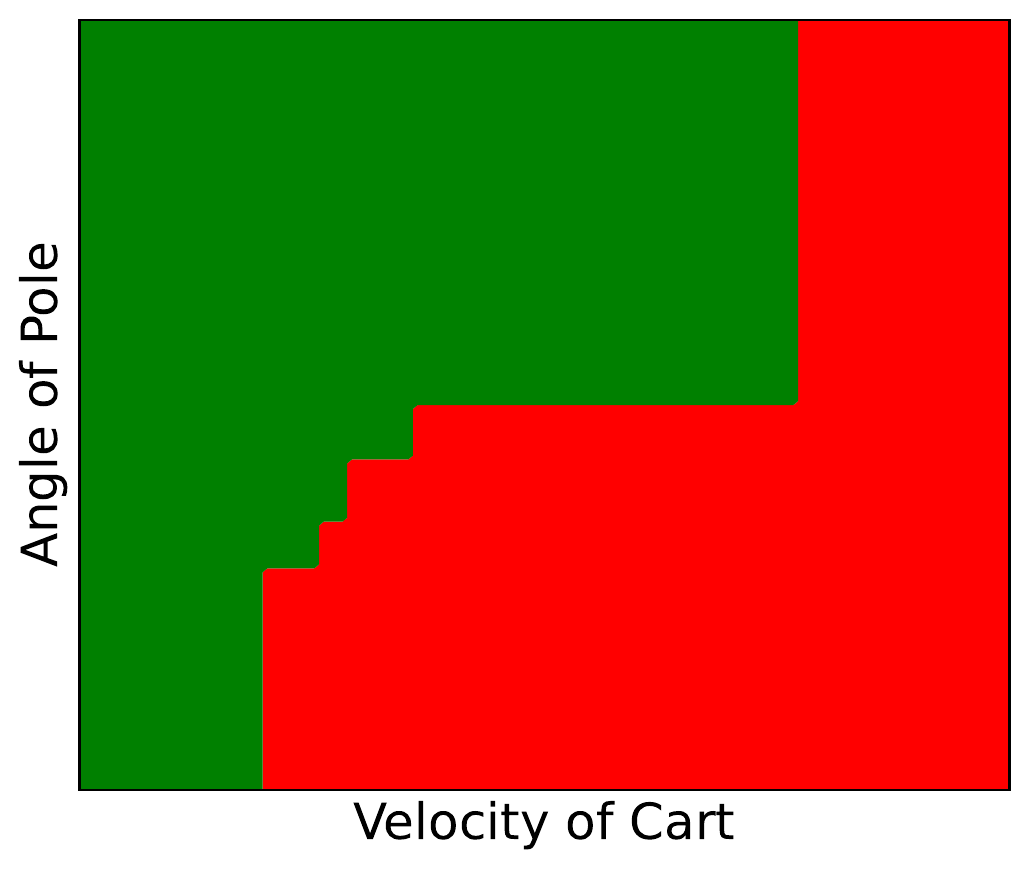}\\[-1ex]
  {\footnotesize ($c$) Decision tree}
  \end{minipage}
  \hfill
  \begin{minipage}[t]{0.22\linewidth}
  \centering
  \includegraphics[width=\linewidth]{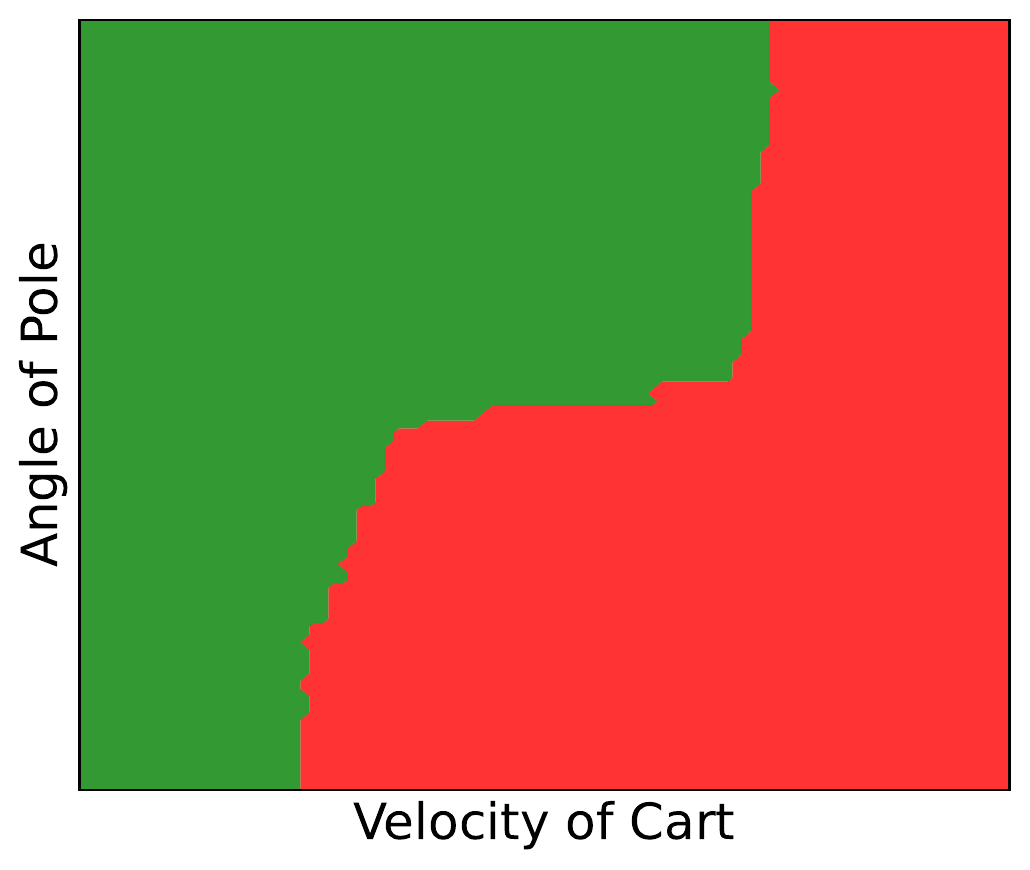}\\[-1ex]
  {\footnotesize ($d$) Random Forest}
  \end{minipage}
  \hfill
  \begin{minipage}[t]{0.062\linewidth}
  \centering
  \includegraphics[width=\linewidth]{images/Color_bar.png}
  \end{minipage}\\
  \includegraphics[width=0.8\linewidth, height=125pt]{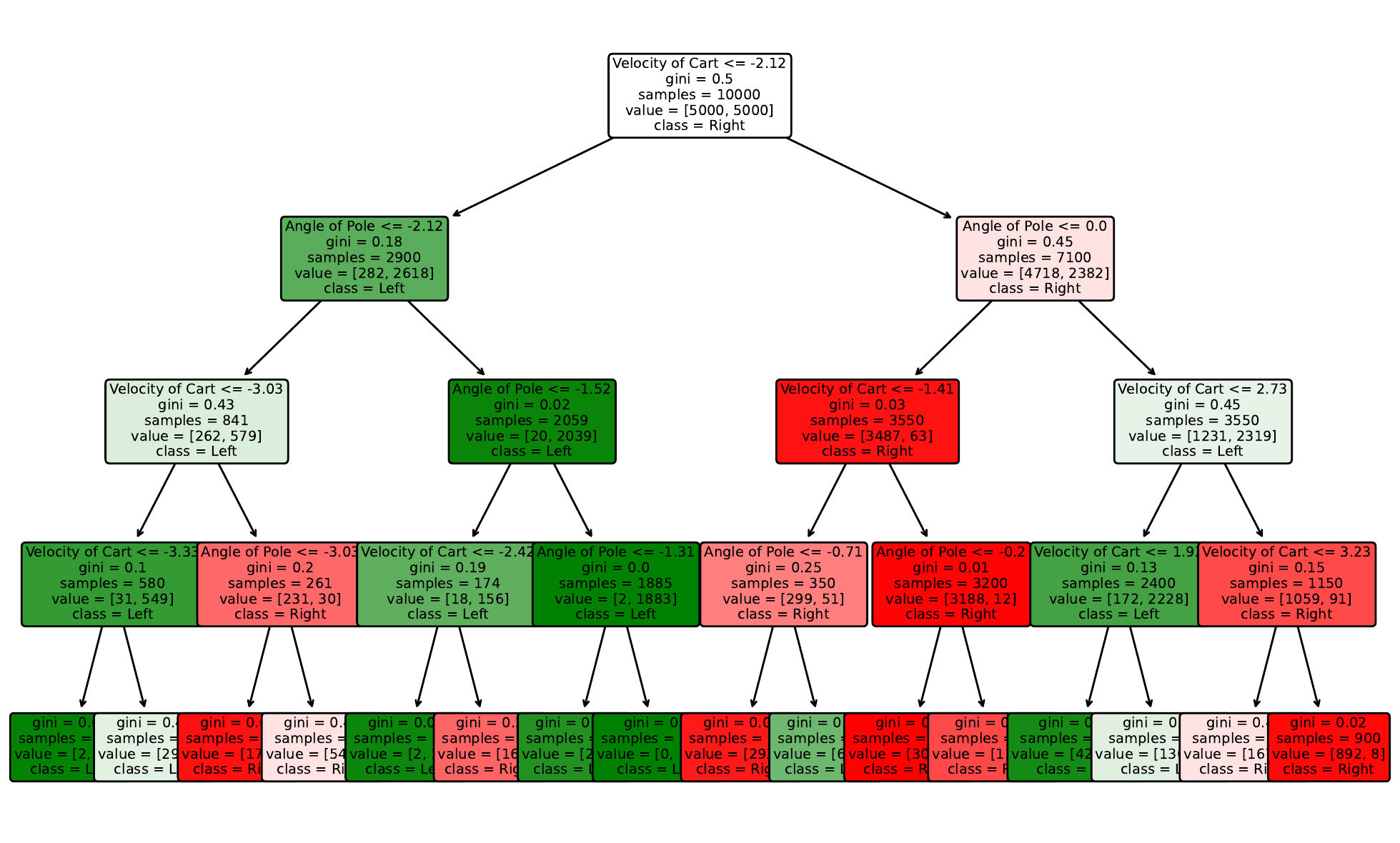}\\
  \vspace*{-2ex}
  {\footnotesize ($e$) The four-level decision tree approximating the nonlinear rule as shown in ($c$)}
\vspace*{-1ex}
\caption{\emph{EVOTER versus other transparent models representing the nonlinear rule "If $(0.11 * \textit{velocity.of.cart}^3 < 0.87 * \textit{angle.of.pole})$ then left; otherwise right."} ($a$) The decision boundary of this rule in EVOTER is smoothly nonlinear; ($b$) The decision boundary of the naive-Bayes model is a straight line. ($c$) The decision boundary of the decision tree with four levels (shown in $e$) consists of several blocks, and ($d$) the decision boundary of the random-forest model with ten estimators is not much better. Both of the tree models would need more parameters to be accurate. Thus, EVOTER's nonlinearity extension to the rule representation not only offers simplicity and ease of understanding but also maintains precision and accuracy.}
\label{fig:exponent}
\end{figure*}

Second, consider the actual nonlinear rule shown in Figure~\ref{fig:cartpole-de}. The decision boundary is now nonlinear, as seen in  Figure~\ref{fig:exponent}$a$. Naive Bayes approximates it with a straight line and the four-level deep decision tree with a series of blocks. Again, with more levels and parameters, as in the random forest, the approximation becomes better, with the cost of increased complexity and decreased explainability.  Since many physical systems are nonlinear, the ability to represent exponents is crucial in uncovering such relationships. For instance, consider Kepler's Third Law stating that the squares of the orbital periods of planets are directly proportional to the cubes of the semi-major axes of their orbits. Like many similar laws of nature, this law was initially discovered as an underlying principle of numerous observations. Yet linear models such as decision trees cannot learn it: It is a power law that requires the representation of exponents. This level of complexity exceeds the capabilities of the other transparent systems, highlighting the importance of the nonlinearity extension in EVOTER in developing insights in scientific domains.

\subsection{Evolving Rule Sets with a Surrogate}
\label{sc:esp}

Many real-world domains are too costly for direct evolution, and the evaluation of candidates needs to be done more economically against a surrogate model \cite{esp-rl,miikkulainen:ieeetec21}. The next experiment demonstrates that solutions evolved with a surrogate can be remarkably similar to those evolved directly in the domain.

\paragraph{Experiment} ESP (Evolutionary Surrogate-assisted Prescription; \cite{esp-rl}) is a general approach to decision making where the decision policy is discovered through evolution and the surrogate is constructed through machine learning.
The surrogate, or the predictor, can be an opaque model such as a random forest or neural network trained with gradient descent. Similarly, the decision policy, or the prescriptor, is often a neural network; however, evolving a rule set to represent the strategy results in an explainable solution instead.

In this experiment, ESP was set up to solve the same cart-pole domain as in the previous section. The prescriptor population was initially random and therefore generated a diverse set of actions. The predictor was trained with the outcome of these actions. The trained predictor was then used to evaluate prescriptor candidates through evolution.

\begin{figure}[!t]
\centering
\footnotesize
\setlength{\lineskip}{-1pt}
\begin{alignat*}{2}
1. &(0.16*{\textit{velocity.of.cart}}^3 > 0.89*\textit{angle.of.pole}) &&\longrightarrow {\textcolor{red}{\textbf{\textit{Right}}}}\\
2. &~\text{Default Action} &&\longrightarrow {\textcolor{teal}{\textbf{\textit{Left}}}}\\
\end{alignat*}
\vspace*{-8ex}
\caption{\emph{A sample rule set discovered through ESP in the Cart-Pole domain.} These rules are logically almost identical to those discovered through direct evolution (i.e., Figure~\ref{fig:cartpole-de}). This result verifies that a surrogate-based optimization is a viable approach, i.e.\ it discovers the same solutions as direct evolution.\label{fig:cartpole-esp}}
\end{figure}

\paragraph{Results} The predictor trained reliably with 100 action examples from the random initial population. Evolution then took only three generations to find agents that solved the examples in the validation set.
Interestingly, the resulting rules (Figure~\ref{fig:cartpole-esp}) are logically almost identical to those resulting from direct evolution (Figure~\ref{fig:cartpole-de}). While this validity of the surrogate modeling approach has been demonstrated before in terms of performance \cite{esp-rl}, rule-set evolution makes this conclusion explicit.

\subsection{Obtaining Insights and Avoiding Biases}
\label{sc:insight}

In this experiment, rule-based ESP was applied to the task of recommending medication for hospitalized diabetes patients that would both minimize hospital readmissions and lead to good discharge outcomes. The results were evaluated together with physicians, giving them a plain-text interpretation and indicating insights that are possible to obtain with this approach.

\paragraph{Experiment}
The Diabetes data set \cite{diabetes} represents ten years (1999-2008) of clinical care at 130 US hospitals and integrated delivery networks. Information was extracted from the database for encounters that satisfied the following criteria:
\begin{itemize}
    \item It is an inpatient encounter (a hospital admission).
    \item It is a diabetic encounter, i.e., one with any kind of diabetes was entered into the system as a diagnosis.
    \item The length of stay was at least one day and at most 14 days.
    \item Laboratory tests were performed during the encounter.
    \item Medications were administered during the encounter.
\end{itemize}
The data set includes over 50 features characterizing the patient and the hospital context, including patient number, race, gender, age, admission type, time in hospital, the medical specialty of admitting physician, number of lab tests performed, HbA1c test results, and diagnosis, as well as the number of outpatients, inpatients, and emergency visits by the patient in the year before the hospitalization. As actions, 21 different treatments can be prescribed, i.e., \ different diabetic and other medications. Two objectives are optimized: readmission rate and discharge disposition. The three possible readmission categories are: "no readmission" (+1 point), "readmission in less than 30 days" (0), and "readmission in more than 30 days" (-1). The discharge-disposition categories were: "sent back home" (+2 points), "remained in the hospital" (-1), "left with advised medical attention" (-1), "sent to another hospital" (-1), and "died" (-4). The goal is to maximize the readmission and discharge-disposition scores simultaneously.

A neural-network predictor was trained with $\sim$66k samples in the historical data, with $\sim$33k reserved for validation and $\sim$33k for testing. A population of 100 rule sets was then evolved for 40 generations to generate a Pareto front: Given the patient and hospital contexts, they represent different tradeoffs between reducing readmissions and improving discharge dispositions. Note that these objectives are somewhat at odds: e.g., \ sending the patient home too early may result in early readmission.

\begin{figure}[!t]
\centering
\footnotesize
\setlength{\lineskip}{-1pt}
\begin{alignat*}{2}
1. &(\textit{diag1.injury} \le 2.2 * \textit{diag1.respiratory}) \And \\
&(\textit{diag2.neoplasms} > 33) &&\longrightarrow \textit{Metformin-Pioglitazone}\\
2. &(\textit{age}[60-70) \ge 1.2*\textit{age}[10-20)) &&\longrightarrow \textit{Glyburide}\\
3. &(\textit{diag2.diabetes} \ge 0.82*\textit{diag1.digestive}) &&\longrightarrow \textit{Glipizide-
Metformin}\\
4. &(\textit{admission.Court/Law} \ge 1.045*\textit{diag3.circulatory}) &&\longrightarrow \textit{Pioglitazone}\\
5. &(\textit{age}[80-90) \ge 2.29*\textit{admission.type.id.Newborn}) &&\longrightarrow \textit{Tolazamide}\\
6. &(\textit{age}[30-40) \le 0.47*\textit{age}[50-60)) \And\\
&(\textit{admission.type.id.Newborn} \le 0.47*\textit{age}[50-60)) \And\\
&(\textit{diag2.respiratory} \ge 1.86*\textit{diag3.musculoskeletal}) \And\\
&(\textit{diag1.diabetes} \ge 1.42*\textit{age}[30-40)) \And\\
&(0.18*\textit{gender} \ge 0.38*\textit{race.Asian}) \And\\
&(0.12*\textit{race.AfricanAmerican} \ge 0.15*\textit{admission.Court/Law}) \And\\
&(0.03*\textit{race.Hispanic} \le 0.76*\textit{admission.Emergency.Room}) &&\longrightarrow \textit{Insulin}\\
7. &~\text{Default} &&\longrightarrow \textit{Glipizide}\\
\end{alignat*}
\vspace*{-8ex}
\caption{\emph{A sample rule set discovered through ESP to reduce hospital readmissions and improve discharge disposition for diabetes.} The rules are transparent but contain redundancies that make them sometimes hard to read. A domain expert can verify that the set is meaningful and translate them into plain English where the insights and biases are easier to see (Figure~\ref{fig:diabetes-2}).\label{fig:diabetes-1}}
\end{figure}

\paragraph{Results}
The predictor training achieved a test accuracy of $\sim$80\% (of correct classifications into the three categories). Rule set evolution started with a no-readmission rate of $\sim$29\%, which improved to $\sim$49\% over the 40 generations; a similar improvement was observed along the discharge-disposition objective. Remarkable tradeoffs were discovered that performed much better than the actual prescriptions in the data set. Whereas 78\% of patients were sent home and the no-readmission rate was 60\% in the data set, the most dominant solution recommended treatments that were predicted to result in sending home as well as no readmission 99\% of the time.

\begin{figure}[!ht]
\centering
\footnotesize
\setlength{\lineskip}{-1pt}
\begin{alignat*}{2}
1. &~\text{If the patient has a respiratory problem}\\
&~\text{but not a neoplasms problem} &&\text{prescribe Metformin-Pioglitazone}\\
2. &~\text{If the patient is any age except 60 to 70} &&\text{prescribe Glyburide}\\
3. &~\text{If the patient has Diabetic Diagnostic} &&\text{prescribe Glipizide-
Metformin}\\
4. &~\text{If the patient has circulatory diagnosis} &&\text{prescribe Pioglitazone}\\
5. &~\text{If the admission type is Newborn} &&\text{prescribe Tolazamide}\\
6. &~\text{If the patient's age is between 30 to 60} \And\\
&~\text{the patient has musculoskeletal diagnosis} \And\\
&~\text{is not Asian} \And\\
&~\text{was not sent by court/law} \And\\
&~\text{was admitted in emergency} &&\text{prescribe Insulin}\\
7. &~\text{If none of the above} &&\text{prescribe Glipizide}\\
\end{alignat*}
\vspace*{-8ex}
\caption{\emph{Expert interpretation of the diabetes treatment rule set.} In particular, Rule 6 is now easier to read. It also makes it possible to identify a potential racial bias; in this case, the bias is medically indicated, but in other applications, it may be due to inaccurate training data. Such biases can be identified and edited out when using rule sets, which would be difficult to do with other machine-learning approaches.}\label{fig:diabetes-2}
\end{figure}

Figure~\ref{fig:diabetes-1} shows this highly effective rule set discovered by EVOTER.
The rules are relatively transparent but again contain redundancies that make them somewhat difficult to read. To clarify and evaluate them, these rules were given to a domain expert who removed the redundancies and translated them into plain text (Figure~\ref{fig:diabetes-2}). For instance, given that their features take on binary values, the last three clauses of Rule 6 can be simplified to "not Asian", "not court/law", and emergency. The expert verified that, indeed, this set constitutes a meaningful treatment policy.

This example also illustrates the potential of the rule set evolution to make biases in machine learning systems explicit. For instance, rule 6 is conditioned upon the race of the patients being Asian. In this case, race is indeed a medically valid consideration, but it is easy to see that in other domains with incomplete data, such constraints could be simply due to biases in the data. With rule sets, it is possible to identify such biases. Furthermore, it is possible to edit them out when they are not warranted, which would be very difficult to do with other machine-learning approaches.

\subsection{Performance of Evolved Rule Set vs.\ Neural-Network Prescriptors}
\label{sc:nn}

The standard approach in ESP is to evolve neural networks as the prescriptor. While neural networks are powerful and flexible, they are opaque. On the other hand, rule sets are transparent and interpretable. An essential question is whether rule sets can achieve performance comparable to neural networks in this role. In this section, this question is answered in two different tasks, one (heart-failure prevention) involving a single objective and the other (diabetes treatment) two objectives. In both cases, the performance of rule sets is found only slightly short of that of the neural networks, demonstrating that explainability can be achieved with a minimal cost.

\begin{figure}[!t]
\begin{minipage}[b]{\linewidth}
\centering
\centering
\includegraphics[width=0.48\textwidth]{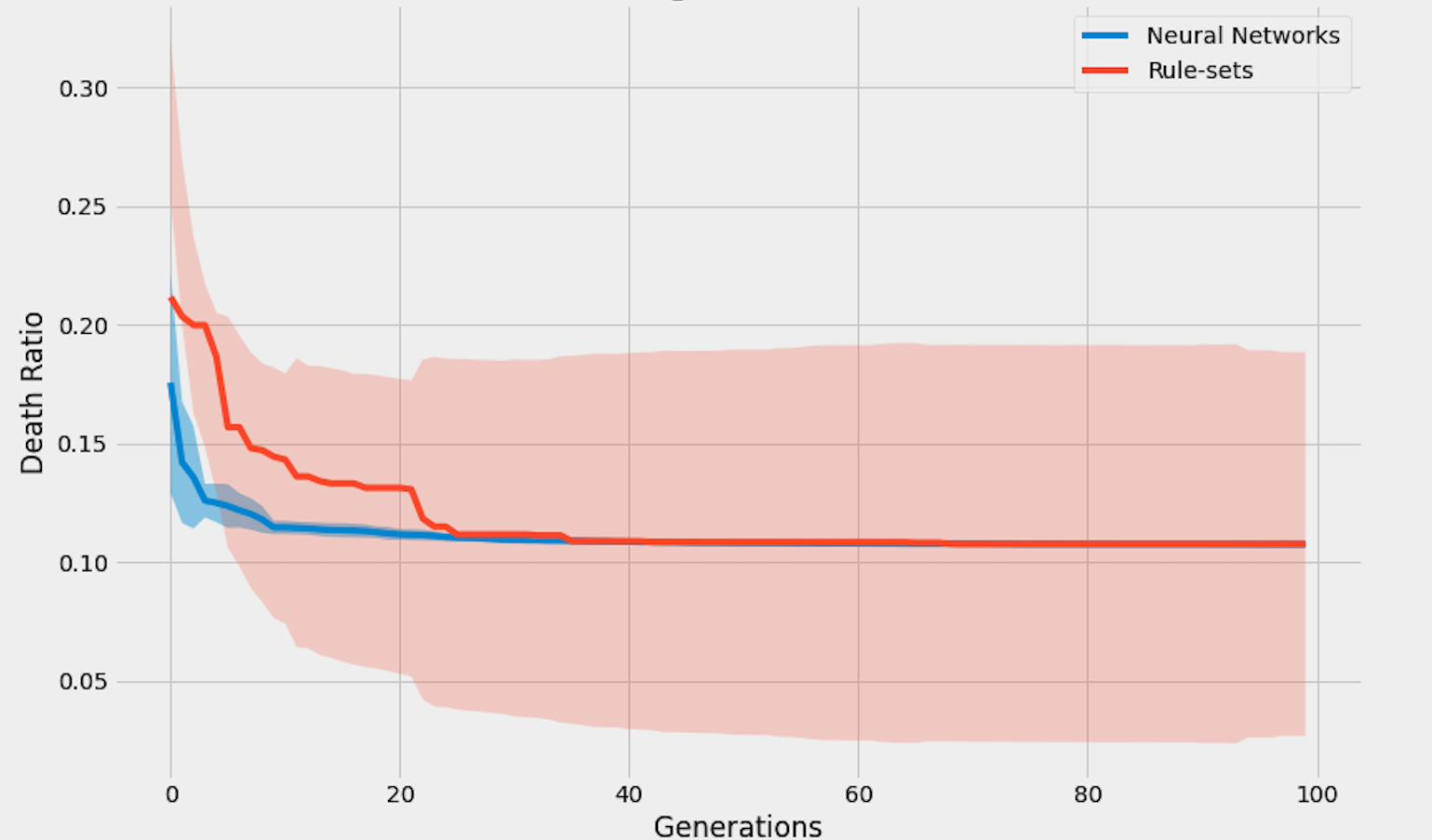}
\vspace*{-1ex}
    \caption{\emph{Evolution of neural-network and rule-set prescriptors in the single-objective domain of Heart-Failure Prevention.}	The plots are averaged over 10 runs and shading indicates 95\% confidence intervals. The rule set population varies more throughout evolution, but the final best solutions perform almost the same on average, with a death rate of slightly less than 0.11. The differences are not statistically significant ($p=0.76$). Thus, explainability is achieved with more diversity and practically no cost in performance.}
    \label{fig:hf_ref}
\end{minipage}
\end{figure}

\subsubsection{Heart-Failure Prevention Recommendations (Single Objective)}
\label{sc:hf}

\paragraph{Experiment}
A random-forest predictor was trained on the Heart-Failure data set \cite{chicco2020machine} to predict the probability of death for a patient given two possible interventions, i.e., ejection fraction and serum creatinine. The predictor achieved an out-of-sample Mathews correlation coefficient of 0.33. 

The prescriptor receives the patient's condition as its input and prescribes one of the two interventions. The predictor is queried to evaluate its fitness in minimizing deaths. The neural network prescriptor's input nodes correspond to the patient's condition and output nodes to the two possible treatments; it has a hidden layer of 16 nodes. Its weights were evolved over 100 generations using a population size of 100. Similarly, the rule-set prescriptor includes left-hand side clauses referring to the patient's condition and right-hand side actions specifying interventions. Evolution was similarly set to run for 100 generations with a population of 100.

\begin{figure}[!ht]
\vspace*{-5ex}
\centering
\scriptsize
\setlength{\lineskip}{-4.75pt}
\begin{align*}
1. &(0.08*\textit{anaemia} < 0.44*\textit{platelets}) &&\hspace*{-2ex}\longrightarrow 0.21*\textit{ejection.fraction}\\
2. &(0.08*\textit{platelets} \le 0.45 [0.0..1.0]) &&\hspace*{-2ex}\longrightarrow 0.59*\textit{ejection.fraction}\\
3. &(0.84*\textit{anaemia}^3 \le 0.40*\textit{age}) &&\hspace*{-2ex}\longrightarrow 0.04*\textit{serum.creatinine}\\
4. &(0.89*\textit{serum.sodium} > 0.97*\textit{smoking}) \And \\
&(0.47*\textit{serum.sodium} > 0.09*\textit{creatinine.phosphokinase}) &&\hspace*{-2ex}\longrightarrow 0.32*\textit{ejection.fraction}\\
5. &(1.00*\textit{age} > 0.97*\textit{platelets}) &&\hspace*{-2ex}\longrightarrow 0.49*\textit{ejection.fraction}\\
6. &(0.39*\textit{creatinine.phosphokinase} < 0.92*\textit{anaemia}) &&\hspace*{-2ex}\longrightarrow 0.03*\textit{serum.creatinine}\\
7. &(0.94*\textit{platelets} \ge 0.02*\textit{anaemia}) &&\hspace*{-2ex}\longrightarrow 0.65*\textit{ejection.fraction}\\
8. &(0.46*\textit{smoking} \le 0.50*\textit{diabetes}) &&\hspace*{-2ex}\longrightarrow 0.51*\textit{ejection.fraction}\\
9. &(0.49*\textit{anaemia} \le 0.92 [0.0..1.0]) \And \\
&(0.46*\textit{smoking} \le 0.50*\textit{diabetes}) \And \\
&(0.09*\textit{platelets} \le 0.78 [0.0..1.0]) &&\hspace*{-2ex}\longrightarrow 0.36*\textit{ejection.fraction}\\
10. &(0.08*\textit{high.blood.pressure}^3 \le 0.85*\textit{diabetes}) &&\hspace*{-2ex}\longrightarrow 0.21*\textit{ejection.fraction}\\
11. &(0.91*\textit{platelets} < 0.62*\textit{high.blood.pressure}) \And \\
&(0.61*\textit{age} < 0.88*\textit{high.blood.pressure}) \And \\
&(0.10*\textit{creatinine.phosphokinase} < 0.43*\textit{sex}) &&\hspace*{-2ex}\longrightarrow 0.46*\textit{ejection.fraction}\\
12. &(0.47*\textit{serum.sodium} > 0.09*\textit{creatinine.phosphokinase}) &&\hspace*{-2ex}\longrightarrow 0.32*\textit{ejection.fraction}\\
13. &(0.46*\textit{smoking} \le 0.50*\textit{diabetes}) \And \\
&(0.06*\textit{anaemia} > 0.82*\textit{creatinine.phosphokinase}) \And \\
&(0.03*\textit{high.blood.pressure} \ge 0.21*\textit{sex}) &&\hspace*{-2ex}\longrightarrow 0.51*\textit{ejection.fraction}\\
14. &(0.31*\textit{creatinine.phosphokinase} >= 0.02*\textit{high.blood.pressure}) &&\hspace*{-2ex}\longrightarrow 0.48*\textit{ejection.fraction}\\
15. &(0.47*\textit{serum.sodium} > 0.09*\textit{creatinine.phosphokinase}) \And \\
&(0.40*\textit{diabetes} \ge 0.33*\textit{serum.sodium}) &&\hspace*{-2ex}\longrightarrow 0.32*\textit{ejection.fraction}\\
16. &(0.17*\textit{smoking} < 0.30*\textit{high.blood.pressure}) \And \\
&(0.08*\textit{high.blood.pressure} \le 0.85*\textit{diabetes}) &&\hspace*{-2ex}\longrightarrow 0.21*\textit{ejection.fraction}\\
17. &(0.94*\textit{platelets} \ge 0.02*\textit{anaemia}) \And \\
&(0.46*\textit{smoking} \le 0.50*\textit{diabetes}) &&\hspace*{-2ex}\longrightarrow 0.65*\textit{ejection.fraction}\\
18. &(1.00*\textit{age} > 0.97*\textit{platelets}) \And \\
&(0.97*\textit{platelets} \le 0.36*\textit{diabetes}) \And \\
&(0.46*\textit{smoking} \le 0.50*\textit{diabetes}) &&\hspace*{-2ex}\longrightarrow 0.49*\textit{ejection.fraction}\\
19. &(0.46*\textit{smoking} \le 0.50*\textit{diabetes}) \And \\
&(0.08*\textit{age} > 0.85*\textit{diabetes}) \And \\
&(0.03*\textit{high.blood.pressure} \ge 0.21*\textit{sex}) &&\hspace*{-2ex}\longrightarrow 0.51*\textit{ejection.fraction}\\
20. &(0.91*\textit{platelets} < 0.62*\textit{high.blood.pressure}) \And \\
&(0.10*\textit{creatinine.phosphokinase} < 0.43*\textit{sex}) &&\hspace*{-2ex}\longrightarrow 0.46*\textit{ejection.fraction}\\
21. &(0.46*\textit{smoking} \le 0.50*\textit{diabetes}) \And \\
&(0.06*\textit{anaemia} > 0.82*\textit{creatinine.phosphokinase}) &&\hspace*{-2ex}\longrightarrow 0.51*\textit{ejection.fraction}\\
22. &(0.88*\textit{anaemia} > 0.76 [0.0..1.0]) \And \\
&(0.84*\textit{creatinine.phosphokinase} < 0.40*\textit{age}) \And \\
&(0.45*\textit{sex} < 0.85 [0.0..1.0]) &&\hspace*{-2ex}\longrightarrow 0.04*\textit{serum.creatinine}\\
23. &(0.46*\textit{smoking} \le 0.50*\textit{diabetes}) \And \\
&(0.08*\textit{anaemia} < 0.44*\textit{platelets}) &&\hspace*{-2ex}\longrightarrow 0.21*\textit{ejection.fraction}\\
24. &(0.63*\textit{high.blood.pressure} \le 0.79*\textit{serum.sodium}) \And \\
&(0.46*\textit{smoking} \le 0.50*\textit{diabetes}) \And \\
&(0.31*\textit{creatinine.phosphokinase} \ge 0.02*\textit{high.blood.pressure}) &&\hspace*{-2ex}\longrightarrow 0.48*\textit{ejection.fraction}\\
25. &(0.08*\textit{age} > 0.85*\textit{diabetes}) &&\hspace*{-2ex}\longrightarrow 0.21*\textit{ejection.fraction}\\
26. &(0.09*\textit{platelets} \le 0.78 [0.0..1.0]) \And \\
&(0.08*\textit{age} > 0.85*\textit{diabetes}) &&\hspace*{-2ex}\longrightarrow 0.36*\textit{ejection.fraction}\\
27. &(0.91*\textit{platelets} < 0.62*\textit{high.blood.pressure}) \And \\
&(0.24*\textit{serum.sodium} > 0.29*\textit{diabetes}) \And \\
&(0.10*\textit{creatinine.phosphokinase} < 0.43*\textit{sex}) \And \\
&(0.08*\textit{age} > 0.85*\textit{diabetes}) &&\hspace*{-2ex}\longrightarrow 0.46*\textit{ejection.fraction}\\
28. &(0.91*\textit{platelets} < 0.62*\textit{high.blood.pressure}) &&\hspace*{-2ex}\longrightarrow 0.46*\textit{ejection.fraction}\\
29. &(0.40*\textit{diabetes} \ge 0.33*\textit{serum.sodium}) &&\hspace*{-2ex}\longrightarrow 0.32*\textit{ejection.fraction}\\
30. &(0.31*\textit{creatinine.phosphokinase} \ge 0.02*\textit{high.blood.pressure}) &&\hspace*{-2ex}\longrightarrow 0.48*\textit{ejection.fraction}\\
31. &(0.84*\textit{creatinine.phosphokinase} \ge 0.40*\textit{age}) &&\hspace*{-2ex}\longrightarrow 0.04*\textit{serum.creatinine}\\
32. &(0.94*\textit{platelets} \ge 0.02*\textit{anaemia}) \And \\
&(0.08*\textit{platelets} \le 0.45 [0.0..1.0]) &&\hspace*{-2ex}\longrightarrow 0.59*\textit{ejection.fraction}\\
33. &(0.08*\textit{anaemia} < 0.44*\textit{platelets}) \And \\
&(0.08*\textit{age} > 0.85*\textit{diabetes}) &&\hspace*{-2ex}\longrightarrow 0.21*\textit{ejection.fraction}\\
34. &(0.47*\textit{serum.sodium} > 0.09*\textit{creatinine.phosphokinase}) \And \\
&(0.46*\textit{smoking} \le 0.50*\textit{diabetes}) \And \\
&(0.40*\textit{diabetes} \ge 0.33*\textit{serum.sodium}) &&\hspace*{-2ex}\longrightarrow 0.32*\textit{ejection.fraction}\\
35. &(0.39*\textit{creatinine.phosphokinase} < 0.92*\textit{anaemia}) \And \\
&(0.10*\textit{anaemia} \ge 0.06*\textit{serum.sodium}) &&\hspace*{-2ex}\longrightarrow 0.03*\textit{serum.creatinine}\\
36. &(0.91*\textit{platelets} < 0.62*\textit{high.blood.pressure}) \And \\
&(0.46*\textit{smoking} \le 0.50*\textit{diabetes}) \And \\
&(0.10*\textit{creatinine.phosphokinase} < 0.43*\textit{sex}) &&\hspace*{-2ex}\longrightarrow 0.46*\textit{ejection.fraction}\\
37. &(0.46*\textit{smoking} \le 0.50*\textit{diabetes}) \And \\
&(0.08*\textit{age} \le 0.85*\textit{diabetes}) &&\hspace*{-2ex}\longrightarrow 0.21*\textit{ejection.fraction}\\
38. &(0.46*\textit{smoking} \le 0.50*\textit{diabetes}) \And \\
&(0.08*\textit{platelets} \le 0.45 [0.0..1.0]) &&\hspace*{-2ex}\longrightarrow 0.59*\textit{ejection.fraction}\\
39. &(0.46*\textit{smoking} \le 0.50*\textit{diabetes}) \And \\
&(0.03*\textit{high.blood.pressure} \ge 0.21*\textit{sex}) &&\hspace*{-2ex}\longrightarrow 0.51*\textit{ejection.fraction}\\
40. &(0.41*\textit{high.blood.pressure} > 0.48*\textit{sex}) \And \\
&(0.09*\textit{platelets} \le 0.78 [0.0..1.0]) \And \\
&(0.08*\textit{age} > 0.85*\textit{diabetes}) &&\hspace*{-2ex}\longrightarrow 0.36*\textit{ejection.fraction}\\
41. &(0.46*\textit{smoking} \le 0.50*\textit{diabetes}) \And \\
&(0.09*\textit{platelets} \le 0.78 [0.0..1.0]) &&\hspace*{-2ex}\longrightarrow 0.36*\textit{ejection.fraction}\\
42. &(0.21*\textit{high.blood.pressure} \le 0.85*\textit{diabetes}) \And \\
&(0.17*\textit{smoking} < 0.30*\textit{high.blood.pressure}) &&\hspace*{-2ex}\longrightarrow 0.21*\textit{ejection.fraction}\\
43. &(1.00*\textit{age} > 0.97*\textit{platelets}) \And \\
&(0.97*\textit{platelets} \le 0.36*\textit{diabetes}) &&\hspace*{-2ex}\longrightarrow 0.49*\textit{ejection.fraction}\\
44. &(0.97*\textit{platelets} \le 0.36*\textit{diabetes}) &&\hspace*{-2ex}\longrightarrow 0.49*\textit{ejection.fraction}\\
45. &(0.23*\textit{diabetes} > 0.89*\textit{platelets}) \And \\
&(0.08*\textit{high.blood.pressure}^3 \le 0.85*\textit{diabetes}) &&\hspace*{-2ex}\longrightarrow 0.21*\textit{ejection.fraction}\\
46. &(0.14*\textit{creatinine.phosphokinase} > 0.75*\textit{smoking}) \And \\
&(0.08*\textit{age} \le 0.85*\textit{diabetes}) &&\hspace*{-2ex}\longrightarrow 0.21*\textit{ejection.fraction}\\
47. &~\text{Default} &&\hspace*{-2ex}\longrightarrow 0.13*\textit{serum.creatinine}\\
\end{align*}
\vspace*{-11ex}
\caption{\emph{A sample solution rule set for the Heart-Failure Prevention domain.} \label{fig:hf_rules}
	The default rule indicates serum, and the others mostly identify situations where ejection fraction is preferable. Many of the rules are redundant, which is a common result of the search process. Based on the $\mathit{times\_applied}$ counters, those rules can be removed, resulting in a compact set shown in Figure~\ref{fig:hf_active_rules}.}
\vspace*{-8ex}
\end{figure}

\begin{figure}[!ht]
\vspace*{-5ex}
\centering
\scriptsize
\setlength{\lineskip}{-4.75pt}
\begin{align*}
1.\hspace*{2.3ex}(1.) \hspace*{1.1ex}83&\hspace*{1.1ex}(0.08*\textit{anaemia} < 0.44*\textit{platelets}) &&\hspace*{-2ex}\longrightarrow 0.21*\textit{ejection.fraction}\\
2.\hspace*{1.1ex}(40.)\hspace*{1.1ex}66&\hspace*{1.1ex}(0.89*\textit{serum.sodium} > 0.97*\textit{smoking}) \And \\
&\hspace*{1.1ex}(0.47*\textit{serum.sodium} > 0.09*\textit{creatinine.phosphokinase}) &&\hspace*{-2ex}\longrightarrow 0.32*\textit{ejection.fraction}\\
3.\hspace*{2.2ex}(8.) \hspace*{0.0ex}102&\hspace*{1.1ex}(0.46*\textit{smoking} \le 0.50*\textit{diabetes}) &&\hspace*{-2ex}\longrightarrow 0.51*\textit{ejection.fraction}\\
4.\hspace*{1.1ex}(10.)\hspace*{1.1ex}78&\hspace*{1.1ex}(0.08*\textit{high.blood.pressure}^3 \le 0.85*\textit{diabetes}) &&\hspace*{-2ex}\longrightarrow 0.21*\textit{ejection.fraction}\\
5.\hspace*{1.1ex}(11.)\hspace*{2.2ex}6&\hspace*{1.1ex}(0.91*\textit{platelets} < 0.62*\textit{high.blood.pressure}) \And \\
&\hspace*{1.1ex}(0.61*\textit{age} < 0.88*\textit{high.blood.pressure}) \And \\
&\hspace*{1.1ex}(0.10*\textit{creatinine.phosphokinase} < 0.43*\textit{sex}) &&\hspace*{-2ex}\longrightarrow 0.46*\textit{ejection.fraction}\\
6.\hspace*{1.1ex}(12.)\hspace*{1.1ex}73&\hspace*{1.1ex}(0.47*\textit{serum.sodium} > 0.09*\textit{creatinine.phosphokinase}) &&\hspace*{-2ex}\longrightarrow 0.32*\textit{ejection.fraction}\\
7.\hspace*{1.1ex}(13.)\hspace*{1.1ex}11&\hspace*{1.1ex}(0.46*\textit{smoking} \le 0.50*\textit{diabetes}) \And \\
&\hspace*{1.1ex}(0.06*\textit{anaemia} > 0.82*\textit{creatinine.phosphokinase}) \And \\
&\hspace*{1.1ex}(0.03*\textit{high.blood.pressure} \ge 0.21*\textit{sex}) &&\hspace*{-2ex}\longrightarrow 0.51*\textit{ejection.fraction}\\
8.\hspace*{1.1ex}(14.)\hspace*{0.0ex}108&\hspace*{1.1ex}(0.31*\textit{creatinine.phosphokinase} >= 0.02*\textit{high.blood.pressure}) &&\hspace*{-2ex}\longrightarrow 0.48*\textit{ejection.fraction}\\
9.\hspace*{1.1ex}(15.)\hspace*{1.1ex}37&\hspace*{1.1ex}(0.47*\textit{serum.sodium} > 0.09*\textit{creatinine.phosphokinase}) \And \\
&\hspace*{1.1ex}(0.40*\textit{diabetes} \ge 0.33*\textit{serum.sodium}) &&\hspace*{-2ex}\longrightarrow 0.32*\textit{ejection.fraction}\\
10.\hspace*{1.1ex}(16.)\hspace*{1.1ex}31&\hspace*{1.1ex}(0.17*\textit{smoking} < 0.30*\textit{high.blood.pressure}) \And \\
&\hspace*{1.1ex}(0.08*\textit{high.blood.pressure} \le 0.85*\textit{diabetes}) &&\hspace*{-2ex}\longrightarrow 0.21*\textit{ejection.fraction}\\
11.\hspace*{1.1ex}(19.)\hspace*{2.2ex}8&\hspace*{1.1ex}(0.46*\textit{smoking} \le 0.50*\textit{diabetes}) \And \\
&\hspace*{1.1ex}(0.08*\textit{age} > 0.85*\textit{diabetes}) \And \\
&\hspace*{1.1ex}(0.03*\textit{high.blood.pressure} \ge 0.21*\textit{sex}) &&\hspace*{-2ex}\longrightarrow 0.51*\textit{ejection.fraction}\\
12.\hspace*{1.1ex}(20.)\hspace*{1.1ex}26&\hspace*{1.1ex}(0.91*\textit{platelets} < 0.62*\textit{high.blood.pressure}) \And \\
&\hspace*{1.1ex}(0.10*\textit{creatinine.phosphokinase} < 0.43*\textit{sex}) &&\hspace*{-2ex}\longrightarrow 0.46*\textit{ejection.fraction}\\
13.\hspace*{1.1ex}(23.)\hspace*{1.1ex}14&\hspace*{1.1ex}(0.46*\textit{smoking} \le 0.50*\textit{diabetes}) \And \\
&\hspace*{1.1ex}(0.08*\textit{anaemia} < 0.44*\textit{platelets}) &&\hspace*{-2ex}\longrightarrow 0.21*\textit{ejection.fraction}\\
14.\hspace*{1.1ex}(28.)\hspace*{1.1ex}73&\hspace*{1.1ex}(0.91*\textit{platelets} < 0.62*\textit{high.blood.pressure}) &&\hspace*{-2ex}\longrightarrow 0.46*\textit{ejection.fraction}\\
15.\hspace*{1.1ex}(30.)\hspace*{0.0ex}103&\hspace*{1.1ex}(0.31*\textit{creatinine.phosphokinase} \ge 0.02*\textit{high.blood.pressure}) &&\hspace*{-2ex}\longrightarrow 0.48*\textit{ejection.fraction}\\
16.\hspace*{1.1ex}(31.)\hspace*{1.1ex}97&\hspace*{1.1ex}(0.84*\textit{creatinine.phosphokinase} \ge 0.40*\textit{age}) &&\hspace*{-2ex}\longrightarrow 0.04*\textit{serum.creatinine}\\
17.\hspace*{1.1ex}(33.)\hspace*{1.1ex}18&\hspace*{1.1ex}(0.08*\textit{anaemia} < 0.44*\textit{platelets}) \And \\
&\hspace*{1.1ex}(0.08*\textit{age} > 0.85*\textit{diabetes}) &&\hspace*{-2ex}\longrightarrow 0.21*\textit{ejection.fraction}\\
18.\hspace*{1.1ex}(35.)\hspace*{1.1ex}21&\hspace*{1.1ex}(0.39*\textit{creatinine.phosphokinase} < 0.92*\textit{anaemia}) \And \\
&\hspace*{1.1ex}(0.10*\textit{anaemia} \ge 0.06*\textit{serum.sodium}) &&\hspace*{-2ex}\longrightarrow 0.03*\textit{serum.creatinine}\\
19.\hspace*{1.1ex}(36.)\hspace*{2.2ex}6&\hspace*{1.1ex}(0.91*\textit{platelets} < 0.62*\textit{high.blood.pressure}) \And \\
&\hspace*{1.1ex}(0.46*\textit{smoking} \le 0.50*\textit{diabetes}) \And \\
&\hspace*{1.1ex}(0.10*\textit{creatinine.phosphokinase} < 0.43*\textit{sex}) &&\hspace*{-2ex}\longrightarrow 0.46*\textit{ejection.fraction}\\
20.\hspace*{1.1ex}(37.)\hspace*{1.1ex}14&\hspace*{1.1ex}(0.46*\textit{smoking} \le 0.50*\textit{diabetes}) \And \\
&\hspace*{1.1ex}(0.08*\textit{age} \le 0.85*\textit{diabetes}) &&\hspace*{-2ex}\longrightarrow 0.21*\textit{ejection.fraction}\\
21.\hspace*{1.1ex}(38.)\hspace*{1.1ex}32&\hspace*{1.1ex}(0.46*\textit{smoking} \le 0.50*\textit{diabetes}) \And \\
&\hspace*{1.1ex}(0.08*\textit{platelets} \le 0.45 [0.0..1.0]) &&\hspace*{-2ex}\longrightarrow 0.59*\textit{ejection.fraction}\\
22.\hspace*{1.1ex}(39.)\hspace*{1.1ex}15&\hspace*{1.1ex}(0.46*\textit{smoking} \le 0.50*\textit{diabetes}) \And \\
&\hspace*{1.1ex}(0.03*\textit{high.blood.pressure} \ge 0.21*\textit{sex}) &&\hspace*{-2ex}\longrightarrow 0.51*\textit{ejection.fraction}\\
23.\hspace*{1.1ex}(40.)\hspace*{2.2ex}9&\hspace*{1.1ex}(0.41*\textit{high.blood.pressure} > 0.48*\textit{sex}) \And \\
&\hspace*{1.1ex}(0.09*\textit{platelets} \le 0.78 [0.0..1.0]) \And \\
&\hspace*{1.1ex}(0.08*\textit{age} > 0.85*\textit{diabetes}) &&\hspace*{-2ex}\longrightarrow 0.36*\textit{ejection.fraction}\\
24.\hspace*{1.1ex}(41.)\hspace*{1.1ex}17&\hspace*{1.1ex}(0.46*\textit{smoking} \le 0.50*\textit{diabetes}) \And \\
&\hspace*{1.1ex}(0.09*\textit{platelets} \le 0.78 [0.0..1.0]) &&\hspace*{-2ex}\longrightarrow 0.36*\textit{ejection.fraction}\\
25.\hspace*{1.1ex}(42.)\hspace*{1.1ex}32&\hspace*{1.1ex}(0.21*\textit{high.blood.pressure} \le 0.85*\textit{diabetes}) \And \\
&\hspace*{1.1ex}(0.17*\textit{smoking} < 0.30*\textit{high.blood.pressure}) &&\hspace*{-2ex}\longrightarrow 0.21*\textit{ejection.fraction}\\
26.\hspace*{1.1ex}(43.)\hspace*{1.1ex}15&\hspace*{1.1ex}(1.00*\textit{age} > 0.97*\textit{platelets}) \And \\
&\hspace*{1.1ex}(0.97*\textit{platelets} \le 0.36*\textit{diabetes}) &&\hspace*{-2ex}\longrightarrow 0.49*\textit{ejection.fraction}\\
27.\hspace*{1.1ex}(44.)\hspace*{1.1ex}72&\hspace*{1.1ex}(0.97*\textit{platelets} \le 0.36*\textit{diabetes}) &&\hspace*{-2ex}\longrightarrow 0.49*\textit{ejection.fraction}\\
28.\hspace*{1.1ex}(45.)\hspace*{1.1ex}23&\hspace*{1.1ex}(0.23*\textit{diabetes} > 0.89*\textit{platelets}) \And \\
&\hspace*{1.1ex}(0.08*\textit{high.blood.pressure}^3 \le 0.85*\textit{diabetes}) &&\hspace*{-2ex}\longrightarrow 0.21*\textit{ejection.fraction}\\
29.\hspace*{1.1ex}(47.)\hspace*{1.1ex}10&\hspace*{1.1ex}\hspace*{0.25ex}\text{Default} &&\hspace*{-2ex}\longrightarrow 0.13*\textit{serum.creatinine}\\
\end{align*}
\vspace*{-8ex}
\caption{\emph{Solution rule set from Figure~\ref{fig:hf_rules} compressed by pruning inactive rules.} \label{fig:hf_active_rules}
	The original rule number is in parenthesis, followed by the number of times the rule was applied (out of 300 samples). This set is functionally equivalent but much smaller. Also, on average only four rules are active for each case in the dataset. Such conciseness makes interpretation and explanation easier, supporting applications in safety-critical fields such as healthcare, finance, and law.}
 \vspace*{-1ex}
\end{figure}

\paragraph{Results}

The effectiveness of the two prescriptor models in preventing death is shown over evolution time in Figure~\ref{fig:hf_ref}, averaged over ten runs. It took more generations for rule-set evolution to reach the same level as the neural network evolution, and there is more diversity and variance in the rule-set populations. The likely reason is that even small changes in rules (e.g., adding or removing a clause) can have a large effect on performance. However, on average, the final results are very similar: The best rule set performs at 10.8\% and the best neural network at 10.7\%. Thus, explainability is achieved with practically no cost in performance. With such variability, an interesting question is:  How can one candidate be chosen in the end such that it most likely generalizes to unseen data, given the multiple hypothesis problem \cite{avdeeva_1966,mht2014}? It turns out that under a smoothness assumption that similar candidates perform similarly, picking a candidate based on the average performance of its neighbors provides a viable strategy \cite{miikkulainen:ssci17}.

A sample rule-set solution is shown in Figure~\ref{fig:hf_rules}. This rule set prioritizes scenarios where the ejection fraction is favorable and utilizes serum creatinine as the default prescription. As discussed in Section~\ref{sc:flappybird}, there are many redundant rules as well. To improve interpretability, rules with $\mathit{times\_applied}=0$ were pruned, resulting in a much smaller set with 29, as shown in Figure~\ref{fig:hf_active_rules}.  Each of these rules participates in making decisions on this dataset, however, only four of them are activated for a given input on average. Such conciseness makes interpretation and explanation easier: It is possible to gain a clear understanding of the key factors driving the decision-making process in the domain. Such transparency and interpretability make it possible to deploy AI systems in safety-critical fields such as healthcare, finance, and law.

\subsubsection{Diabetes Treatment Recommendations (Multiobjective)}
\label{sc:diabetes}

\begin{figure}[!ht]
  \begin{minipage}[b]{0.48\textwidth}
  \centering
  \includegraphics[width=\textwidth]{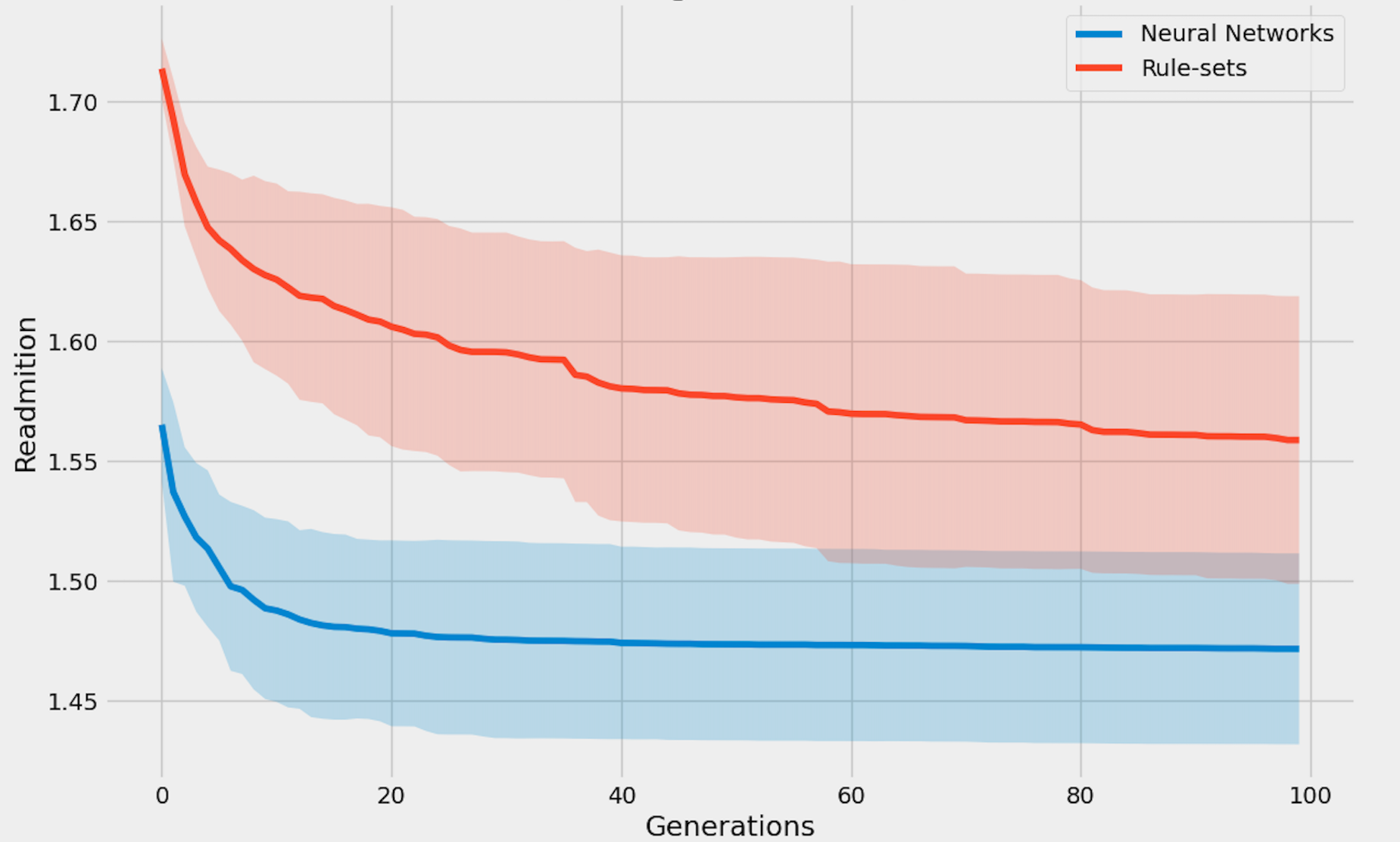}\\
  {\footnotesize ($a$) Readmission likelihood}
  \end{minipage}
  \hfill
  \begin{minipage}[b]{0.48\textwidth}
  \centering
  \includegraphics[width=\textwidth]{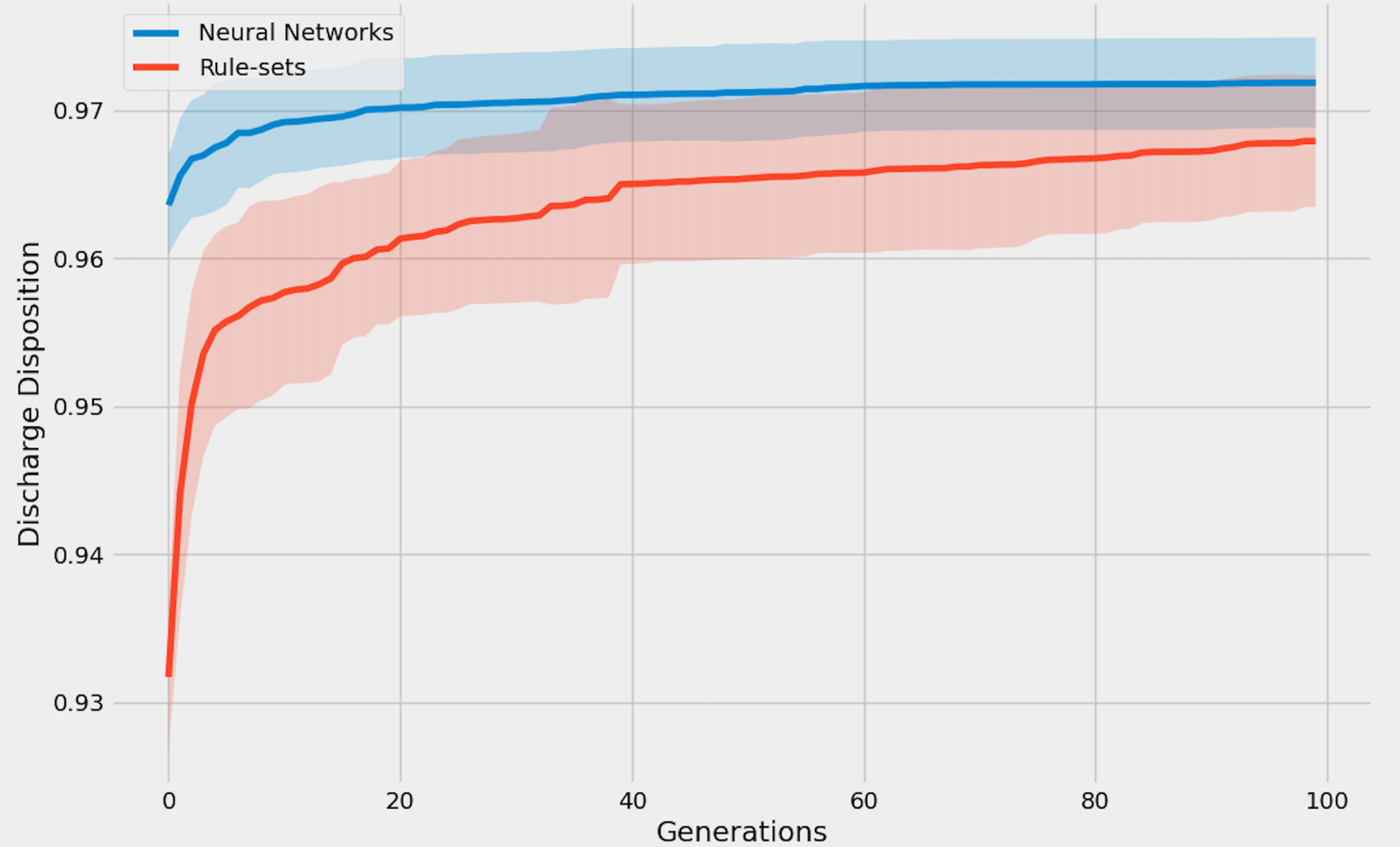}\\
  {\footnotesize ($b$) Discharge disposition quality}
  \end{minipage}
\vspace*{-2ex}
   \caption{\emph{Evolution of neural-network and rule-set prescriptors in the multiobjective domain of Diabetes Treatment.} ($a$) Progress in the readmission objective and ($b$) discharge disposition objective over evolution; the graphs are averages over 10 runs and shaded areas indicate 95\% confidence intervals. While the rule-set population exhibits more variation during evolution, the overlapping confidence intervals suggest that the performance differences between the two approaches are small ($p=0.0012$ in $a$ and $p=0.0327$ in $b$), thus justifying the cost of explainability. }
    \label{fig:diabetes_ref}
\end{figure}

\paragraph{Experiment}
In order to assess the cost of explainability in a domain with multiple objectives, a second comparison was run on the same Diabetes Treatment recommendation task as in Section~\ref{sc:insight}. Again, the goal was to minimize hospital readmissions while simultaneously improving discharge dispositions. Rule-based ESP was again used to evolve a population of 100 rule sets and 100 neural networks over 100 generations, and evolution repeated ten times.

\paragraph{Results}
Figure~\ref{fig:diabetes_ref} provides an overview of the results obtained. Again, the variation in the evolved rule sets is slightly higher compared to the neural networks across the ten runs. This variation can be attributed to the discrete nature of the rule set components in contrast to the continuous values of the weights in neural networks. Although the average performance curves are noticeably different, their confidence intervals still overlap. This result suggests that the performance cost associated with the explainability provided by the rule sets is small (i.e.\ barely significant).

In sum, the experiments in both single and multiobjective domains suggest that there may be a small cost in performance when evolving rule sets vs. neural network prescriptors. Depending on the domain, such cost may be well worth it in order to achieve transparent, explainable performance.

\section{Discussion and Future Work}
\label{sc:discussion}

The experiments presented in this paper suggest that rule-set evolution in general and its EVOTER implementation in particular is a practical and effective approach to machine learning. EVOTER applies to a wide range of tasks, including prediction, classification, prescription, and policy search, both in static and time-series contexts. It is possible to discover solutions directly in the domain, or through surrogate models that represent the domain safely and cost-effectively. Most importantly, compared to black-box models such as neural networks, rule-set solutions are explainable, i.e.\ both interpretable and transparent. They can provide insight into the domain, uncovering fundamental principles as well as hidden biases. While there may be a slight performance cost compared to black-box models, it may be well worth it in many safety-critical domains.

There are several compelling directions for future work. In terms of technical extensions, simplicity could be further encouraged through an explicit secondary evolution objective to favor smaller rule sets, incorporated e.g.\ through NSGA-II \cite{Deb2002}. Similarly, exploration and creativity can be encouraged through a secondary novelty objective \cite{shahrzad:alife18,shahrzad:gptp20}.  To encourage general solutions, the $\mathit{times\_applied}$ counter can be useful: A very low count may indicate overfitting or even a corner case bias, and evolution can be directed to avoid such cases. Further, the rule grammar could be expanded during evolution gradually. The process could be started with a restricted set of operators that could then be expanded as evolution progresses, thus implementing a form of curricular learning. In cases where the data set may be excessively large, partial and incremental evaluation through age-layering can be used to make the process more efficient \cite{bp-sepsis,Hodjat2013,Shahrzad2016}. EVOTER also lends itself well to parallelization and federation, both at the level of evaluating individuals and at the level of evaluating conditions.  Further, condition evaluation can be tensorized and evaluated in parallel on different data points, taking advantage of GPUs.

A useful practical extension to EVOTER would be to include a facility for the expert to edit the discovered rule sets. First, such a facility could be used to identify and remove redundancies, making the rules easier to read. Although much of such simplification can be done automatically, there is still room for human insight. Second, the facility can provide a means for removing biases and making sure that the models are fair. Unlike with opaque models, the learned biases are explicit and can be evaluated and edited out if desired. Third, it may be possible to edit the rules to add further knowledge that may be well understood but challenging to learn, such as safety limits and other real-world constraints.

More generally, it may be possible to build rule-set twins for existing SoTA black-box models. In this manner, converting an opaque but well-performing model into a transparent rule-set equivalent may be possible. This model can then be used to explain the learned behavior and identify its biases and potential weaknesses. Such rule-set models are executable, and if they replicate the performance of the black-box model accurately enough, they could be deployed instead of the black box in explanation-critical applications. In this manner, the process of rule set distillation can play a vital role in deploying machine learning systems in the real world.

It may also be possible to create hybrid systems that take advantage of both black-box and explainable elements. Such a system could e.g.\ use deep neural networks to extract visual features from medical images, such as the position, dimension, and texture of lumps in X-rays, followed by EVOTER to derive transparent, verifiable rules based on these features. This methodology would not only make it possible to audit the system but also allow an expert to verify the features extracted by the deep network, thus resulting in a more robust and trustworthy machine-learning application.

To encourage adoption further, Large Language Models (LLMs) \cite{anil:arxiv23,bubeck:arxiv23,touvron:arxiv23} could be used to translate EVOTER rules into plain language. Given the propositional rules as input, LLMs should be able to generate natural language explanations, descriptions, and interpretations. This approach would enable users with varying levels of expertise to comprehend the underlying concepts and their functions. It could thus potentially bridge the gap between intricate machine learning techniques and a broader audience, making it adoption of such systems easier.

\section{Conclusion}
\label{sc:conclusion}

This study demonstrated rule-set evolution, implemented in EVOTER, as a promising approach to developing inherently transparent and explainable models. The approach contrasts with the dominant black-box models, such as neural networks, in machine learning. Rule-set evolution can be applied to various tasks, including prediction, classification, prescription, and policy search, to both static and time-series problems, and with and without surrogate models of the domain. While a slight performance cost can be expected compared to black-box models, it may be acceptable in many cases considering the benefits of explainability. Rule-set evolution can thus enhance the trust, understanding, and adoption of AI systems in practical applications.

\newpage
\bibliographystyle{ACM-Reference-Format}
\bibliography{refs}

\begin{thebibliography}{}

\end{thebibliography}



\begin{thebibliography}{50}


\ifx \showCODEN    \undefined \def \showCODEN     #1{\unskip}     \fi
\ifx \showDOI      \undefined \def \showDOI       #1{#1}\fi
\ifx \showISBNx    \undefined \def \showISBNx     #1{\unskip}     \fi
\ifx \showISBNxiii \undefined \def \showISBNxiii  #1{\unskip}     \fi
\ifx \showISSN     \undefined \def \showISSN      #1{\unskip}     \fi
\ifx \showLCCN     \undefined \def \showLCCN      #1{\unskip}     \fi
\ifx \shownote     \undefined \def \shownote      #1{#1}          \fi
\ifx \showarticletitle \undefined \def \showarticletitle #1{#1}   \fi
\ifx \showURL      \undefined \def \showURL       {\relax}        \fi
\providecommand\bibfield[2]{#2}
\providecommand\bibinfo[2]{#2}
\providecommand\natexlab[1]{#1}
\providecommand\showeprint[2][]{arXiv:#2}

\bibitem[Adadi and Berrada(2018)]%
        {xai-survey-2018}
\bibfield{author}{\bibinfo{person}{Amina Adadi} {and} \bibinfo{person}{Mohammed Berrada}.} \bibinfo{year}{2018}\natexlab{}.
\newblock \showarticletitle{Peeking Inside the Black-Box: A Survey on Explainable Artificial Intelligence (XAI)}.
\newblock \bibinfo{journal}{\emph{IEEE Access}}  \bibinfo{volume}{6} (\bibinfo{year}{2018}), \bibinfo{pages}{52138--52160}.
\newblock


\bibitem[Akaike(1974)]%
        {akaike1974}
\bibfield{author}{\bibinfo{person}{H. Akaike}.} \bibinfo{year}{1974}\natexlab{}.
\newblock \showarticletitle{A new look at the statistical model identification}.
\newblock \bibinfo{journal}{\emph{IEEE Trans. Automat. Control}} \bibinfo{volume}{19}, \bibinfo{number}{6} (\bibinfo{year}{1974}), \bibinfo{pages}{716--723}.
\newblock


\bibitem[Austin et~al\mbox{.}(2014)]%
        {mht2014}
\bibfield{author}{\bibinfo{person}{RS Austin}, \bibinfo{person}{Isaac Dialsingh}, {and} \bibinfo{person}{NS Altman}.} \bibinfo{year}{2014}\natexlab{}.
\newblock \showarticletitle{Multiple Hypothesis Testing: A Review}.
\newblock \bibinfo{journal}{\emph{Journal of the Indian Society of Agricultural Statistics}}  \bibinfo{volume}{68} (\bibinfo{date}{01} \bibinfo{year}{2014}), \bibinfo{pages}{303--314}.
\newblock


\bibitem[Avdeeva(1966)]%
        {avdeeva_1966}
\bibfield{author}{\bibinfo{person}{Ligiia~Igorevna Avdeeva}.} \bibinfo{year}{1966}\natexlab{}.
\newblock \bibinfo{booktitle}{\emph{Simultaneous statistical inference}}.
\newblock \bibinfo{publisher}{Springer}.
\newblock


\bibitem[Belle and Papantonis(2021)]%
        {explainable-ML}
\bibfield{author}{\bibinfo{person}{Vaishak Belle} {and} \bibinfo{person}{Ioannis Papantonis}.} \bibinfo{year}{2021}\natexlab{}.
\newblock \showarticletitle{Principles and Practice of Explainable Machine Learning}.
\newblock \bibinfo{journal}{\emph{Frontiers in Big Data}}  \bibinfo{volume}{4} (\bibinfo{year}{2021}).
\newblock
\showISSN{2624-909X}


\bibitem[Bodria et~al\mbox{.}(2021)]%
        {Bodria2021BenchmarkingAS}
\bibfield{author}{\bibinfo{person}{Francesco Bodria}, \bibinfo{person}{Fosca Giannotti}, \bibinfo{person}{Riccardo Guidotti}, \bibinfo{person}{Francesca Naretto}, \bibinfo{person}{Dino Pedreschi}, {and} \bibinfo{person}{Salvatore Rinzivillo}.} \bibinfo{year}{2021}\natexlab{}.
\newblock \showarticletitle{Benchmarking and survey of explanation methods for black box models}.
\newblock \bibinfo{journal}{\emph{Data Mining and Knowledge Discovery}}  \bibinfo{volume}{37} (\bibinfo{year}{2021}), \bibinfo{pages}{1719--1778}.
\newblock
\urldef\tempurl%
\url{https://api.semanticscholar.org/CorpusID:232046272}
\showURL{%
\tempurl}


\bibitem[Brockman et~al\mbox{.}(2016)]%
        {brockman2016gym}
\bibfield{author}{\bibinfo{person}{Greg Brockman}, \bibinfo{person}{Vicki Cheung}, \bibinfo{person}{Ludwig Pettersson}, \bibinfo{person}{Jonas Schneider}, \bibinfo{person}{John Schulman}, \bibinfo{person}{Jie Tang}, {and} \bibinfo{person}{Wojciech Zaremba}.} \bibinfo{year}{2016}\natexlab{}.
\newblock \showarticletitle{{O}pen{AI} {G}ym}.
\newblock \bibinfo{journal}{\emph{arXiv:1606.01540}} (\bibinfo{year}{2016}).
\newblock


\bibitem[Bubeck et~al\mbox{.}(2023)]%
        {bubeck:arxiv23}
\bibfield{author}{\bibinfo{person}{Sébastien Bubeck}, \bibinfo{person}{Varun Chandrasekaran}, \bibinfo{person}{Ronen Eldan}, \bibinfo{person}{Johannes Gehrke}, \bibinfo{person}{Eric Horvitz}, \bibinfo{person}{Ece Kamar}, \bibinfo{person}{Peter Lee}, \bibinfo{person}{Yin~Tat Lee}, \bibinfo{person}{Yuanzhi Li}, \bibinfo{person}{Scott Lundberg}, \bibinfo{person}{Harsha Nori}, \bibinfo{person}{Hamid Palangi}, \bibinfo{person}{Marco~Tulio Ribeiro}, {and} \bibinfo{person}{Yi Zhang}.} \bibinfo{year}{2023}\natexlab{}.
\newblock \showarticletitle{Sparks of Artificial General Intelligence: Early experiments with GPT-4}.
\newblock \bibinfo{journal}{\emph{arXiv:2303.12712}} (\bibinfo{year}{2023}).
\newblock


\bibitem[Chicco and Jurman(2020)]%
        {chicco2020machine}
\bibfield{author}{\bibinfo{person}{Davide Chicco} {and} \bibinfo{person}{Giuseppe Jurman}.} \bibinfo{year}{2020}\natexlab{}.
\newblock \showarticletitle{Machine learning can predict survival of patients with heart failure from serum creatinine and ejection fraction alone}.
\newblock \bibinfo{journal}{\emph{BMC medical informatics and decision making}} \bibinfo{volume}{20}, \bibinfo{number}{1} (\bibinfo{year}{2020}), \bibinfo{pages}{1--16}.
\newblock


\bibitem[Deb et~al\mbox{.}(2002)]%
        {Deb2002}
\bibfield{author}{\bibinfo{person}{K. Deb}, \bibinfo{person}{A. Pratap}, \bibinfo{person}{S. Agarwal}, {and} \bibinfo{person}{T. Meyarivan}.} \bibinfo{year}{2002}\natexlab{}.
\newblock \showarticletitle{A fast and elitist multiobjective genetic algorithm: NSGA-II}.
\newblock \bibinfo{journal}{\emph{IEEE Transactions on Evolutionary Computation}} \bibinfo{volume}{6}, \bibinfo{number}{2} (\bibinfo{year}{2002}), \bibinfo{pages}{182--197}.
\newblock
\urldef\tempurl%
\url{https://doi.org/10.1109/4235.996017}
\showDOI{\tempurl}


\bibitem[Doshi-Velez and Kim(2017)]%
        {Doshi_2017}
\bibfield{author}{\bibinfo{person}{Finale Doshi-Velez} {and} \bibinfo{person}{Been Kim}.} \bibinfo{year}{2017}\natexlab{}.
\newblock \showarticletitle{Towards A Rigorous Science of Interpretable Machine Learning}.
\newblock \bibinfo{journal}{\emph{arXiv:1702:08608}} (\bibinfo{year}{2017}).
\newblock


\bibitem[Dua and Graff(2017)]%
        {Dua:2019}
\bibfield{author}{\bibinfo{person}{Dheeru Dua} {and} \bibinfo{person}{Casey Graff}.} \bibinfo{year}{2017}\natexlab{}.
\newblock \bibinfo{title}{{UCI} Machine Learning Repository}.
\newblock
\newblock
\urldef\tempurl%
\url{http://archive.ics.uci.edu/ml}
\showURL{%
\tempurl}


\bibitem[Efron(1992)]%
        {efron1992bootstrap}
\bibfield{author}{\bibinfo{person}{Bradley Efron}.} \bibinfo{year}{1992}\natexlab{}.
\newblock \showarticletitle{Bootstrap methods: another look at the jackknife}.
\newblock \bibinfo{journal}{\emph{Springer Series in Statistics}}  \bibinfo{volume}{7} (\bibinfo{year}{1992}), \bibinfo{pages}{1--26}.
\newblock


\bibitem[et~al.(2023)]%
        {anil:arxiv23}
\bibfield{author}{\bibinfo{person}{Rohan~Anil et al.}} \bibinfo{year}{2023}\natexlab{}.
\newblock \showarticletitle{PaLM 2 Technical Report}.
\newblock \bibinfo{journal}{\emph{arXiv:2305.10403}} (\bibinfo{year}{2023}).
\newblock


\bibitem[Fernandes et~al\mbox{.}(2023)]%
        {fernandes2023hotgp}
\bibfield{author}{\bibinfo{person}{Matheus~Campos Fernandes}, \bibinfo{person}{Fabrício Olivetti~De França}, {and} \bibinfo{person}{Emilio Francesquini}.} \bibinfo{year}{2023}\natexlab{}.
\newblock \showarticletitle{HOTGP-Higher-Order Typed Genetic Programming}. In \bibinfo{booktitle}{\emph{Proceedings of the Genetic and Evolutionary Computation Conference}}. ACM.
\newblock


\bibitem[Fonseca and Poças(2023)]%
        {fonseca2023comparing}
\bibfield{author}{\bibinfo{person}{Alcides Fonseca} {and} \bibinfo{person}{Diogo Poças}.} \bibinfo{year}{2023}\natexlab{}.
\newblock \showarticletitle{Comparing the Expressive Power of Strongly-Typed and Grammar-Guided Genetic Programming}. In \bibinfo{booktitle}{\emph{Proceedings of the Genetic and Evolutionary Computation Conference}}. ACM.
\newblock


\bibitem[Francon et~al\mbox{.}(2020)]%
        {esp-rl}
\bibfield{author}{\bibinfo{person}{Olivier Francon}, \bibinfo{person}{Santiago Gonzalez}, \bibinfo{person}{Babak Hodjat}, \bibinfo{person}{Elliot Meyerson}, \bibinfo{person}{Risto Miikkulainen}, \bibinfo{person}{Xin Qiu}, {and} \bibinfo{person}{Hormoz Shahrzad}.} \bibinfo{year}{2020}\natexlab{}.
\newblock \showarticletitle{Effective reinforcement learning through evolutionary surrogate-assisted prescription}. In \bibinfo{booktitle}{\emph{Proceedings of the 2020 Genetic and Evolutionary Computation Conference}}. \bibinfo{pages}{814--822}.
\newblock


\bibitem[Fushiki(2011)]%
        {fushiki2011bootstrap}
\bibfield{author}{\bibinfo{person}{T Fushiki}.} \bibinfo{year}{2011}\natexlab{}.
\newblock \showarticletitle{Estimation of prediction error by using K-fold cross-validation and bootstrap for sparse logit models}.
\newblock \bibinfo{journal}{\emph{Journal of Statistical Computation and Simulation}} \bibinfo{volume}{81}, \bibinfo{number}{2} (\bibinfo{year}{2011}), \bibinfo{pages}{141--155}.
\newblock


\bibitem[Gomez and Miikkulainen(1997)]%
        {gomez}
\bibfield{author}{\bibinfo{person}{Faustino Gomez} {and} \bibinfo{person}{Risto Miikkulainen}.} \bibinfo{year}{1997}\natexlab{}.
\newblock \showarticletitle{Incremental evolution of complex general behavior}.
\newblock \bibinfo{journal}{\emph{Adaptive Behavior}} \bibinfo{volume}{5}, \bibinfo{number}{3-4} (\bibinfo{year}{1997}), \bibinfo{pages}{317--342}.
\newblock


\bibitem[Gunning and Aha(2019)]%
        {darpa_xai_2019}
\bibfield{author}{\bibinfo{person}{David Gunning} {and} \bibinfo{person}{David Aha}.} \bibinfo{year}{2019}\natexlab{}.
\newblock \showarticletitle{DARPA’s Explainable Artificial Intelligence (XAI) Program}.
\newblock \bibinfo{journal}{\emph{AI Magazine}} \bibinfo{volume}{40}, \bibinfo{number}{2} (\bibinfo{date}{Jun.} \bibinfo{year}{2019}), \bibinfo{pages}{44--58}.
\newblock


\bibitem[Hastie et~al\mbox{.}(2009)]%
        {hastie2009elements}
\bibfield{author}{\bibinfo{person}{Trevor Hastie}, \bibinfo{person}{Robert Tibshirani}, {and} \bibinfo{person}{Jerome Friedman}.} \bibinfo{year}{2009}\natexlab{}.
\newblock \bibinfo{booktitle}{\emph{The Elements of Statistical Learning: Data Mining, Inference, and Prediction}}.
\newblock \bibinfo{publisher}{Springer}.
\newblock


\bibitem[Hemberg et~al\mbox{.}(2014)]%
        {bp-sepsis}
\bibfield{author}{\bibinfo{person}{Erik Hemberg}, \bibinfo{person}{Kalyan Veeramachaneni}, \bibinfo{person}{Babak Hodjat}, \bibinfo{person}{Prashan Wanigasekara}, \bibinfo{person}{Hormoz Shahrzad}, {and} \bibinfo{person}{Una-May O’Reilly}.} \bibinfo{year}{2014}\natexlab{}.
\newblock \showarticletitle{Learning Decision Lists with Lags for Physiological Time Series}. In \bibinfo{booktitle}{\emph{Third Workshop on Data Mining for Medicine and Healthcare, at the 14th SIAM International Conference on Data Mining}}. \bibinfo{pages}{82--87}.
\newblock


\bibitem[Hodjat and Shahrzad(2013)]%
        {Hodjat2013}
\bibfield{author}{\bibinfo{person}{Babak Hodjat} {and} \bibinfo{person}{Hormoz Shahrzad}.} \bibinfo{year}{2013}\natexlab{}.
\newblock \bibinfo{booktitle}{\emph{Introducing an Age-Varying Fitness Estimation Function}}.
\newblock \bibinfo{pages}{59--71}.
\newblock
\showISBNx{978-1-4614-6845-5}
\urldef\tempurl%
\url{https://doi.org/10.1007/978-1-4614-6846-2_5}
\showDOI{\tempurl}


\bibitem[Hodjat et~al\mbox{.}(2018)]%
        {Hodjat2018}
\bibfield{author}{\bibinfo{person}{Babak Hodjat}, \bibinfo{person}{Hormoz Shahrzad}, \bibinfo{person}{Risto Miikkulainen}, \bibinfo{person}{Lawrence Murray}, {and} \bibinfo{person}{Chris Holmes}.} \bibinfo{year}{2018}\natexlab{}.
\newblock \showarticletitle{{PRETSL}: Distributed Probabilistic Rule Evolution for Time-Series Classification}.
\newblock In \bibinfo{booktitle}{\emph{Genetic Programming Theory and Practice XIV}}. \bibinfo{publisher}{Springer}, \bibinfo{pages}{139--148}.
\newblock
\showISBNx{978-3-319-97088-2}


\bibitem[Jacques et~al\mbox{.}(2015)]%
        {MOCA}
\bibfield{author}{\bibinfo{person}{Julie Jacques}, \bibinfo{person}{Julien Taillard}, \bibinfo{person}{David Delerue}, \bibinfo{person}{Clarisse Dhaenens}, {and} \bibinfo{person}{Laetitia Jourdan}.} \bibinfo{year}{2015}\natexlab{}.
\newblock \showarticletitle{Conception of a dominance-based multi-objective local search in the context of classification rule mining in large and imbalanced data sets}.
\newblock \bibinfo{journal}{\emph{Applied Soft Computing}}  \bibinfo{volume}{34} (\bibinfo{year}{2015}), \bibinfo{pages}{705--720}.
\newblock


\bibitem[Lage et~al\mbox{.}(2019)]%
        {lage2019evaluation}
\bibfield{author}{\bibinfo{person}{Isaac Lage}, \bibinfo{person}{Emily Chen}, \bibinfo{person}{Jeffrey He}, \bibinfo{person}{Menaka Narayanan}, \bibinfo{person}{Been Kim}, \bibinfo{person}{Sam Gershman}, {and} \bibinfo{person}{Finale Doshi-Velez}.} \bibinfo{year}{2019}\natexlab{}.
\newblock \showarticletitle{An evaluation of the human-interpretability of explanation}.
\newblock \bibinfo{journal}{\emph{arXiv:1902.00006}} (\bibinfo{year}{2019}).
\newblock


\bibitem[Lakkaraju et~al\mbox{.}(2019)]%
        {MUSE}
\bibfield{author}{\bibinfo{person}{Himabindu Lakkaraju}, \bibinfo{person}{Ece Kamar}, \bibinfo{person}{Rich Caruana}, {and} \bibinfo{person}{Jure Leskovec}.} \bibinfo{year}{2019}\natexlab{}.
\newblock \showarticletitle{Faithful and Customizable Explanations of Black Box Models}. In \bibinfo{booktitle}{\emph{Proceedings of the 2019 AAAI/ACM Conference on AI, Ethics, and Society}} (Honolulu, HI, USA) \emph{(\bibinfo{series}{AIES '19})}. \bibinfo{publisher}{Association for Computing Machinery}, \bibinfo{address}{New York, NY, USA}, \bibinfo{pages}{131–138}.
\newblock
\showISBNx{9781450363242}


\bibitem[Linardatos et~al\mbox{.}(2020)]%
        {Explainable-AI}
\bibfield{author}{\bibinfo{person}{Pantelis Linardatos}, \bibinfo{person}{Vasilis Papastefanopoulos}, {and} \bibinfo{person}{Sotiris Kotsiantis}.} \bibinfo{year}{2020}\natexlab{}.
\newblock \showarticletitle{Explainable ai: A review of machine learning interpretability methods}.
\newblock \bibinfo{journal}{\emph{Entropy}} \bibinfo{volume}{23}, \bibinfo{number}{1} (\bibinfo{year}{2020}), \bibinfo{pages}{18}.
\newblock


\bibitem[Lundberg and Lee(2017)]%
        {shap}
\bibfield{author}{\bibinfo{person}{Scott~M. Lundberg} {and} \bibinfo{person}{Su-In Lee}.} \bibinfo{year}{2017}\natexlab{}.
\newblock \showarticletitle{A Unified Approach to Interpreting Model Predictions}. In \bibinfo{booktitle}{\emph{Proceedings of the 31st International Conference on Neural Information Processing Systems}}. Curran Associates Inc.
\newblock


\bibitem[Miikkulainen(2023)]%
        {miikkulainen:emlchapter23}
\bibfield{author}{\bibinfo{person}{Risto Miikkulainen}.} \bibinfo{year}{2023}\natexlab{}.
\newblock \showarticletitle{Evolutionary Supervised Machine Learning}.
\newblock In \bibinfo{booktitle}{\emph{Handbook of Evolutionary Machine Learning}}, \bibfield{editor}{\bibinfo{person}{W.~Banzhaf}, \bibinfo{person}{P.~Machado}, {and} \bibinfo{person}{M.~Zhang}} (Eds.). \bibinfo{publisher}{Springer}, \bibinfo{address}{New York}.
\newblock


\bibitem[Miikkulainen et~al\mbox{.}(2021)]%
        {miikkulainen:ieeetec21}
\bibfield{author}{\bibinfo{person}{Risto Miikkulainen}, \bibinfo{person}{Olivier Francon}, \bibinfo{person}{Elliot Meyerson}, \bibinfo{person}{Xin Qiu}, \bibinfo{person}{Darren Sargent}, \bibinfo{person}{Elisa Canzani}, {and} \bibinfo{person}{Babak Hodjat}.} \bibinfo{year}{2021}\natexlab{}.
\newblock \showarticletitle{From Prediction to Prescription: Evolutionary Optimization of Non-Pharmaceutical Interventions in the {COVID-19} Pandemic}.
\newblock \bibinfo{journal}{\emph{IEEE Transactions on Evolutionary Computation}}  \bibinfo{volume}{25} (\bibinfo{year}{2021}), \bibinfo{pages}{386--401}.
\newblock


\bibitem[Miikkulainen et~al\mbox{.}(2017)]%
        {miikkulainen:ssci17}
\bibfield{author}{\bibinfo{person}{Risto Miikkulainen}, \bibinfo{person}{Hormoz Shahrzad}, \bibinfo{person}{Nigel Duffy}, {and} \bibinfo{person}{Phil Long}.} \bibinfo{year}{2017}\natexlab{}.
\newblock \showarticletitle{How to Select a Winner in Evolutionary Optimization?}. In \bibinfo{booktitle}{\emph{Proceedings of the IEEE Symposium Series in Computational Intelligence}}. \bibinfo{publisher}{IEEE}.
\newblock
\urldef\tempurl%
\url{http://www.cs.utexas.edu/users/ai-lab?miikkulainen:ssci17}
\showURL{%
\tempurl}


\bibitem[Miller(2019)]%
        {miller2019}
\bibfield{author}{\bibinfo{person}{Tim Miller}.} \bibinfo{year}{2019}\natexlab{}.
\newblock \showarticletitle{Explanation in artificial intelligence: Insights from the social sciences}.
\newblock \bibinfo{journal}{\emph{Artificial intelligence}}  \bibinfo{volume}{267} (\bibinfo{year}{2019}), \bibinfo{pages}{1--38}.
\newblock


\bibitem[Nauta et~al\mbox{.}(2022)]%
        {Nauta2022FromAE}
\bibfield{author}{\bibinfo{person}{Meike Nauta}, \bibinfo{person}{Jan Trienes}, \bibinfo{person}{Shreyasi Pathak}, \bibinfo{person}{Elisa Nguyen}, \bibinfo{person}{Michelle Peters}, \bibinfo{person}{Yasmin Schmitt}, \bibinfo{person}{J{\"o}rg Schl{\"o}tterer}, \bibinfo{person}{Maurice van Keulen}, {and} \bibinfo{person}{Christin Seifert}.} \bibinfo{year}{2022}\natexlab{}.
\newblock \showarticletitle{From Anecdotal Evidence to Quantitative Evaluation Methods: A Systematic Review on Evaluating Explainable AI}.
\newblock \bibinfo{journal}{\emph{Comput. Surveys}}  \bibinfo{volume}{55} (\bibinfo{year}{2022}), \bibinfo{pages}{1 -- 42}.
\newblock
\urldef\tempurl%
\url{https://api.semanticscholar.org/CorpusID:246063780}
\showURL{%
\tempurl}


\bibitem[O'Neill and Ryan(1998)]%
        {oneill1998grammatical}
\bibfield{author}{\bibinfo{person}{Michael O'Neill} {and} \bibinfo{person}{Conor Ryan}.} \bibinfo{year}{1998}\natexlab{}.
\newblock \showarticletitle{Grammatical evolution: Evolving programs for an arbitrary language}. In \bibinfo{booktitle}{\emph{Genetic Programming}}. Springer, \bibinfo{pages}{83--96}.
\newblock


\bibitem[Pe{\~n}arroya(2004)]%
        {Pearroya2004PittsburghGM}
\bibfield{author}{\bibinfo{person}{Jaume~Bacardit Pe{\~n}arroya}.} \bibinfo{year}{2004}\natexlab{}.
\newblock \showarticletitle{Pittsburgh genetic-based machine learning in the data mining era: representations, generalization, and run-time}.
\newblock


\bibitem[Poghosyan et~al\mbox{.}(2021)]%
        {RIPPER}
\bibfield{author}{\bibinfo{person}{Arnak Poghosyan}, \bibinfo{person}{Ashot Harutyunyan}, \bibinfo{person}{Naira Grigoryan}, {and} \bibinfo{person}{Nicholas Kushmerick}.} \bibinfo{year}{2021}\natexlab{}.
\newblock \showarticletitle{Incident Management for Explainable and Automated Root Cause Analysis in Cloud Data Centers}.
\newblock \bibinfo{journal}{\emph{Journal of Universal Computer Science}} \bibinfo{volume}{27}, \bibinfo{number}{11} (\bibinfo{year}{2021}), \bibinfo{pages}{1152--1173}.
\newblock


\bibitem[Ribeiro et~al\mbox{.}(2016)]%
        {LIME}
\bibfield{author}{\bibinfo{person}{Marco~Tulio Ribeiro}, \bibinfo{person}{Sameer Singh}, {and} \bibinfo{person}{Carlos Guestrin}.} \bibinfo{year}{2016}\natexlab{}.
\newblock \showarticletitle{"Why Should I Trust You?": Explaining the Predictions of Any Classifier}. In \bibinfo{booktitle}{\emph{Proceedings of the 22nd ACM SIGKDD International Conference on Knowledge Discovery and Data Mining}} (San Francisco, California, USA) \emph{(\bibinfo{series}{KDD '16})}. \bibinfo{publisher}{Association for Computing Machinery}, \bibinfo{address}{New York, NY, USA}, \bibinfo{pages}{1135–1144}.
\newblock
\showISBNx{9781450342322}


\bibitem[Schmidt and Lipson(2009)]%
        {lipson}
\bibfield{author}{\bibinfo{person}{Michael Schmidt} {and} \bibinfo{person}{Hod Lipson}.} \bibinfo{year}{2009}\natexlab{}.
\newblock \showarticletitle{Distilling Free-Form Natural Laws from Experimental Data}.
\newblock \bibinfo{journal}{\emph{Science}} \bibinfo{volume}{324}, \bibinfo{number}{5923} (\bibinfo{year}{2009}), \bibinfo{pages}{81--85}.
\newblock


\bibitem[Schwalbe and Finzel(2021)]%
        {Schwalbe2021ACT}
\bibfield{author}{\bibinfo{person}{Gesina Schwalbe} {and} \bibinfo{person}{Bettina Finzel}.} \bibinfo{year}{2021}\natexlab{}.
\newblock \showarticletitle{A comprehensive taxonomy for explainable artificial intelligence: {A} systematic survey of surveys on methods and concepts}.
\newblock \bibinfo{journal}{\emph{Data Mining and Knowledge Discovery}} (\bibinfo{year}{2021}), \bibinfo{pages}{1--59}.
\newblock


\bibitem[Schwarz(1978)]%
        {schwarz1978}
\bibfield{author}{\bibinfo{person}{G. Schwarz}.} \bibinfo{year}{1978}\natexlab{}.
\newblock \showarticletitle{Estimating the dimension of a model}.
\newblock \bibinfo{journal}{\emph{The Annals of Statistics}} \bibinfo{volume}{6}, \bibinfo{number}{2} (\bibinfo{year}{1978}), \bibinfo{pages}{461--464}.
\newblock


\bibitem[Shahrzad et~al\mbox{.}(2018)]%
        {shahrzad:alife18}
\bibfield{author}{\bibinfo{person}{Hormoz Shahrzad}, \bibinfo{person}{Daniel Fink}, {and} \bibinfo{person}{Risto Miikkulainen}.} \bibinfo{year}{2018}\natexlab{}.
\newblock \showarticletitle{Enhanced Optimization with Composite Objectives and Novelty Selection}. In \bibinfo{booktitle}{\emph{Proceedings of the 2018 Conference on Artificial Life}}. \bibinfo{address}{Tokyo, Japan}.
\newblock
\urldef\tempurl%
\url{http://www.cs.utexas.edu/users/ai-lab?shahrzad:alife18}
\showURL{%
\tempurl}


\bibitem[Shahrzad et~al\mbox{.}(2020)]%
        {shahrzad:gptp20}
\bibfield{author}{\bibinfo{person}{Hormoz Shahrzad}, \bibinfo{person}{Babak Hodjat}, \bibinfo{person}{Camille Dolle}, \bibinfo{person}{Andrej Denissov}, \bibinfo{person}{Simon Lau}, \bibinfo{person}{Donn Goodhew}, \bibinfo{person}{Justin Dyer}, {and} \bibinfo{person}{Risto Miikkulainen}.} \bibinfo{year}{2020}\natexlab{}.
\newblock \showarticletitle{Enhanced Optimization with Composite Objectives and Novelty Pulsation}.
\newblock In \bibinfo{booktitle}{\emph{Genetic Programming Theory and Practice XVII}}, \bibfield{editor}{\bibinfo{person}{Wolfgang Banzhaf}, \bibinfo{person}{Erik Goodman}, \bibinfo{person}{Leigh Sheneman}, \bibinfo{person}{Leonardo Trujillo}, {and} \bibinfo{person}{Bill Worzel}} (Eds.). \bibinfo{publisher}{Springer}, \bibinfo{address}{New York}, \bibinfo{pages}{275--293}.
\newblock


\bibitem[Shahrzad et~al\mbox{.}(2016)]%
        {Shahrzad2016}
\bibfield{author}{\bibinfo{person}{Hormoz Shahrzad}, \bibinfo{person}{Babak Hodjat}, {and} \bibinfo{person}{Risto Miikkulainen}.} \bibinfo{year}{2016}\natexlab{}.
\newblock \showarticletitle{Estimating the Advantage of Age-Layering in Evolutionary Algorithms}. \bibinfo{pages}{693--699}.
\newblock
\urldef\tempurl%
\url{https://doi.org/10.1145/2908812.2908911}
\showDOI{\tempurl}


\bibitem[Srinivasan and Ramakrishnan(2011)]%
        {Srinivasan2011}
\bibfield{author}{\bibinfo{person}{Sujatha Srinivasan} {and} \bibinfo{person}{Sivakumar Ramakrishnan}.} \bibinfo{year}{2011}\natexlab{}.
\newblock \showarticletitle{Evolutionary multi objective optimization for rule mining: {A} review}.
\newblock \bibinfo{journal}{\emph{Artificial Intelligence Review}}  \bibinfo{volume}{36} (\bibinfo{date}{10} \bibinfo{year}{2011}), \bibinfo{pages}{205--248}.
\newblock
\urldef\tempurl%
\url{https://doi.org/10.1007/s10462-011-9212-3}
\showDOI{\tempurl}


\bibitem[Strack et~al\mbox{.}(2014)]%
        {diabetes}
\bibfield{author}{\bibinfo{person}{Beata Strack}, \bibinfo{person}{Jonathan Deshazo}, \bibinfo{person}{Chris Gennings}, \bibinfo{person}{Juan~Luis Olmo~Ortiz}, \bibinfo{person}{Sebastian Ventura}, \bibinfo{person}{Krzysztof Cios}, {and} \bibinfo{person}{John Clore}.} \bibinfo{year}{2014}\natexlab{}.
\newblock \showarticletitle{Impact of HbA1c Measurement on Hospital Readmission Rates: Analysis of 70,000 Clinical Database Patient Records}.
\newblock \bibinfo{journal}{\emph{BioMed Research International}}  \bibinfo{volume}{2014} (\bibinfo{year}{2014}), \bibinfo{pages}{781670}.
\newblock


\bibitem[Tasfi(2016)]%
        {tasfi2016PLE}
\bibfield{author}{\bibinfo{person}{Norman Tasfi}.} \bibinfo{year}{2016}\natexlab{}.
\newblock \bibinfo{title}{PyGame Learning Environment}.
\newblock \bibinfo{howpublished}{\url{https://github.com/ntasfi/PyGame-Learning-Environment}}.
\newblock


\bibitem[Tjoa and Guan(2021)]%
        {Tjoa2021ASO}
\bibfield{author}{\bibinfo{person}{Erico Tjoa} {and} \bibinfo{person}{Cuntai Guan}.} \bibinfo{year}{2021}\natexlab{}.
\newblock \showarticletitle{A Survey on Explainable Artificial Intelligence (XAI): Toward Medical XAI}.
\newblock \bibinfo{journal}{\emph{IEEE Transactions on Neural Networks and Learning Systems}}  \bibinfo{volume}{32} (\bibinfo{year}{2021}), \bibinfo{pages}{4793--4813}.
\newblock


\bibitem[Touvron et~al\mbox{.}(2023)]%
        {touvron:arxiv23}
\bibfield{author}{\bibinfo{person}{Hugo Touvron}, \bibinfo{person}{Thibaut Lavril}, \bibinfo{person}{Gautier Izacard}, \bibinfo{person}{Xavier Martinet}, \bibinfo{person}{Marie-Anne Lachaux}, \bibinfo{person}{Timothée Lacroix}, \bibinfo{person}{Baptiste Rozière}, \bibinfo{person}{Naman Goyal}, \bibinfo{person}{Eric Hambro}, \bibinfo{person}{Faisal Azhar}, \bibinfo{person}{Aurelien Rodriguez}, \bibinfo{person}{Armand Joulin}, \bibinfo{person}{Edouard Grave}, {and} \bibinfo{person}{Guillaume Lample}.} \bibinfo{year}{2023}\natexlab{}.
\newblock \showarticletitle{LLaMA: Open and Efficient Foundation Language Models}.
\newblock \bibinfo{journal}{\emph{arXiv:2302.13971}} (\bibinfo{year}{2023}).
\newblock


\bibitem[Townsend et~al\mbox{.}(2019)]%
        {explain-deepNN}
\bibfield{author}{\bibinfo{person}{Joseph Townsend}, \bibinfo{person}{Thomas Chaton}, {and} \bibinfo{person}{Joao~M Monteiro}.} \bibinfo{year}{2019}\natexlab{}.
\newblock \showarticletitle{Extracting relational explanations from deep neural networks: A survey from a neural-symbolic perspective}.
\newblock \bibinfo{journal}{\emph{IEEE transactions on neural networks and learning systems}} \bibinfo{volume}{31}, \bibinfo{number}{9} (\bibinfo{year}{2019}), \bibinfo{pages}{3456--3470}.
\newblock


\end{thebibliography}

\newpage
\appendix

\section{Implementation of initialization, mutation, crossover, and pruning}
\label{ap:algos}

Algorithm~\ref{alg:generate_random_ruleset} describes how rule sets are randomly initialized in the beginning of evolution; Algorithms~\ref{alg:crossover} through~\ref{alg:combine_rules} describe how rule sets are crossed over; Algorithms~\ref{alg:mutate_condition} through~\ref{alg:mutate_ruleset} specify how individual rules are changed through mutation; Algorithm~\ref{alg:combine_rules} implements the novel method of combining conditions and actions between two rule sets. Algorithm~\ref{alg:prune} illustrates how the rule sets are simplified through pruning based on the $\mathit{times\_applied}$ counters. 

\begin{algorithm}
\caption{Generate Random Rule Set}
\label{alg:generate_random_ruleset}
\begin{algorithmic}[1]
\Require $\textit{max\_rules}$: Maximum number of building block rules
\Require $\textit{max\_conditions}$: Maximum number of building block conditions per rule
\Require $\textit{action\_set}$: Set of possible domain actions
\Ensure Returns a randomly generated RuleSet

\Function{GenerateRandomRuleSet}{$max\_rules$, $max\_conditions$, $action\_set$}
    \State $\textit{ruleSet} \gets \text{new RuleSet}()$
    \State $\textit{numRules} \gets \Call{Random}{1, max\_rules}$
    \For{$i \gets 1$ \textbf{to} $\textit{numRules}$}
        \State $\textit{rule} \gets \text{new Rule}()$
        \State $\textit{numConditions} \gets \Call{Random}{1, max\_conditions}$
        \For{$j \gets 1$ \textbf{to} $\textit{numConditions}$}
            \State $\textit{condition} \gets \text{new Condition}()$
            \State Create a $\textit{condition}$ with random parameters
            \State $\Call{AddCondition}{\textit{rule}, \textit{condition}}$
        \EndFor
        \State $rule.action \gets \Call{RandomChoice}{\text{action\_set}}$
        \State $\Call{AddRule}{\textit{ruleSet}, \textit{rule}}$
    \EndFor
    \State \Return $\textit{ruleSet}$
\EndFunction
\end{algorithmic}
\end{algorithm}

\begin{algorithm}
\caption{Rule-Set Crossover}
\label{alg:crossover}
\begin{algorithmic}[1]
\Require $parent_1$, $parent_2$: The two-parent RuleSets
\Require $\mathit{config}$: Hyper-parameters for crossover
\Ensure Returns an $\mathit{offspring}$ RuleSet derived from two parents

\Function{RuleSetCrossover}{$parent_1$, $parent_2$}
    \State Create an empty $\mathit{offspring}$ RuleSet
    \State Filter inactive rules from $parent_1$ and $parent_2$ \Comment{Keep only rules with {\footnotesize $\mathit{times\_applied}>0$}}
    \State $crossoverStrategy \gets \Call{RandomChoice}{\{\text{'single\_point'}, \text{'uniform'}, \text{'combine\_rules'}\}}$
    
    \If{$crossoverStrategy$ == 'single\_point'}
        \State $\mathit{offspring} \gets \Call{SinglePointCrossover}{parent_1, parent_2}$
    \ElsIf{$crossoverStrategy$ == 'uniform'}
        \State $\mathit{offspring} \gets \Call{UniformCrossover}{parent_1, parent_2}$
    \Else \Comment{crossoverStrategy is 'combine\_rules'}
        \State $\mathit{offspring} \gets \Call{CombineRules}{parent_1, parent_2}$
    \EndIf
    
    \If{$\mathit{offspring}$ is invalid or empty}
        \State $\mathit{offspring} \gets \text{randomly generated RuleSet}$
    \EndIf
    
    \State $\mathit{offspring} \gets \Call{PruneRules}{\mathit{offspring}}$ \Comment{Remove redundant rules from $\mathit{offspring}$}
    \State \Return $\mathit{offspring}$
\EndFunction
\end{algorithmic}
\end{algorithm}

\begin{algorithm}
\caption{Single-Point Crossover}
\label{alg:single_point_crossover}
\begin{algorithmic}[1]
\Require $parent_1$, $parent_2$: The two-parent RuleSets
\Require $\mathit{config}$: Hyper-parameters for crossover
\Ensure Returns an $\mathit{offspring}$ RuleSet derived through single-point crossover
\Function{SinglePointCrossover}{$parent_1$, $parent_2$}
    \State $crossoverIndex \gets \text{A random number between 1 and }\min(\Call{Size}{parent_1}, \Call{Size}{parent_2})$
    \State $maxLength \gets $
    \State Create an empty $\mathit{offspring}$ RuleSet
    \For{$i \gets 1$ \textbf{to} $crossoverIndex$}
        \State Append $parent_1[i]$ to $\mathit{offspring}$
    \EndFor
    \For{$i \gets crossoverIndex + 1$ \textbf{to} $\Call{Size}{parent_2}$}
        \State Append $parent_2[i]$ to $\mathit{offspring}$
    \EndFor
    \State \Return $\mathit{offspring}$
\EndFunction
\end{algorithmic}
\end{algorithm}

\begin{algorithm}
\caption{Uniform Crossover}
\label{alg:uniform_crossover}
\begin{algorithmic}[1]
\Require $parent_1$, $parent_2$: The two-parent RuleSets
\Require $\mathit{config}$: Hyper-parameters for crossover
\Ensure Returns an $\mathit{offspring}$ RuleSet derived through uniform crossover
\Function{UniformCrossover}{$parent_1$, $parent_2$}
    \State $maxLength \gets \max(\Call{Size}{parent_1}, \Call{Size}{parent_2})$
    \State Create an empty $\mathit{offspring}$ RuleSet
    \For{$i \gets 1$ \textbf{to} $maxLength$}
        \If{$i \leq \Call{Size}{parent_1}$ \textbf{and} $i \leq \Call{Size}{parent_2}$}
            \State $\mathit{offspring}Rule \gets \Call{RandomChoice}{\{parent_1[i], parent_2[i]\}}$
        \ElsIf{$i \leq \Call{Size}{parent_1}$}
            \State $\mathit{offspring}Rule \gets parent_1[i]$
        \Else
            \State $\mathit{offspring}Rule \gets parent_2[i]$
        \EndIf
        \State Append $\mathit{offspring}Rule$ to $\mathit{offspring}$
    \EndFor
    \State \Return $\mathit{offspring}$
\EndFunction
\end{algorithmic}
\end{algorithm}

\begin{algorithm}
\caption{Combine-Rules Crossover}
\label{alg:combine_rules}
\begin{algorithmic}[1]
\Require $parent_1$, $parent_2$: The two-parent RuleSets
\Ensure Returns an $\mathit{offspring}$ RuleSet, enhancing $parent_2$ with conditions from $parent_1$
\Function{CombineRules}{$parent_1$, $parent_2$}
    \State $\mathit{offspring} \gets \text{Clone(}parent_2\text{)}$ \Comment{Initialize $\mathit{offspring}$ as a copy of $parent_2$}
    \State $selectedRule \gets \Call{RandomSelect}{parent_1}$ \Comment{Randomly select a rule from $parent_1$}
    \State $selectedAction \gets \Call{GetAction}{selectedRule}$ \Comment{Extract action of the selected rule}
    \ForAll{$\mathit{rule}$ in $\mathit{offspring}$}
        \If{\Call{GetAction}{$\mathit{rule}$} == $selectedAction$}
            \State \Call{AppendConditionsFrom}{$\mathit{rule}$, $selectedRule$} \Comment{Append conditions to rules in} \\\Comment{$\mathit{offspring}$ with the same action}
        \EndIf
    \EndFor
    \State \Return $\mathit{offspring}$
\EndFunction
\algnewcommand\algorithmicforeach{\textbf{for each}}
\algdef{S}[FOR]{ForAll}[1]{\algorithmicforeach\ #1\ \algorithmicdo}
\end{algorithmic}
\end{algorithm}

\begin{algorithm}[!h]
\caption{Mutate Conditions}
\label{alg:mutate_condition}
\begin{algorithmic}[1]
\Require $\mathit{condition}$: The Condition object to be mutated
\Require $\mathit{config}$: Hyper-parameters for condition mutation
\Require $\mathit{states}$: Dictionary of possible input features
\Ensure Returns a mutated Condition object
\Function{MutateCondition}{$\mathit{condition}$}
    \State Randomly determine mutation type based on $\mathit{config}$
    \If{selected mutation is "perturbation"}
        \State Perturb the value of $\mathit{condition}$
    \ElsIf{selected mutation is changing "operator"}
        \State Randomly change the "operator" of $\mathit{condition}$
    \ElsIf{selected mutation is modifying the "lag"}
        \State Randomly modify the lag of $\mathit{condition}$
    \ElsIf{selected mutation is changing $\mathit{states}$}
        \State Randomly change the $\mathit{states}$ involved in $\mathit{condition}$
    \EndIf
    \State Apply any necessary adjustments to maintain logical consistency
    \State \Return the mutated $\mathit{condition}$
\EndFunction
\end{algorithmic}
\end{algorithm}

\begin{algorithm}[!h]
\caption{Mutate Rule}
\label{alg:mutate_rule}
\begin{algorithmic}[1]
\Require $\mathit{rule}$: The Rule object to be mutated
\Require $\mathit{config}$: Hyper-parameters for rule mutation
\Require $\mathit{actions}$: Dictionary of possible actions or classes
\Ensure Returns a mutated Rule object
\Function{MutateRule}{$\mathit{rule}$}
    \State Randomly select mutation aspect based on $\mathit{config}$
    \If{selected mutation is changing "action"}
        \State $\mathit{rule.action} \gets \Call{MutateAction}{\mathit{rule.action}}$
    \ElsIf{selected mutation is changing "certainty"}
        \State $\mathit{rule} \gets \Call{MutateCertainty}{\mathit{rule}}$
    \ElsIf{selected mutation of a "condition"}
        \State Randomly determine specific condition mutation strategy based on $\mathit{config}$
        \If{selected condition strategy is in place mutation}
            \State $\mathit{condition} \gets \Call{PickRandomCondition}{\mathit{rule}}$
            \Comment{Picks a random $\mathit{condition}$ from $\mathit{rule}$}
            \State $\mathit{rule} \gets \Call{MutateCondition}{\mathit{rule, condition}}$
        \ElsIf{selected condition strategy is removal $\And$ multiple conditions exist}
            \State $\mathit{condition} \gets \Call{PickRandomCondition}{\mathit{rule}}$
            \Comment{Picks a random $\mathit{condition}$ from $\mathit{rule}$}
            \State $\mathit{rule} \gets \Call{RemoveCondition}{\mathit{rule, condition}}$
        \Else
            \State $\mathit{condition} \gets \Call{CreateRandomCondition}{ }$
            \State $\mathit{rule} \gets \Call{AddCondition}{\mathit{condition}}$
        \EndIf
    \EndIf
    \State Apply any necessary adjustments to maintain logical consistency
    \State \Return $\mathit{rule}$
    \EndFunction
\end{algorithmic}
\end{algorithm}

\begin{algorithm}[!h]
\caption{Mutate Rule Set}
\label{alg:mutate_ruleset}
\begin{algorithmic}[1]
\Require $\mathit{ruleset}$: The RuleSet object to be mutated
\Require $\mathit{config}$: Hyper-parameters for ruleset mutation
\Require $\mathit{actions}$: Dictionary of possible actions or classes
\Ensure Returns a mutated RuleSet object
\Function{MutateRuleSet}{$\mathit{ruleset}$}
    \State Randomly select mutation type based on $\mathit{config}$
    \If{selected mutation is "addition"}
        \State Create a random new rule with actions from $\mathit{actions}$
        \State Append the new rule to $\mathit{ruleset}$
    \ElsIf{selected mutation is "removal"}
        \If{$\mathit{ruleset}$ has more than one rule}
            \State Randomly remove an existing rule from $\mathit{ruleset}$
        \EndIf
    \ElsIf{selected mutation is "modify default action"}
        \State Randomly select a new action from $\mathit{actions}$
        \State Set this action as the new default action for $\mathit{ruleset}$
    \ElsIf{selected mutation is "reorder"}
        \State Randomly shuffle the order of rules in $\mathit{ruleset}$
    \EndIf
    \State \Return $\mathit{ruleset}$
\EndFunction
\end{algorithmic}
\end{algorithm}

\begin{algorithm}
\caption{Prune Redundant Rules and Conditions}
\label{alg:prune}
\begin{algorithmic}[1]
\Require $\mathit{rules}$: A list of Rule objects
\Ensure $\mathit{rules}$ is pruned of redundant rules, redundant conditions, tautologies, and falsehoods
\Function{PruneRules}{$\mathit{rules}$}
    \For{each $\mathit{rule}$ in $\mathit{rules}$}
        \State Identify and remove any tautological or false conditions within $\mathit{rule}$
        \For{each $\mathit{condition}$ in $rule.conditions$}
            \If{$\mathit{condition}$ is redundant or a tautology}
                \State Remove $\mathit{condition}$ from $\mathit{rule}$
            \EndIf
        \EndFor
        \If{$\mathit{rule}$ is now redundant or contains only tautologies or falsehoods}
            \State Remove $\mathit{rule}$ from $rule\_set$
        \EndIf
    \EndFor
    \For{each pair of $rule_i, rule_j$ in $\mathit{rules}$}
        \If{$\mathit{rule_i}$ and $\mathit{rule_j}$ are logically equivalent}
            \State Remove $\mathit{rule_j}$ from $\mathit{rules}$
        \EndIf
    \EndFor
    \State \Return $\mathit{rules}$ \Comment{Return the pruned set of rules}
    \EndFunction
\end{algorithmic}
\end{algorithm}

\clearpage
\section{Implementation of the AIC and BIC metrics}
\label{ap:aicbic}

Algorithm~\ref{alg:aic_bic_calc} specifies the simplified calculation of AIC and BIC for the special conditions of the experiments: The classifiers always predict one class with certainty (probability of one) and assign a probability of zero to all other classes. Under these conditions, AIC and BIC can be calculated using the model's accuracy. Further, to make the method practical, a small value (1e-15) is used for the logarithm of zero (which is mathematically undefined), similarly to the Scikit-learn library's $\mathit{log\_loss}$ function (found in $\mathit{sklearn.metrics.log\_loss}$).

\begin{algorithm}
\caption{Calculate AIC and BIC under the special conditions of the experiments}
\label{alg:aic_bic_calc}
\begin{algorithmic}[1]
\Procedure{CalculateAICBIC}{$accuracy$, $n\_parameters$, $n\_data\_points$}
    \State $n\_incorrect \gets (1 - accuracy) \times n\_data\_points$
    \State $small\_prob \gets 1 \times 10^{-15}$
    \State $log\_likelihood \gets n\_incorrect \times \log(small\_prob)$
    \State $AIC \gets 2 \times n\_parameters - 2 \times log\_likelihood$
    \State $BIC \gets \log(n\_data\_points) \times n\_parameters - 2 \times log\_likelihood$
    \State \textbf{return} $AIC$, $BIC$
\EndProcedure
\end{algorithmic}
\end{algorithm}

\clearpage
\section{Balancing accuracy vs.\ complexity}
\label{ap:tradeoffs}

\begin{figure*}[!ht]
  \centering
  \hfill
  \begin{minipage}[t]{0.48\linewidth}
  \centering
  \includegraphics[width=\linewidth]{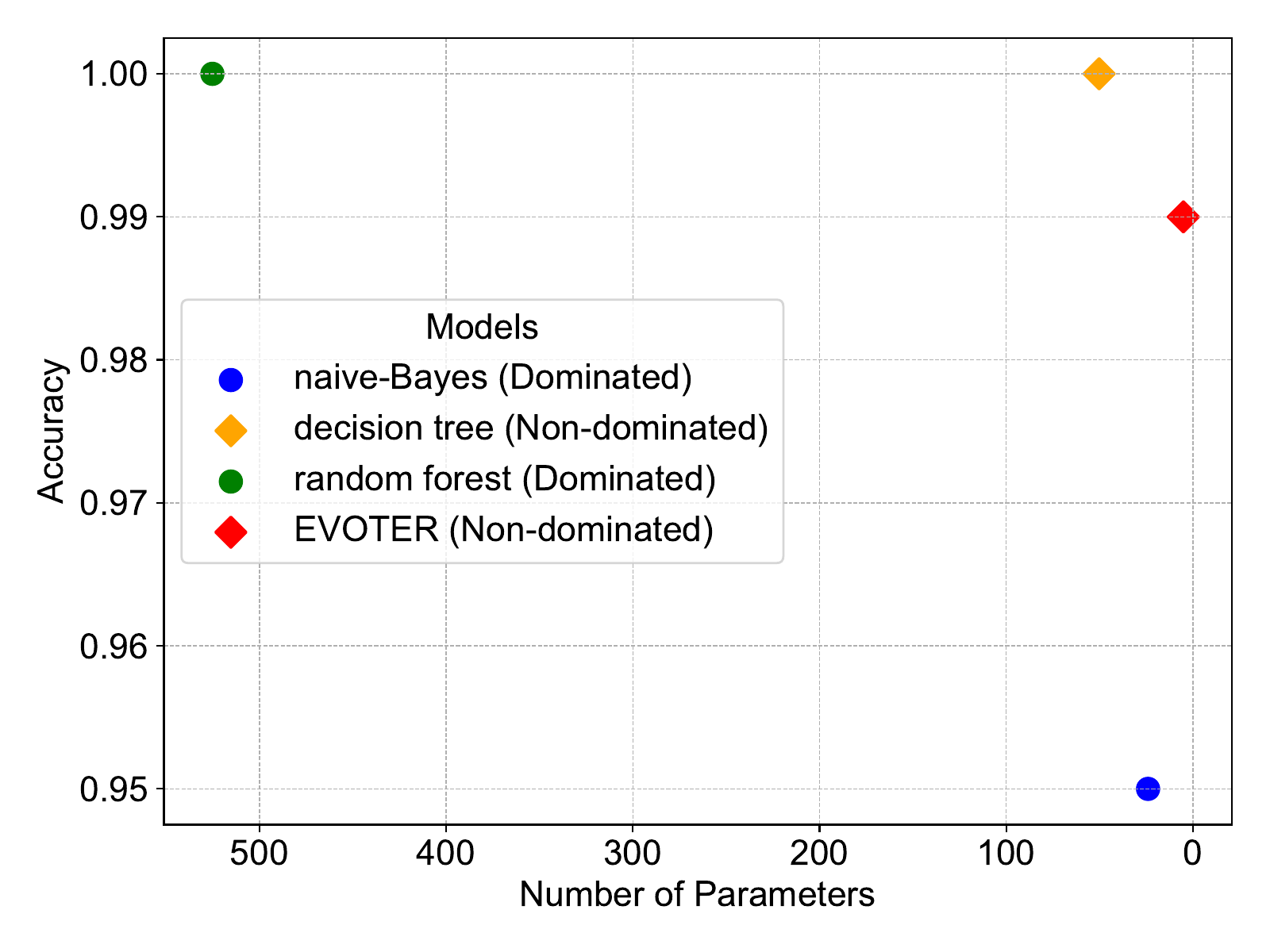}\\[-2ex]
  {\footnotesize ($a$) Iris}
  \end{minipage}
  \hfill
  \begin{minipage}[t]{0.48\linewidth}
  \centering
  \includegraphics[width=\linewidth]{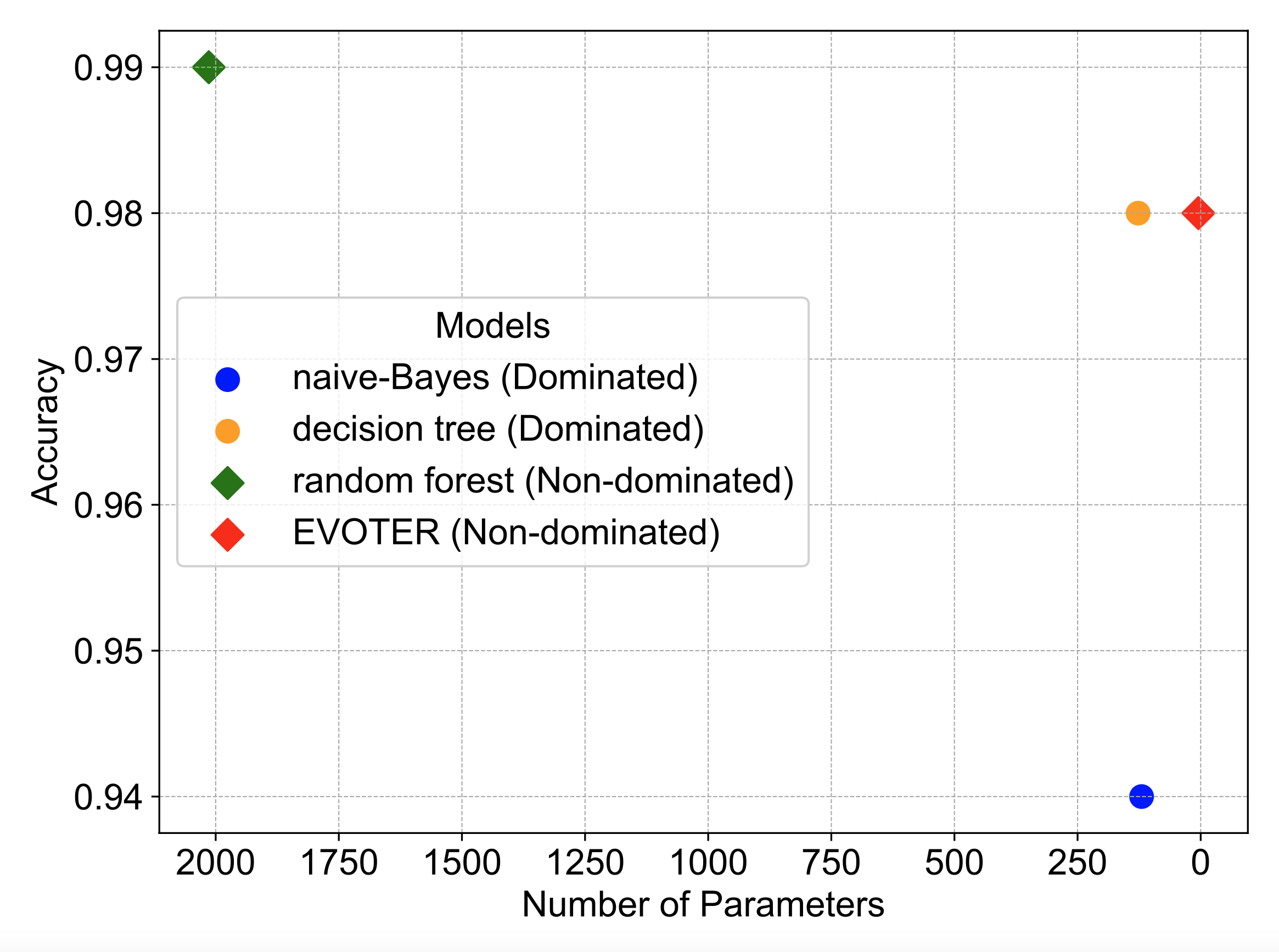}\\[-2ex]
  {\footnotesize ($b$) Breast Cancer}
  \end{minipage}
  \hfill
  \begin{minipage}[t]{0.48\linewidth}
  \centering
  \includegraphics[width=\linewidth]{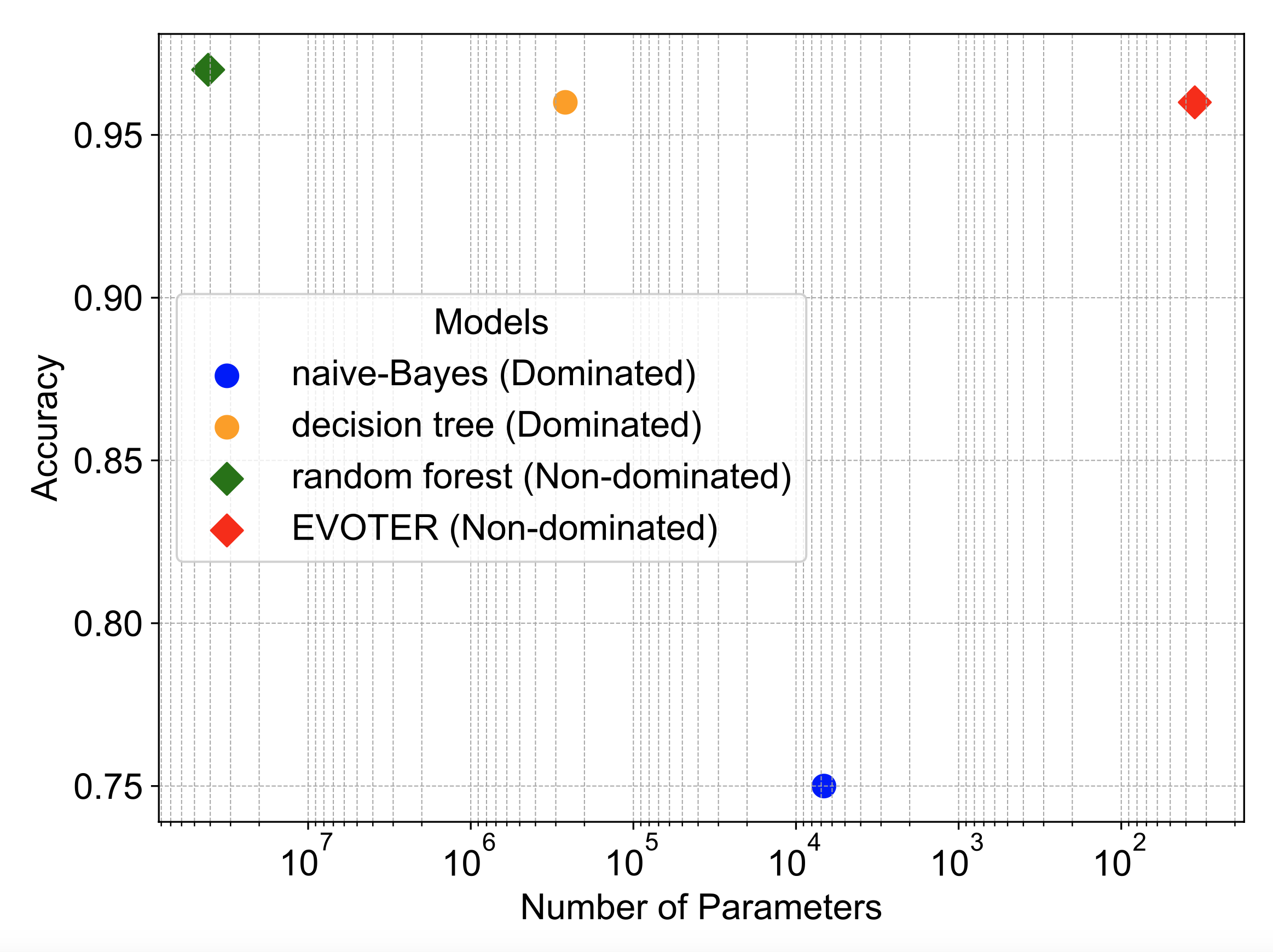}\\[-2ex]
  {\footnotesize ($c$) Human Activity (Static Features)}
  \end{minipage}
  \hfill
  \begin{minipage}[t]{0.48\linewidth}
  \centering
  \includegraphics[width=\linewidth]{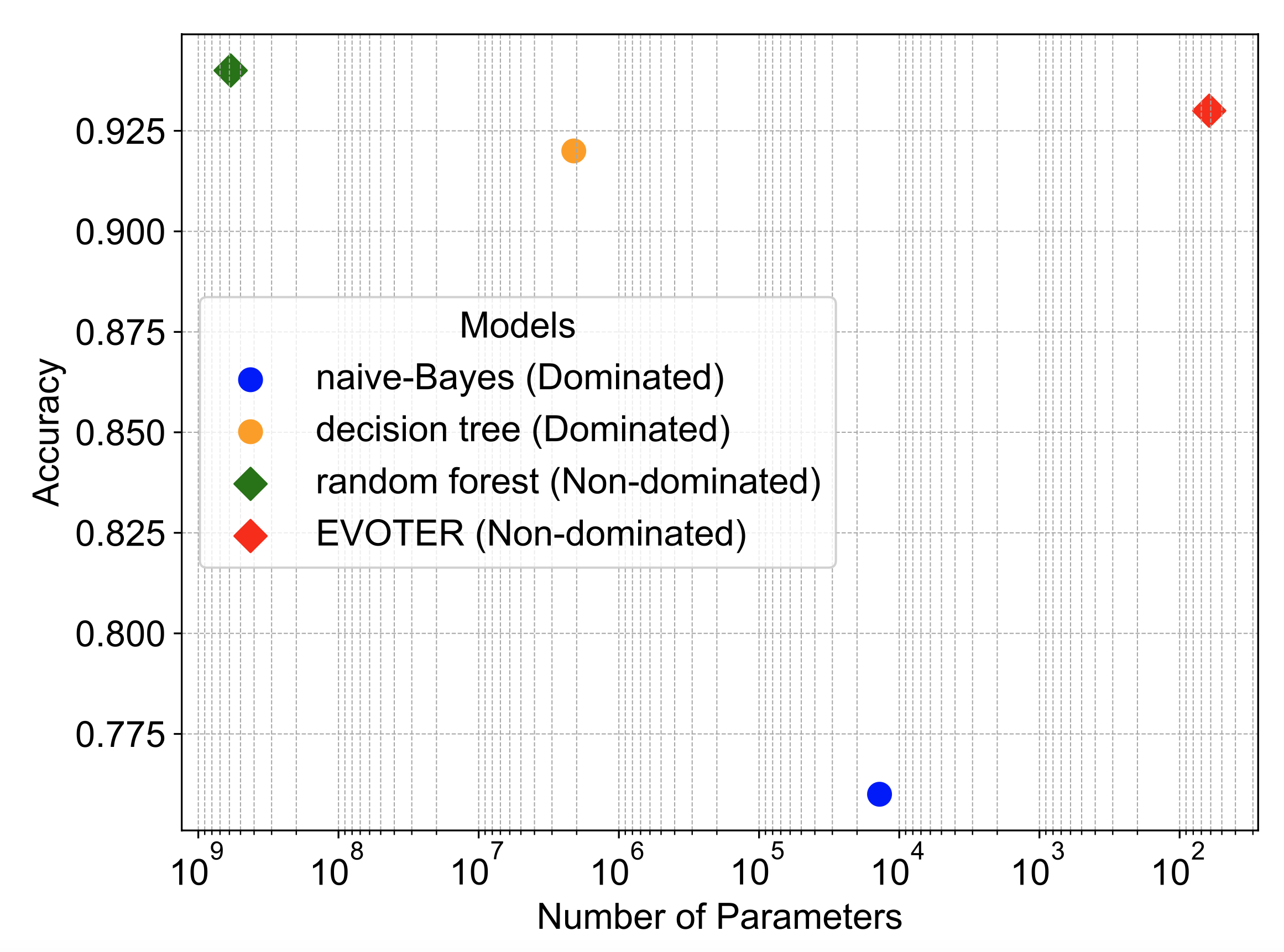}\\[-2ex]
  {\footnotesize ($d$) Human Activity (Time-Series Features)}
  \end{minipage}
  \hfill\mbox{}
\caption{\emph{Comparison of accuracy vs.\ complexity trade-offs.} Each point represents the average of ten runs as shown in Tables~\ref{tab:model_comparison_iris}~--~\ref{tab:model_comparison_activity_raw}.  In each dataset, the EVOTER solution is non-dominated by the other solutions, that is, EVOTER is either to the right or above the others, or both. It therefore finds a superior balance between accuracy and simplicity.}
\label{fig:Pareto}
\end{figure*}

Accuracy and complexity are conflicting properties in that accuracy can often be improved by making the model more complex. It is therefore illuminating to show the trade-offs that the different methods pick along these metrics.  As illustrated in Figure~\ref{fig:Pareto}, EVOTER discovers non-dominated solutions in every domain. Although in these four datasets non-dominance follows immediately because EVOTER is no less accurate and always much simpler, this result suggests even in domains where accuracy may be lower when the models are simpler, EVOTER's solutions may provide a useful and practical tradeoff.

\end{document}